\documentclass{article}

\usepackage{wrapfig}

\PassOptionsToPackage{numbers, compress}{natbib}

\usepackage[final]{neurips_2022}

\usepackage[utf8]{inputenc} 
\usepackage[T1]{fontenc}    
\usepackage{hyperref}       
\usepackage{url}            
\usepackage{booktabs}       
\usepackage{amsfonts}       
\usepackage{nicefrac}       
\usepackage{microtype}      
\usepackage{xcolor}         

\title{Ensemble of Averages: Improving Model Selection and Boosting Performance in Domain Generalization}

\usepackage{listings}

\usepackage{color, colortbl}
\definecolor{Gray}{gray}{0.9}

\usepackage{microtype}
\usepackage{booktabs} 

\usepackage{hyperref}

\usepackage{microtype}
\usepackage{graphicx}
\usepackage{subfigure}
\usepackage{booktabs} 

\usepackage{hhline}
\usepackage{url}
\usepackage{multirow}

\usepackage[T1]{fontenc}
\usepackage[utf8]{inputenc}
\usepackage{multirow}
\usepackage{placeins}
\usepackage{url}
\usepackage{amsthm}
\usepackage{float}
\usepackage{amsmath}
\usepackage{microtype}
\usepackage{xspace}
\usepackage{float}
\usepackage{booktabs} 

\usepackage{amssymb}
\usepackage{amsmath,bm}
\usepackage{enumitem}

\author{%
  Devansh Arpit, Huan Wang, Yingbo Zhou, Caiming Xiong\\
  Salesforce Research, USA\\
  \texttt{devansharpit@gmail.com} \\
}

\begin{document}

\maketitle
\begin{abstract}
In Domain Generalization (DG) settings, models trained independently on a given set of training domains have notoriously chaotic performance on distribution shifted test domains, and stochasticity in optimization (e.g. seed) plays a big role. This makes deep learning models unreliable in real world settings. We first show that this chaotic behavior exists even along the training optimization trajectory of a single model, and propose a simple model averaging protocol that both significantly boosts domain generalization and diminishes the impact of stochasticity by improving the rank correlation between the in-domain validation accuracy and out-domain test accuracy, which is crucial for reliable early stopping. Taking advantage of our observation, we show that instead of ensembling unaveraged models (that is typical in practice), ensembling moving average models (EoA) from independent runs further boosts performance. We theoretically explain the boost in performance of ensembling and model averaging by adapting the well known Bias-Variance trade-off to the domain generalization setting. On the DomainBed benchmark, when using a pre-trained ResNet-50, this ensemble of averages achieves an average of $68.0\%$, beating vanilla ERM (w/o averaging/ensembling) by $\sim 4\%$, and when using a pre-trained RegNetY-16GF, achieves an average of $76.6\%$, beating vanilla ERM by $6\%$. Our code is available at \url{https://github.com/salesforce/ensemble-of-averages}.
\end{abstract}

\section{Introduction}

Domain generalization (DG, \cite{blanchard2011generalizing}) aims at learning predictors that generalize well on data sampled from test distributions that are different from the training distribution. Currently, deep learning models have been shown to be poor at this form of generalization \cite{d2020underspecification}, and excel primarily in the IID setting \cite{zhou2021domain}.

While a number of algorithms have been proposed to mitigate this problem (cf \cite{zhou2021domain} for a survey), \cite{gulrajani2020search} demonstrate that models trained using empirical risk minimization (ERM, \cite{vapnik1998statistical}) along with proper model selection (i.e. early stopping using validation set), using a subset of data from all the training domains, largely match or even outperform the performance of most existing domain generalization algorithms. This suggests that model selection plays an important role in domain generalization.
Despite its importance, \textit{there has not been much investigation into the reliability of model selection}.
As we demonstrate in Figure \ref{fig:instability}, the out-domain performance varies greatly along the optimization trajectory of a model during training, even though the in-domain performance does not. This instability therefore hurts the reliability of model selection, and can become a problem in realistic settings where test domain data is unavailable, because it causes the rank correlation between in-domain validation accuracy and out-domain test accuracy to be weak.

In this paper, we first investigate a simple protocol for model averaging that both boosts DG within the ERM framework, and mitigates performance instability of deep models on out-domain data, specifically with respect to in-domain validation data. This makes model selection more reliable. Next, taking advantage of our observation, we show that ensembling moving average models further boosts performance, making it a better choice for practical scenarios. Note that we do not claim that model averaging or ensembling can fully solve the problem of DG. The observation that model averaging can boost domain generalization performance is not new, and was exposed by SWAD \cite{cha2021swad}, which inspired our work. Our contribution in this respect are as follows:

\begin{figure}
    \centering
        \includegraphics[width=0.45\columnwidth]{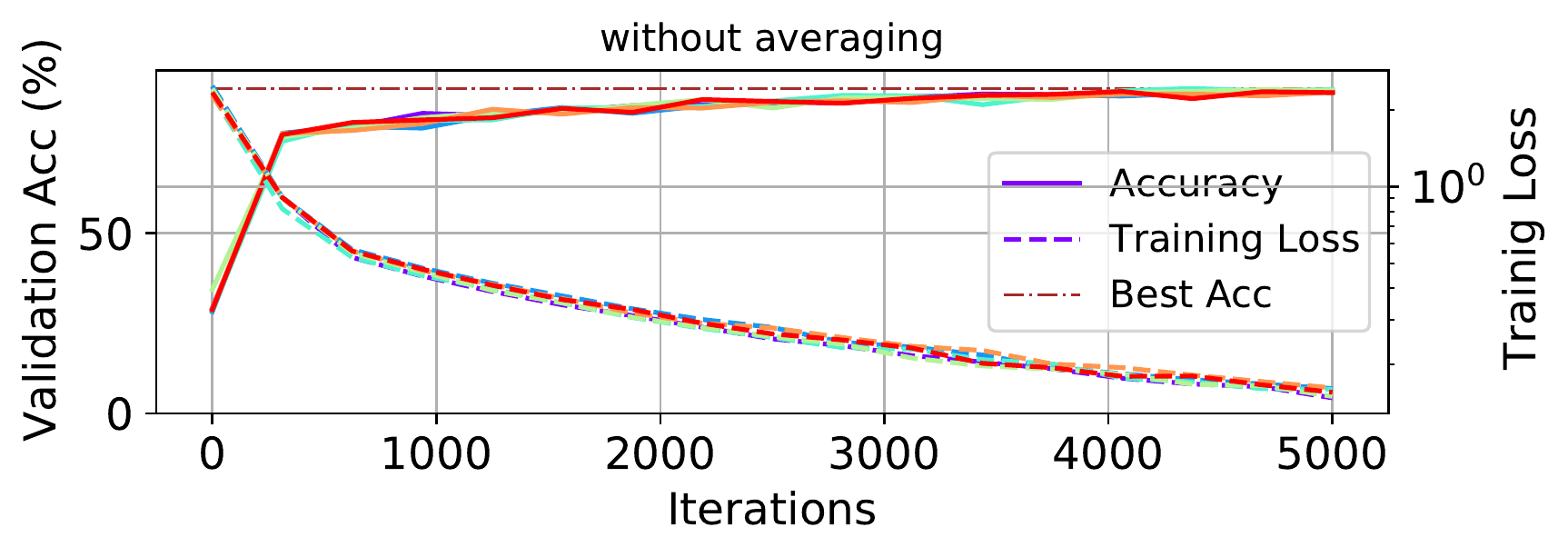}
        \includegraphics[width=0.45\columnwidth]{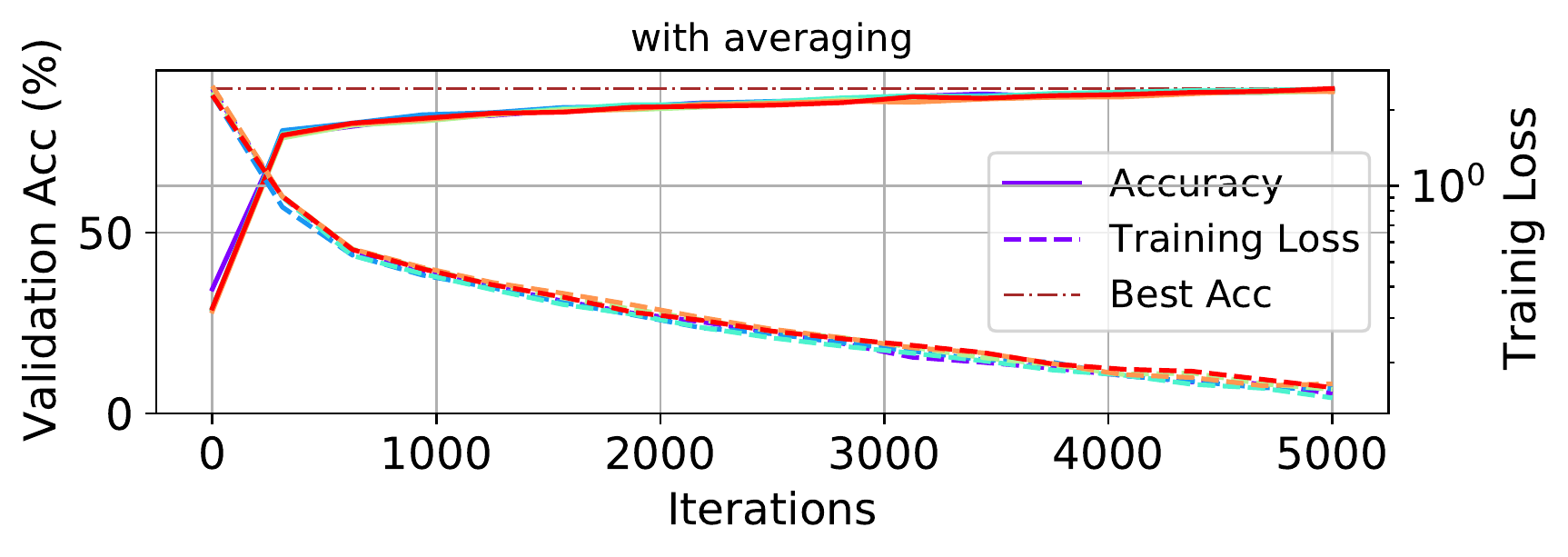}
      \includegraphics[width=0.45\columnwidth]{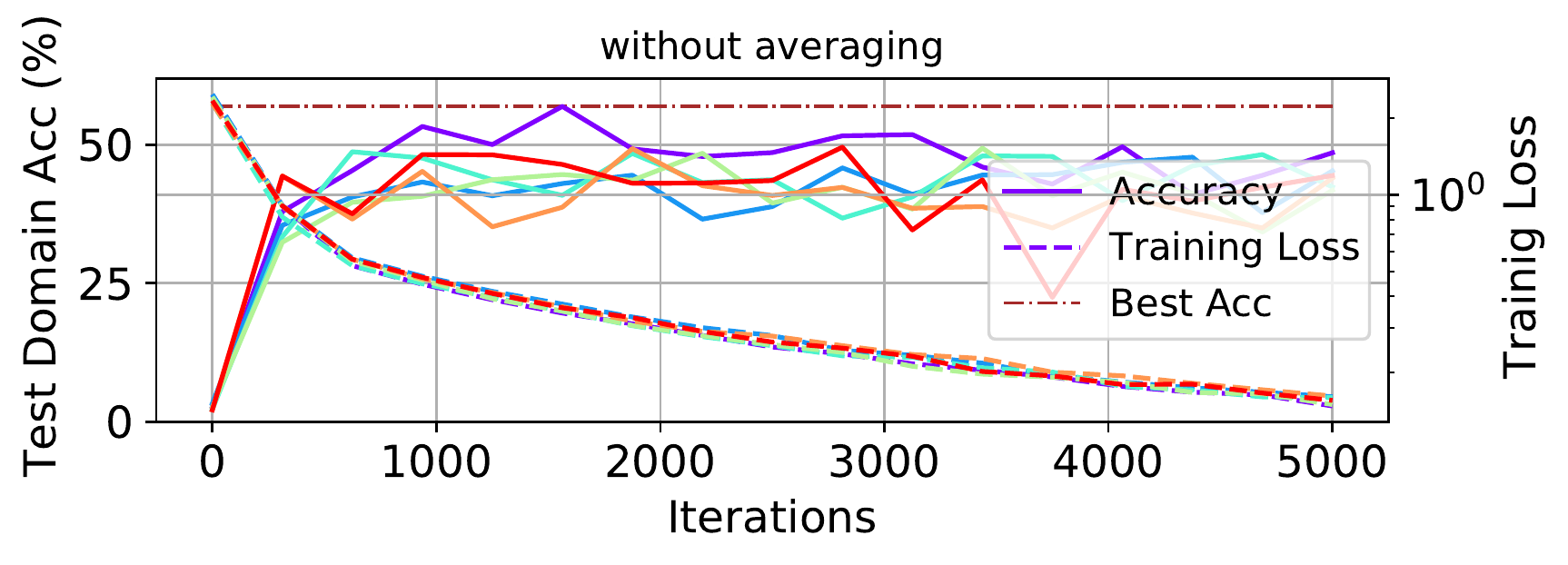}
      \includegraphics[width=0.45\columnwidth]{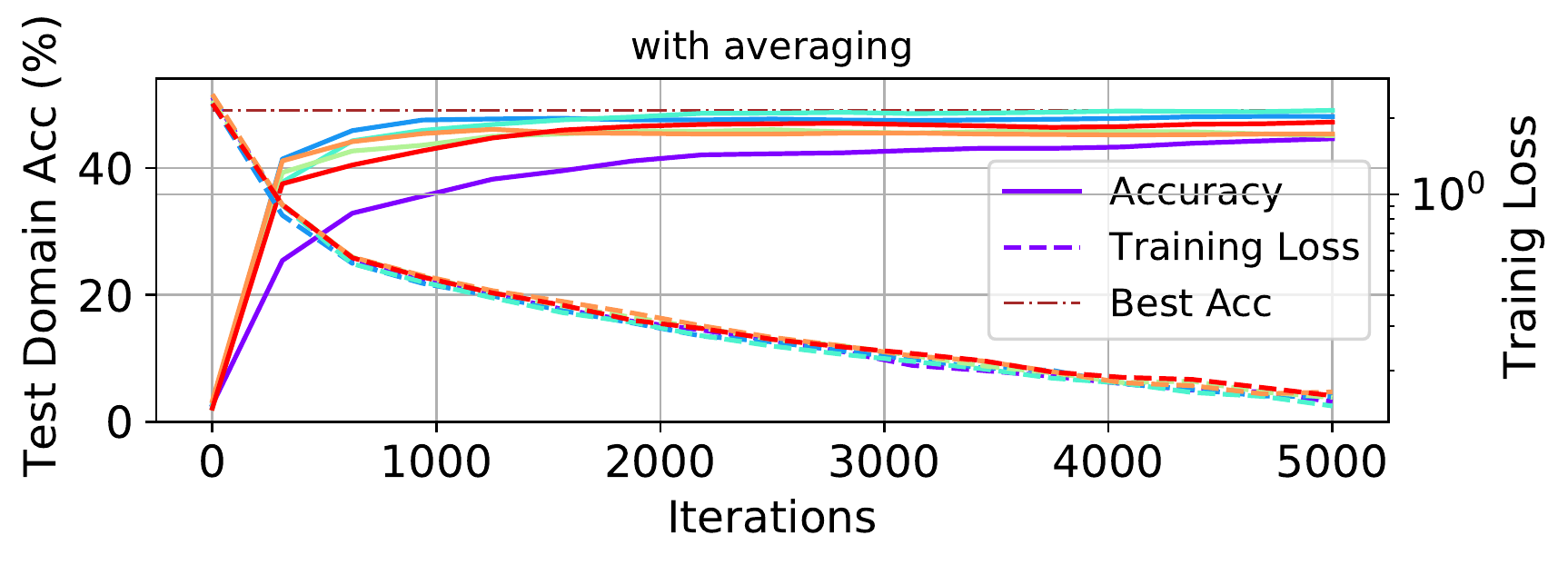}
    \caption{Model averaging improves out-domain performance \textit{stability}. \textbf{Left}: In-domain validation accuracy and out-domain test accuracy during training of models using ERM. \textbf{Right}: Same as left, except validation and test predictions are made using a simple moving average of the model being optimized, along its optimization path. \textbf{Details}: The plots are for the TerraIncognita dataset with domain L38 used as the test domain, and others as training/validation data, and ResNet-50. Solid lines denote accuracy, dashed lines denote training loss, and dash-dot lines denote best accuracy achieved during training and all runs (for reference). Each color denotes a different run with a different random seed and training/validation split. \textbf{Gist}: Model averaging reduces out-domain performance instability, and makes the test curves correlate better with the validation curves, making model selection using in-domain validation set more reliable during optimization. We see a similar pattern when using ensemble of models, with and without model averaging, in Figure \ref{fig:instability_ensemble_terra_l38}.}
    \label{fig:instability}
\vspace{-11pt}
\end{figure}

1. \textbf{Hyperparameter-free}:In contrast to SWAD, which introduces three additional hyper-parameters for its model averaging algorithm that need tuning, we show that the simple strategy of maintaining a simple moving average (SMA) of the model parameters throughout the optimization trajectory, starting near initialization (Appendix Figure \ref{fig:mo_start_iter}), works just as well (when a pre-trained model is used as initialization). Although model averaging technically requires two hyper-parameters-- averaging frequency and starting iteration, through empirical analysis, we show that setting the frequency to 1 and setting the start iteration close to 0 works well on multiple datasets and architectures, making our proposal hyperparameter-free in practice.

2. \textbf{Computationally efficient}:SWAD requires computing validation performance more frequently than is typically done (2x-6x on the DomainBed datasets), which is needed because it needs to find the start and end iteration between which model averaging is done. This increases compute requirements. This segment is selected based on the validation performance computed using the model being trained. Our proposal to instead use the SMA model to perform early stopping and inference, side-steps this need and does not require frequent validation performance check. We show that the root cause for this difference is that the model being trained has unstable performance on OOD data, while the SMA model has a more stable OOD performance (see Figure \ref{fig:instability} and Table \ref{tab:rank_correlation_single}). Thus this observation results in our hyperparameter-free and more efficient model averaging strategy.

3.  \textbf{EoA}: Taking advantage of our efficient model averaging protocol (section \ref{sec_protocol}), we find that an ensemble of moving average models (EoA) outperforms a traditional ensemble of unaveraged models (Table \ref{tab:benchmark}). We also show ablation analysis that the rank correlation between in-domain validation performance and out-domain test performance is better for the ensemble of average models (Table \ref{tab:rank_correlation_ensemble_single}).

4. \textbf{Theoretical explanation}: To explain why both model averaging and ensembling improve OOD performance under a unified theoretical framework, we adapt the well known Bias-Variance decomposition to the domain generalization setting, and argue that the expected OOD loss for individual models comprises of both the bias and the variance term, while the expected OOD loss for ensembles and averaged models comprises mainly of the bias term only, and is thus strictly lower (section \ref{sec_histogram_ablation}). Our explanation is in contrast with SWAD, which uses flat minima to explain the improved OOD generalization, which applies to model averaging, but is less straight forward for explaining the boost by ensembles.

5. \textbf{Benchmarking}: For benchmarking, we experiment with three different pre-trained models as initializations for DG training, with increasing pre-training dataset size and model size. In these experiments we find that EoA provides a larger gain over the \textit{corresponding} ERM baseline with increasing dataset and model size. These gains range from $4\%-6\%$ (Table \ref{tab:benchmark}). Notice that this claim is different from existing work \cite{hendrycks2021many}, which states that the baseline ERM performance improves with larger pre-training data and model size.

\section{Model Averaging}

\subsection{Terminology}

\textbf{Online Model}: For a given supervised learning objective function, let $f_{\theta}(.)$ denote the deep network being optimized using gradient based optimizer, where $\theta$ denotes the parameters of this model. We refer to $f_{\theta}$ as the \textit{online model}, or \textit{unaveraged model}. The output of $f_{\theta}(.)$ is a vector of $K$ logits corresponding to the $K$ classes in the supervised task.

\textbf{Moving Average (MA) Model}: While the online model is being trained, we maintain a moving average of the online model's parameters. This process is sometime referred to as \textit{iterate averaging} in existing literature. The deep network whose parameters are set to be this moving average is referred to as the \textit{moving average model}, or more specifically \textit{simple moving average (SMA) model} because of its use in our work. We denote the parameters of this model by $\hat{\theta}.$ 

\subsection{Model Averaging Protocol}
\label{sec_protocol}

We use a simple moving average (SMA) of the online model. Instead of calculating the moving average starting from initialization (as done in Polyak-Ruppert averaging), we instead start after a certain number of iterations $t_0$ during training (tail averaging), and maintain the moving average until the end of training. As we discuss in the next section, $t_0$ is chosen to be close, but not equal to the initialization when a pre-trained model is used as initialization. At any iteration $t$, we denote:
\begin{equation}
\label{eq_my_sma}
    \hat{\theta}_t = \begin{cases}
    \theta_{t},& \text{if } t\le t_0\\
    \frac{t-t_0}{t-t_0 +1} \cdot \hat{\theta}_{t-1} + \frac{1}{t-t_0 +1} \cdot \theta_{t},              & \text{otherwise}
\end{cases}
\end{equation}
where $\theta_t$ is the online model's state at iteration $t$. Note that effectively, $\hat{\theta}_t := \frac{1}{t-t_0+1}\cdot \sum_{t^\prime=t_0}^t {\theta}_{t^\prime}$. Further, at iteration $t$, if we need to calculate validation performance, we use $\hat{\theta}_t$ to do so, and not $\theta_{t}$. As we show in the next section, the benefit of doing so is that the rank correlation between in-domain validation accuracy and out-domain test accuracy is significantly better when predictions are made using $\hat{\theta}_t$. This makes model selection more reliable for domain generalization. Finally, for a given run, model selection selects $\hat{\theta}_{t^*}$ for making test set predictions, such that $\hat{\theta}_{t^*}$ achieves the best validation performance. We discuss some theoretical perspectives on why model averaging can help domain generalization in section \ref{sec_relatedw_ma_models}.

\subsection{Ablation Analysis}
\label{sec_ma_analysis}

Here we perform four ablation studies: 1) impact of the start iteration $t_0$ used in our SMA protocol in Eq. \ref{eq_my_sma}; 2) the frequency of model averaging; 3) instability reduction of SMA model compared to the online mode along the optimization trajectory on out-domain data; 4) correlation between in-domain and out-domain accuracy across independently trained models. 

Due to space limitation, we show experiments for 1,2 and 4 in Appendix section \ref{sec_ablation_sm}. In summary, we find that: 1) starting averaging close to initialization results in improved out-domain performance (Figure \ref{fig:mo_start_iter} in Appendix) when the parameters are initialized used a pre-trained model; 2) the frequency of SMA does not have a significant impact on performance, unless sampling is done at too large intervals (Figure \ref{fig:sma_freq} in Appendix); 4) the rank correlation is poor between validation and test accuracy of \textit{independently} trained models (Figure \ref{fig:cross_run_spearman} in Appendix). An implication of this is that it is difficult to discover the best model (for out-domain performance) from a pool of independently trained models, based only on their in-domain validation performance (echoing the findings of \cite{d2020underspecification}).

\begin{table*}[t]
\caption{Spearman correlation (closer to 1 is better) between within-run in-domain validation accuracy and out-domain test accuracy on multiple datasets. Model averaging improves rank correlation for both individual models (left) and ensemble of averages (right).}
\begin{minipage}{.5\linewidth}
      \caption{Individual Models}
      \centering
        \begin{tabular}{lll}
        \hhline{===}
        TerraIncognita    & w/o avg & w/ avg   \\ \hline
        L100     & 0.21 $\pm$ 0.07   & \textbf{0.90 $\pm$ 0.05}   \\
        L38 & 0.12 $\pm$ 0.13   & \textbf{0.83 $\pm$ 0.05}  \\
        L43   & 0.30 $\pm$ 0.06   & \textbf{0.67 $\pm$ 0.18} \\
        L46  & 0.03 $\pm$ 0.11   & \textbf{0.52 $\pm$ 0.14}  \\ 
        \hline
        \end{tabular}
        \label{tab:rank_correlation_single}
    \end{minipage}%
    \begin{minipage}{.5\linewidth}
      \centering
        \caption{Ensembles}
        \begin{tabular}{lll}
        \hhline{===}
        TerraIncognita    & w/o avg & w/ avg   \\ \hline
        L100     & 0.48   & \textbf{1}   \\
        L38 & 0.17   & \textbf{0.95}  \\
        L43   & \textbf{0.59}   & 0.38 \\
        L46  & 0.08   & \textbf{0.61}  \\ 
        \hline
        \end{tabular}
        \label{tab:rank_correlation_ensemble_single}
    \end{minipage} 
\end{table*}

\subsubsection{Instability Reduction: Rank Correlation}
\label{sec_rank_corr}

We study the reliability of model selection for domain generalization when using online models vs moving average models, using rank correlation (see Appendix \ref{sec_rank_corr_intro} for definition). To do so, we train models on a dataset, both with and without model averaging, and compute Spearman correlation between the in-domain validation accuracy and out-domain test accuracy sampled at regular intervals during the training process. Since there are multiple runs where a given domain acts as the test domain, we calculate the mean and standard error of these values over these runs.

The rank correlations are shown in Table \ref{tab:rank_correlation_single} (and Table \ref{tab:rank_correlation} in Appendix) for the PACS, VLCS, OfficeHome, TerraIncognita and DomainNet datasets. We find that in majority of the cases, using model averaging results in a significantly better rank correlation compared to using the online model. These experiments therefore suggest that the reliability of model selection is significantly higher within a run when using model averaging.

\section{Ensemble of Averages (EoA)}
\label{sec_eoa}

\cite{gulrajani2020search} propose a rigorous framework for evaluation in the domain generalization setting which accounts for randomness due to seed and hyper-parameter values, and recommend reporting the average test accuracy over all the runs computed using a model selection criteria. However, in practice, it is desirable to have a single predictor that has a high accuracy. An ensemble combines predictions from multiple models, and is a well known approach for achieving this goal \cite{dietterich2000ensemble} by exploiting function diversity \cite{fort2019deep}. However, as we show, even ensembles suffer from instability in the domain generalization setting. Building on the observations of the previous section, we investigate the behavior of ensemble of moving average models and find that it mitigates this issue. We begin by describing the EoA protocol below.

\textbf{EoA Protocol}: We perform experiments with ensemble of multiple independently trained models (i.e., with different hyper-parameters and seeds). When each of these models are moving average models from their corresponding runs, we refer to this ensemble in short as the \textit{ensemble of averages (EoA)}. Identical to how we make predictions for traditional ensembles (specifically the bagging method \cite{breiman1996bagging}), the class $\hat{y}$ predicted by an EoA for an input $\mathbf{x}$ is given by the formula:
\begin{align}
    \hat{y} = \arg \max_{k} Softmax(\frac{1}{E} \sum_{i=1}^E f(\mathbf{x}; {\hat{\theta}_i}))_k
\end{align}
where $E$ is the total number of models in the ensemble, $\hat{\theta_i}$ denotes the parameters of the $i^{th}$ moving average model, and the sub-script $(.)_{k}$ denotes the $k^{th}$ element of the vector argument. Finally, the state $\hat{\theta_i}$ of the $i^{th}$ moving average model used in the ensemble is selected from its corresponding run using its in-domain validation set performance (described in section \ref{sec_protocol}). We now investigate the behavior of EoA compared with ensembles of online models on domain generalization tasks.

\begin{figure*}[t]
    \centering
        \includegraphics[width=0.45\columnwidth]{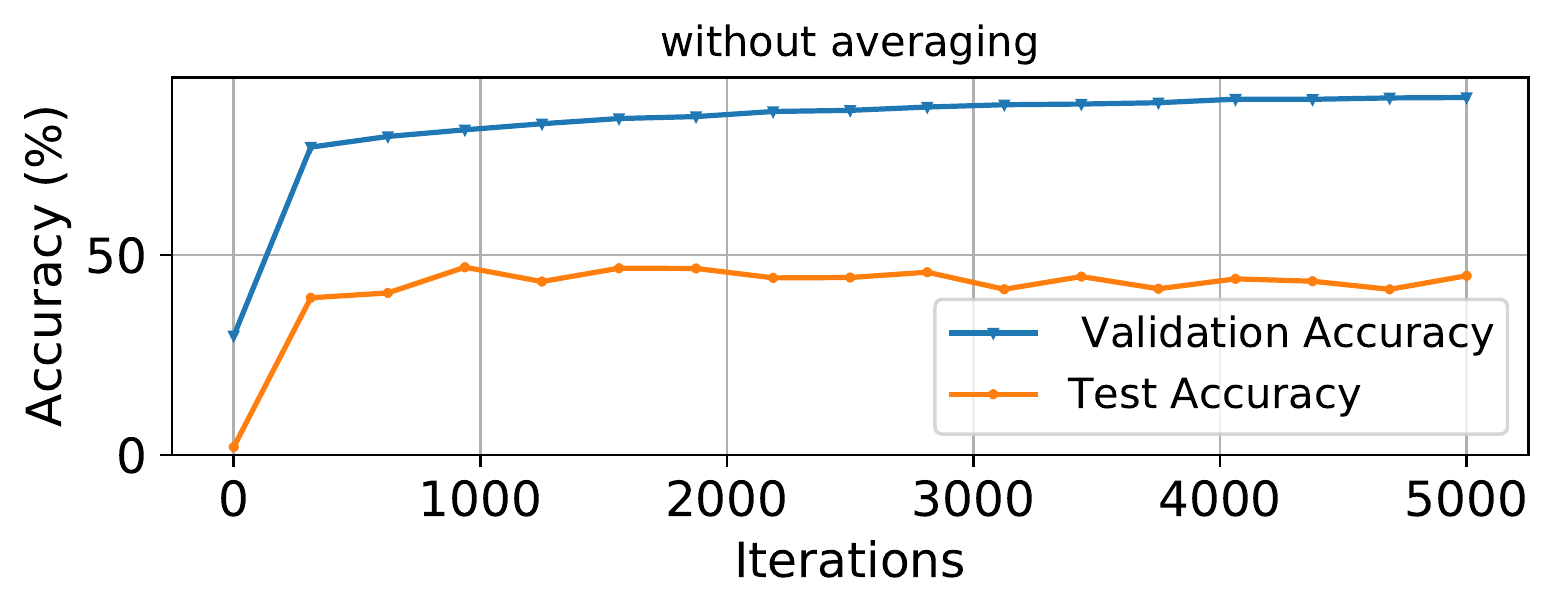}
      \includegraphics[width=0.45\columnwidth]{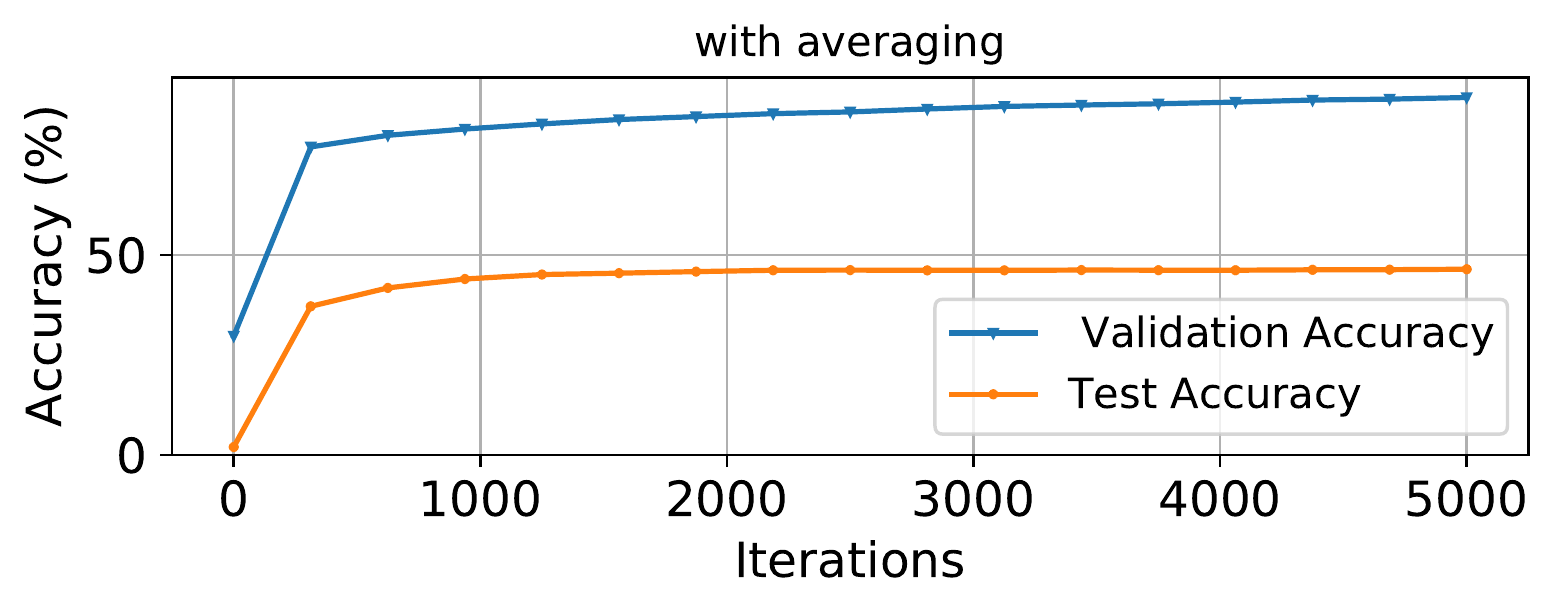}
    \caption{Ensemble of moving averages (EoA) (right) has better out-domain test performance \textit{stability} compared with ensemble of online models (left), w.r.t. in-domain validation accuracy. \textbf{Details}: The plots are for the TerraIncognita dataset with domain L38 used as the test domain, and others as training/validation domain, and ResNet-50. Each ensemble has 6 different models from independent runs with different random seeds, hyper-parameters, and training/validation split.}
    \label{fig:instability_ensemble_terra_l38}
\vspace{-11pt}
\end{figure*}

\subsection{Analysis}

\textbf{Qualitative visualization}: For the purpose of contrasting the behavior of traditional ensembles vs ensemble of averages, we begin by qualitatively studying the stability of out-domain performance of these two ensembling techniques during the training process. To do so, we use the TerraIncognita dataset, and fix one of its domains as the test domain while using the others as training/validation data. We then train 6 different models independently for $5,000$ iterations with different seeds, hyper-parameters and training-validation splits identical to the \cite{gulrajani2020search} protocol. We also maintain moving average models corresponding to each of these 6 models. At every $300$ iterations, we form an ensemble of the 6 online models from their corresponding runs and compute the out-domain test accuracy. Since, each run has a different training-validation split, we calculate the mean validation accuracy of each of these online models at that iteration. We follow an identical procedure for the moving average models and plot these performances in Figure \ref{fig:instability_ensemble_terra_l38}. We find that the ensemble of averages has a better stability on out-domain test set compared to the ensemble of online models. 

For clarity, note that this procedure for calculating test accuracy at regular intervals is different from what we proposed earlier for EoA for practical purposes. This experiment is only meant to highlight the fact that making predictions on out-domain data using an ensemble of online models suffers from instability along the optimization trajectory, while an ensemble of averages mitigates this issue. For plots on other domains of TerraIncognita, see Figure \ref{fig:instability_ensemble_terra_all} in the Appendix.

\textbf{Rank correlation}: We now measure the rank correlation between in-domain validation accuracy and out-domain test accuracy for a quantitative evaluation. The details of the metric and motivations behind this experiment are same as those described in section \ref{sec_rank_corr}. Here we use the same experimental setup described in the qualitative analysis above. But in addition, we also conduct experiments on VLCS, OfficeHome and DomainNet datasets. The results are shown in Table \ref{tab:rank_correlation_ensemble_single} (and Table \ref{tab:rank_correlation_ensemble} in Appendix).  We find that in majority of the cases, using EoA results in a significantly better rank correlation compared to using the online model ensemble. These results show more concretely the fact that predictions by an ensemble of online models on out-domain data suffers from instability along the optimization trajectory, and EoA mitigates this problem.

\subsection{Why does Ensembling and Model Averaging Improve Performance?}
\label{sec_histogram_ablation}
We explain the performance boost achieved by ensemble of averages (see next section) by adapting the Bias-Variance decomposition \cite{geman1992neural} to the domain generalization setting. For classification tasks with one-hot labels, the Bias-Variance decomposition is given as \cite{yang2020rethinking},
\begin{align}
\mathbb{E}_{\mathbf{x}, {y}} \mathbb{E}_{\mathcal{T}} [CE({y}, f(\mathbf{x}; \mathcal{T}))] &=
    \underbrace{\mathbb{E}_{\mathbf{x}, {y}}[CE({y}, \bar{f}(\mathbf{x}))]}_{\text{Bias}^2} + \underbrace{\mathbb{E}_{\mathbf{x},\mathcal{T}}[KL(\bar{f}(\mathbf{x}), f(\mathbf{x}; \mathcal{T}))]}_{\text{Variance}}\nonumber
\end{align}
where $CE$ denotes the cross entropy loss, $KL$ denotes KL divergence, $\mathcal{T} = \{(\mathbf{x}^{in}_i, y^{in}_i)\}_{i=1}^N$ are $N$ IID samples drawn from the in-domain training distribution $\mathbb{P}^{in}$, $f(\mathbf{x}; \mathcal{T})$ denotes the prediction of the model $f$ on sample $\mathbf{x}$ such that the model is trained on the dataset $\mathcal{T}$, and $\bar{f}(\mathbf{x}) = \mathbb{E}_{\mathcal{T}}[f(\mathbf{x}; \mathcal{T})]$. Finally $(\mathbf{x}, y) \sim \mathbb{P}^{out}$ where $\mathbb{P}^{out}$ is the out-domain distribution. Notice how $\mathcal{T}$ and $(\mathbf{x}, y)$ come from different distributions. For instance, in PACS dataset, $\mathbb{P}^{in}$ could be the union of art, cartoon and photo domains, and $\mathbb{P}^{out}$ could be the sketch domain. 

The L.H.S. of the above equation is the expected cross entropy loss on the out-domain distribution achieved by individual models, i.e., when we train an individual model on a particular instance of the training dataset $\mathcal{T}$, the expected out-domain test loss is denoted by L.H.S. Importantly, the Bias term on the R.H.S. denotes the expected cross entropy loss on the out-domain distribution achieved by the function $\bar{f}(.)$, which is essentially an \textit{ensemble}. Finally, the variance term captures how much the prediction of individual models differs in expectation from the ensemble prediction, which makes this term strictly greater than zero. 

Therefore, the above decomposition tells us that the expected test domain error of an ensemble is strictly less than that of an individual model. This interpretation directly explains why a traditional ensemble of unaveraged models can be expected to perform better than individual unaveraged models. However, it is still not clear why EoA performs better that a traditional ensemble in practice. To establish this connection, we note that in practice, we typically train a small number of independent models to form a traditional ensemble due to computational constraints. Thus such ensembles do not behave identically to the expected ensemble $\bar{f}(.)$ described above. Model averaging on the other hand has been shown to approximate an ensemble \cite{izmailov2018averaging}. To see this, consider without any loss of generality that the ensemble contains models with parameters $\{\theta_1, \theta_2 \hdots \theta_T\}$, and denote $\hat{\theta}_T := \frac{1}{T}\cdot \sum_{t=1}^T {\theta}_t$. Then note that the second order Taylor's expansion around $\hat{\theta}_{T}$ of each model's $k^{th}$ dimension's prediction is given by, 
\begin{align}
\label{eq_taylor_expansion}
    \frac{1}{T}\cdot \sum_{t=1}^{T} f({\theta_t})_k &\approx f(\hat{\theta}_{T})_k + \frac{1}{T}\cdot \sum_{t=1}^{T}  (\hat{\theta}_{T} - {\theta}_t)^T \frac{\partial f({\hat{\theta}_{T}})_k }{\partial \hat{\theta}_{T}}\nonumber+  0.5(\hat{\theta}_{T} - {\theta}_t)^T  \frac{\partial^2 f({\hat{\theta}_{T}})_k }{\partial \hat{\theta}_{T}^2} (\hat{\theta}_{T} - {\theta}_t)
\end{align}

\begin{figure*}[t]
    \centering
        \includegraphics[width=0.45\columnwidth,trim={0cm 2cm 15cm 0},clip]{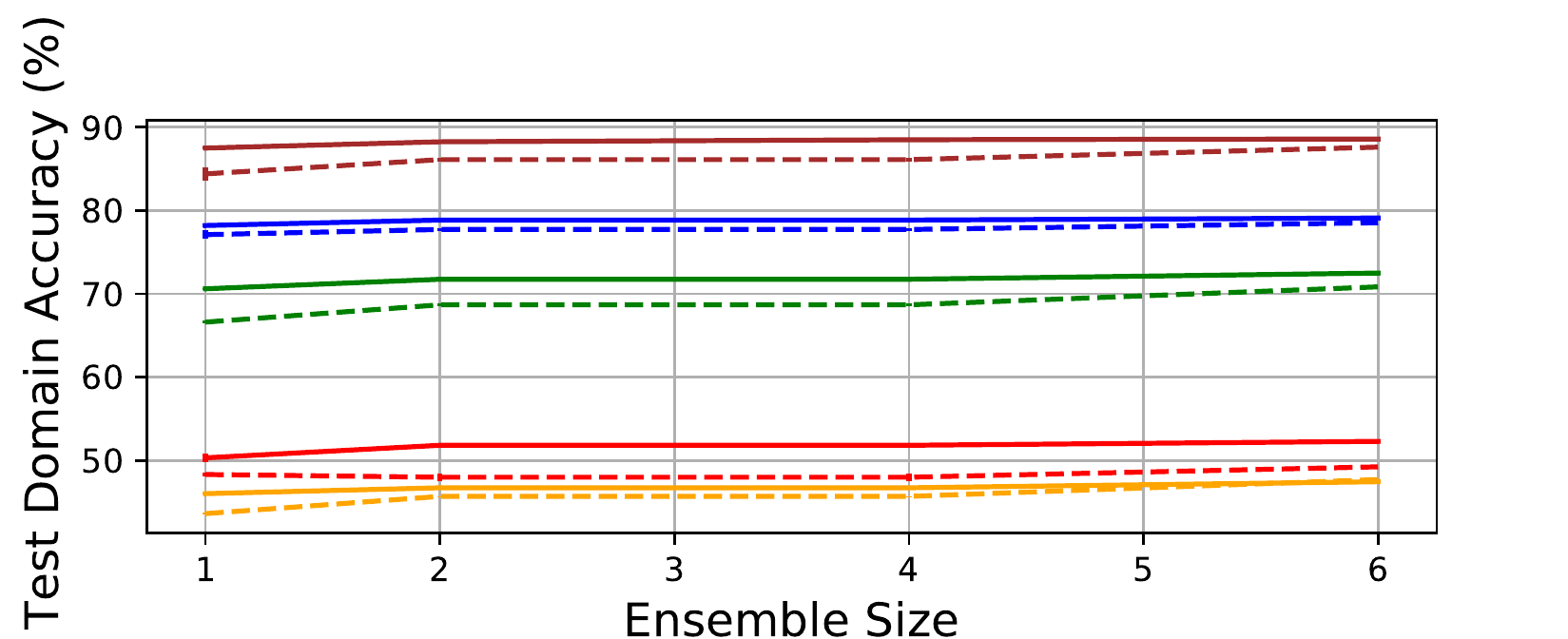}
      \includegraphics[width=0.45\columnwidth,trim={0cm 2cm 15cm 0},clip]{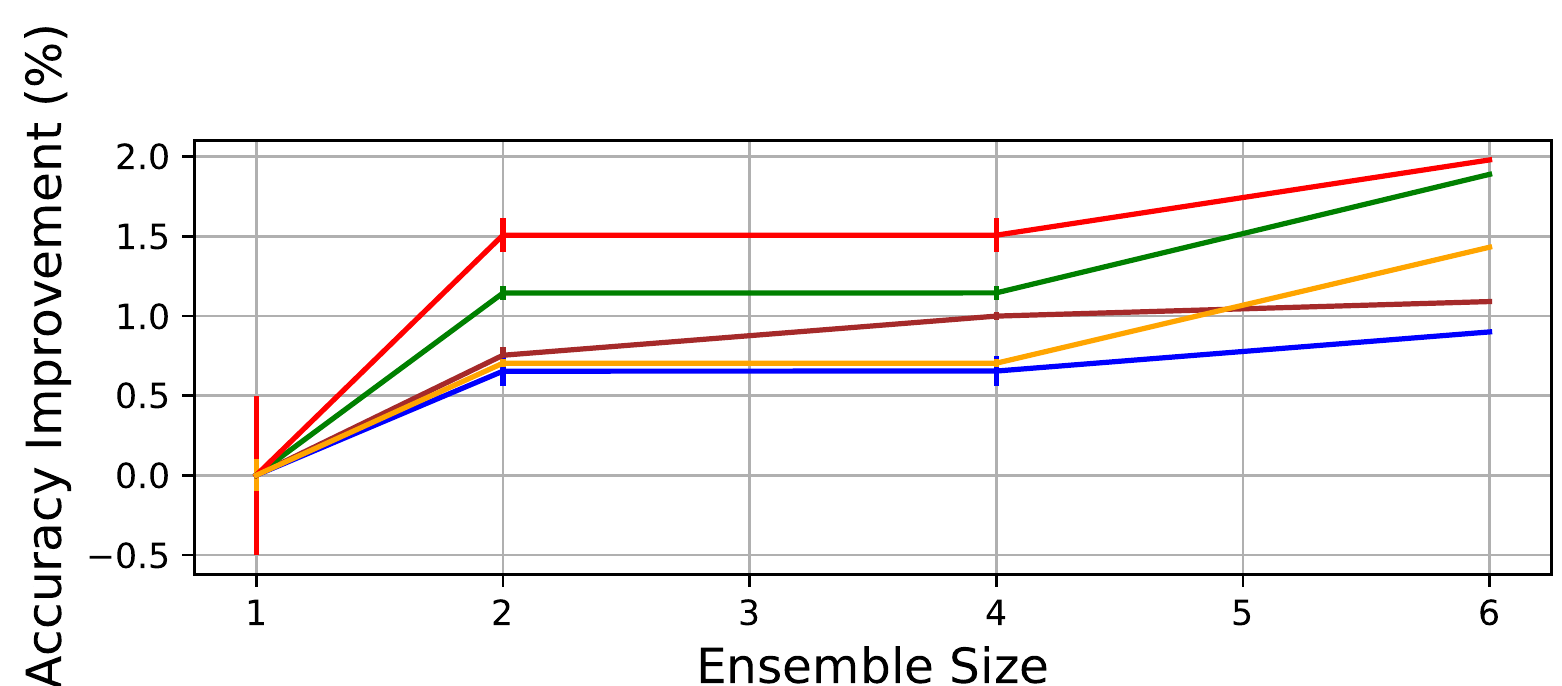}
      \vfill
      \includegraphics[width=1\columnwidth,trim={1cm 0cm 0 7.5cm},clip]{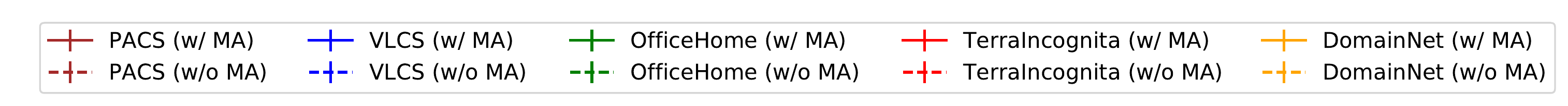}
    \caption{\textbf{Left}: Effect of ensemble size (number of models in an ensemble) on out-domain performance (mean and standard error) for models with and without moving average (MA) parameters for ResNet-50 pre-trained on ImageNet. \textbf{Right}: Using the performance of ensemble of size 1 (shown in the left plot) as reference, right plot shows the percentage point improvement for ensembles of size $> 1$. The plots show that i) ensemble of averages (solid lines in left plot) are consistently better than ensemble of models without averaging (dashed lines in left plot); ii) ensemble of averages consistently improves performance over averaged models (ensemble of size 1 in right plot). }
    \label{fig:ensemble_size}
\vspace{-11pt}
\end{figure*}

\begin{wrapfigure}{r}{0.5\textwidth}
    \centering
        \includegraphics[width=0.45\columnwidth,trim={0cm 0.25cm 0cm 0},clip]{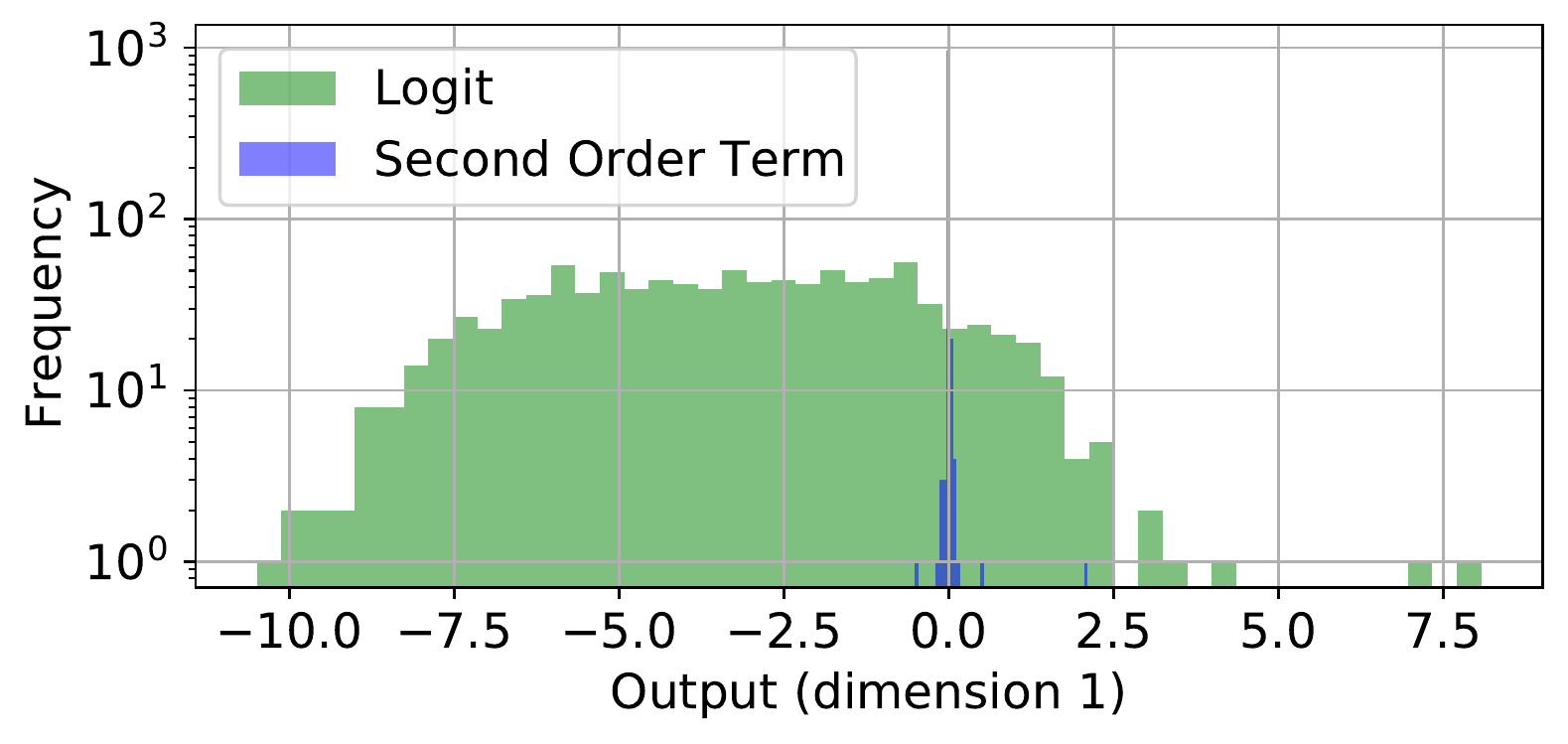}
    \caption{The scale of terms-- moving average model's logit and the second order term in Eq. \ref{eq_taylor_expansion}. The latter concentrates around 0, suggesting our model averaging protocol approximates ensembles.}
    \label{fig:ablation_dim1}
\end{wrapfigure}
Notice that $f(.)$ is the model output and therefore the first and second order terms are the derivatives of the model output and not the loss gradient and Hessian. The first order term is zero due to $\hat{\theta}_T := \frac{1}{T}\cdot \sum_{t=1}^T {\theta}_t$. A crucial difference of our analysis compared to \cite{izmailov2018averaging} is that they average model states that lie near different loss minima, while we perform tail averaging. Therefore, the term $(\hat{\theta}_{T} - {\theta}_t)$ may not behave similar to that in their case. To shed light on its behavior, we plot the histogram of the second order term and the moving average model's logit $f(\hat{\theta}_{T})_k$ in Eq. \ref{eq_taylor_expansion} for the first dimension ($k=1$) for test domain data in figure \ref{fig:ablation_dim1} (details and additional experiments provided in Appendix \ref{sec_ablation_details}). The histogram shows that the second order term concentrates near zeros while the logit values span a wider range, which implies that under the second order approximation, the model averaging protocol used in our work behaves like an ensemble. Finally, to study the impact of ensemble size on out-domain performance, we plot the test domain accuracy as a function of ensemble size in figure \ref{fig:ensemble_size}. The plots show that i. EoA outperforms traditional ensembles for all ensemble sizes (left); and ii. ensembles of larger size typically have better out-domain performance. See a discussion on functional diversity of ensembles vs model averaging in Appendix \ref{sec_discussion}.

\begin{table*}[t]
\small
\vspace{-12pt}
\centering
\caption{Performance benchmarking on 5 datasets of the DomainBed benchmark using two different pre-trained models. SWAD and MIRO are the previous SOTA. \textbf{See Table \ref{tab:benchmark_full} in Appendix for comparison with more methods.} Note that ensembles do not have confidence interval because an ensemble uses all the models to make a prediction. Gray background shows our proposal. \textit{Our runs} implies we ran experiments, but we did not propose it. Experiments use the training-domain validation protocol from \cite{gulrajani2020search}.}
\begin{tabular}{llllll|l}
\hline
{Algorithm}          & {PACS}           & {VLCS}           & {OfficeHome}     & {TerraIncognita} & {DomainNet}      & {Avg}. \\ \hhline{======|=}
\multicolumn{7}{c}{{ResNet-50 (25M Parameters, Pre-trained on ImageNet)}}\\
\hline
ERM (our runs) & 84.4 $\pm$ 0.8 & 77.1 $\pm$ 0.5 & 66.6 $\pm$ 0.2 & 48.3 $\pm$ 0.2 & 43.6 $\pm$ 0.1 & 64.0 \\
Ensemble (our runs) & 87.6             & 78.5           &       70.8         & 49.2           &          \textbf{47.7}      &   66.8   \\
ERM \cite{gulrajani2020search} & 85.7 $\pm$ 0.5 & 77.4 $\pm$ 0.3 & 67.5 $\pm$ 0.5 & 47.2 $\pm$ 0.4 & 41.2 $\pm$ 0.2 & 63.8\\ 
SWAD \cite{cha2021swad} & 88.1 $\pm$ 0.4 & \textbf{79.1 $\pm$ 0.4} & 70.6 $\pm$ 0.3 & 50.0 $\pm$ 0.4 & 46.5 $\pm$ 0.2 & 66.9\\
MIRO \cite{cha2022domain} & 85.4 $\pm$ 0.4 & {79.0 $\pm$ 0.} & 70.5 $\pm$ 0.4 & 50.4 $\pm$ 1.1 & 44.3 $\pm$ 0.2 & 65.9\\
\hline
\rowcolor{Gray}
SMA (ours)    & 87.5 $\pm$ 0.2 & 78.2 $\pm$ 0.2   & 70.6 $\pm$ 0.1   & 50.3 $\pm$ 0.5 & 46 $\pm$ 0.1 & 66.5 \\
\rowcolor{Gray}
EoA (ours) & \textbf{88.6}           & \textbf{79.1}           & \textbf{72.5}          & \textbf{52.3}           &        {47.4}        & \textbf{68.0}\\
\hhline{======|=}
\multicolumn{7}{c}{{ResNeXt-50 32x4d  \cite{yalniz2019billion} (25M Parameters, Pre-trained 1B Images)}}\\
\hline
ERM (our runs) & 88.9 $\pm$ 0.3 & 79.0 $\pm$ 0.1 & 70.9 $\pm$ 0.5 & 51.4 $\pm$ 1.2 & 48.1 $\pm$ 0.2 & 67.7 \\ 
Ensemble (our runs) & 91.2             & 80.3           &       77.8         & 53.5           &          52.8      &   71.1   \\
 \hline
 \rowcolor{Gray}
SMA (ours)    & {92.7 $\pm$ 0.3} & {79.7 $\pm$ 0.3}   & {78.6 $\pm$ 0.1}   & {53.3 $\pm$ 0.1} & {53.5 $\pm$ 0.1} & {71.6} \\ 
\rowcolor{Gray}
EoA (ours) & \textbf{93.2}           & \textbf{80.4}           &  \textbf{80.2}  & \textbf{55.2}   &  \textbf{54.6}   & \textbf{72.7} \\
\hhline{======|=}
\multicolumn{7}{c}{{RegNetY-16GF \cite{singh2022revisiting} (81M Parameters, Pre-trained on 3.6B Images)}}\\
\hline
ERM (our runs) & 92 $\pm$ 0.4 & 78.6 $\pm$ 0.6 & 73.8 $\pm$ 0.5 & 55.6 $\pm$ 0.9 & 53.1 $\pm$ 0.2 & 70.6\\ 
Ensemble (our runs) &     95.1        &     80.6       &        80.5       &      59.5     &     57.8       &   74.7\\
ERM \cite{cha2022domain} & 89.6 $\pm$ 0.4 & 78.6 $\pm$ 0.3 & 71.9 $\pm$ 0.6 & 51.4 $\pm$ 1.8 & 48.5 $\pm$ 0.6 & 68.0 \\
SWAD \cite{cha2022domain}& 94.7 $\pm$ 0.2 & 79.7 $\pm$ 0.2 & 80.0 $\pm$ 0.1 & 57.9 $\pm$ 0.7 & 53.6 $\pm$ 0.6 & 73.2 \\
MIRO \cite{cha2022domain}& \textbf{97.4 $\pm$ 0.2} & 79.9 $\pm$ 0.6 & 80.4 $\pm$ 0.2 & 58.9 $\pm$ 1.3 & 53.8 $\pm$ 0.1 & 74.1\\
 \hline
 \rowcolor{Gray}
SMA (ours)      & 95.5 $\pm$ 0.0 & {80.7 $\pm$ 0.1} & {82.0 $\pm$ 0.0} & {59.7 $\pm$ 0.0} & {60.0 $\pm$ 0.0} & {75.6}\\
\rowcolor{Gray}
EoA (ours) & {95.8}           & \textbf{81.1}           &  \textbf{83.9}  & \textbf{61.1}   &  \textbf{60.9}   & \textbf{76.6} \\
\hline
\end{tabular}
\label{tab:benchmark}
\vspace{-12pt}
\end{table*}

\section{Empirical Results}
\subsection{DomainBed Benchmarking}
\vspace{-5pt}
\label{sec_benchmarking}

We now benchmark our model averaging protocol (SMA) and ensemble of averages against online models (ERM, without MA) and ensemble of online models (ensembles). Note that all these models are trained using the ERM objective as before. We evaluate on PACS \cite{li2017deeper}, VLCS \cite{fang2013unbiased}, OfficeHome \cite{venkateswara2017deep}, TerraIncognita \cite{beery2018recognition} and DomainNet \cite{peng2019moment} datasets in DomainBed. The training-evaluation protocols are the same as described in section \ref{sec_ma_analysis} for moving average and online models, and in section \ref{sec_eoa} for ensembles. Full details can be found in section \ref{sec_training_protocol} in the Appendix.

\textbf{Comparison with existing results using ResNet-50 pre-trained on ImageNet}: Here we compare existing methods with our runs. All methods use ResNet-50 (25M parameters) \cite{he2016deep} pre-trained on ImageNet as initialization. Comparing ERM \cite{gulrajani2020search} and ERM (our runs), we find that they perform similarly, especially considering we have used a smaller hyper-parameter space (further discussion in Appendix \ref{sec_discussion}). A comparison between SWAD and SMA shows that SWAD is slightly better (by $0.4\%$ on average). However, recall that our protocol retains the advantage of not tuning any hyper-parameters while SWAD has 3 additional ones that they tune separately in addition to the optimization hyper-parameters. Interestingly, traditional ensembles and SMA achieve similar performance ($66.8\%$ and $66.5\%$ respectively). Finally, EoA outperforms all the existing results: ERM by $4\%$ and SWAD (previous SOTA) by $1.1\%$. Importantly, note that while all non-ensemble models report the average test accuracy of multiple models following the protocol of \cite{gulrajani2020search}, EoA test accuracy is achieved by a single predictor that combines the output of multiple models.

\textbf{Experiments with larger pre-training datasets and larger models}: In addition to ResNet-50 pre-trained on ImageNet, we now also experiment with ResNeXt-50 32x4d (25M parameters), that is pre-trained using semi-weakly supervised objective on Instagram 1B images and ImageNet labeled data \cite{yalniz2019billion}, and RegNetY-16GF (81M parameters) pre-trained using Instagram 3.6B images. Note that both ResNet-50 and ResNeXt-50 32x4d have similar number of parameters, while RegNetY-16GF has more than 3x the number of parameters. On the other hand, also notice that the three architectures are respectively pre-trained on an increasing size of datasets. The rationale behind this choice is that recent trends in deep learning has shown that models pre-trained on larger datasets and architectures achieve better downstream transfer performance \cite{dosovitskiy2020image,mahajan2018exploring,hendrycks2021many}. Therefore, we expect the latter models to improve the ERM baseline, and our goal is to investigate the out-domain performance gain by model averaging and EoA relative to the \textit{corresponding} ERM baseline with increasing pre-training dataset size and model size. 

The experimental results are shown in Table \ref{tab:benchmark}. To investigate models with the same size, but one pre-trained on a larger dataset, we compare the results of ResNet-50 and ResNeXt-50 32x4d. On average across all five datasets, the gain of SMA over ERM (our runs) is $2.5\%$ for ResNet-50 and $3.9\%$ for ResNeXt-50 32x4d. The gain of EoA over ERM is larger: $4\%$ vs $5\%$ respectively. This suggests that pre-training the model on a larger dataset increases the gain of model averaging and EoA over the \textit{corresponding} ERM baseline, while the ERM performance itself improves.

Next, to investigate the impact of both larger model size and larger pre-training dataset, we compare the results of ResNeXt-50 32x4d and RegNetY-16GF. On average across all five datasets, the gain of SMA over ERM (our runs) is $3.9\%$ for ResNeXt-50 32x4d and $5\%$ for RegNetY-16GF. The gain of EoA over ERM is again larger: $5\%$ vs $6\%$ respectively. This suggests that increasing both model size and pre-training dataset size allow model averaging and EoA to provide larger out-domain gains over the {corresponding} ERM baseline. Notice that these claims are different from existing work \cite{hendrycks2021many}, which states that the baseline ERM performance improves with larger pre-training data and model size.

\begin{wraptable}{r}{0.5\textwidth}
\small
\vspace{-20pt}
\centering
\caption{ SMA outperforms ERM without model averaging in the IID setting.}
\begin{tabular}{|l|l|l|}
\hline
Algorithm & PACS             & OfficeHome       \\ \hline
ERM (no averaging)      & 94.39 $\pm$ 0.46 & 77.09 $\pm$ 0.57 \\ \hline
SMA (ours) & \textbf{96.77 $\pm$ 0.20} & \textbf{83.56 $\pm$ 0.21} \\ \hline
\end{tabular}
\label{tab_iid}
\end{wraptable}

\subsection{In-domain Performance Improvement using Model Averaging}

We study the in-domain test accuracy on PACS and OfficeHome datasets using ImageNet pre-trained ResNet-50 with and without our SMA protocol. In this experiment, we combine all the domains of PACS and split it into training/validation/test splits (0.8/0.1/0.1). We run 10 different runs with different seeds and randomly chosen splits for each dataset. The best model for each run is chosen using the validation set. The remaining optimization details are identical to those used in the previous section. The test accuracy mean and standard error using these best models are shown in Table \ref{tab_iid}. As expected, SMA outperforms models without averaging.

\section{Related Work}
\label{sec_related_work}

\subsection{Model Averaging}
\vspace{-7pt}
\label{sec_relatedw_ma_models}

\textbf{A theoretical perspective}: In our model averaging protocol, we compute a simple moving average of the model parameters starting early during training. This is known as \textit{tail-averaging} \cite{jain2018parallelizing}, which is slightly different from Polyak-Ruppert averaging \cite{polyak1992acceleration} in that the latter starts averaging from the very beginning of training. 
In the context of least square regression in the IID setting, \cite{jain2018parallelizing} theoretically study the behavior of tail averaging and show that the excess risk of the moving average model is upper bounded by a bias and a variance term. This bias term depends on the initialization state of the parameter, but interestingly, it decays exponentially with $t_0$, where $t_0$ is the iteration at which model averaging is started. The variance term on the other hand depends on the covariance of the noise inherent in the data w.r.t. the optimal parameter, and is shown to decay at a faster rate when using model averaging, as opposed to a slower rate without averaging. This motivated them to propose \textit{tail-averaging}.
 
Model averaging has also been shown to have a regularization effect \cite{neu2018iterate} similar to that of Tikhonov regularization \cite{tikhonov1943stability}. This regularization has been classically used in ill-posed optimization problems (typically least squared regression), which are \textit{under-specified}. This property provides an interesting connection between model averaging and the \textit{under-specification} problem discussed in \cite{d2020underspecification}, where the authors perform large scale experiments showing that the performance of multiple over-parameterized deep models, trained independently with different hyper-parameters and seeds, have a high variance on out-domain data, even though their in-domain performances are very close together. Based on this connection, a simple intuition why one can expect model averaging to help in domain generalization is its Tikhonov regularization effect. However, this intuition requires a more thorough investigation.

\textbf{SWAD} \cite{cha2021swad}: SWAD propose flat minima as a means for improving domain generalization. Following the intuition of stochastic weight averaging (SWA, \cite{izmailov2018averaging}), they use model averaging to find flat minima. However, their proposal is different from sampling model states at regular intervals and towards the end of training (as done in SWA). SWAD selects contiguous model states along the optimization path for averaging, based on their validation loss. This is done to prevent including an under-performing state (determined using the in-domain validation set) in the moving average model. SWAD however adds additional hyper-parameters of its own: the validation loss threshold below which the the model states are selected, and patience parameters (number of iterations that determine the start and end of the averaging process). Note that this also requires computing validation loss more frequently during training. In this context, we show that instead of finding the start and end period for model averaging meticulously, we can simply start model averaging early during training and continue till the end. This difference arises from the fact that SWAD uses the online network to calculate validation performance while we use the SMA model in our protocol. This is explained further in section \ref{sec_protocol}. The benefit our observations provide over SWAD is that they allow us to take advantage of model averaging without the additional hyper-parameters and compute required by SWAD.

\subsection{Domain Generalization}
\label{sec_related_work_dg}
\vspace{-8pt}
Existing methods aimed at domain generalization can be broadly categorized into techniques that perform domain alignment, regularization, data augmentation, and meta-learning. Domain alignment is perhaps the most intuitive direction, in which methods aim to learn latent representations which have similar distributions across different domains \cite{sun2016deep,li2018domain,shi2021gradient,rame2021fishr}. There are different variants of this idea, such as minimizing some divergence metric between the latent representation of different domains (E.g. DANN \cite{ganin2016domain}), or less strictly, minimizing the difference between the latent statistics of different domains (E.g. DICA \cite{muandet2013domain}, CORAL \cite{sun2016deep}).
In the meta learning category, source domains are typically split into 2 subsets to be used as the training and test domains in episodes to simulate the domain generalization setting \cite{li2018learning,li2019episodic}. Data augmentation is also a popular tool used for improving domain generalization. It ranges from introducing various types of augmentations to simulate unseen test domain conditions (E.g. style transfer \cite{yue2019domain, zhou2021semi}) to self-supervised learning involving matching the representations of an image with different augmentations (E.g. \cite{albuquerque2020improving,bucci2021self}). Finally, different ways of regularizing models (implicit and explicit) have also been developed with the goal of encouraging domain-invariant feature learning \cite{sagawa2019distributionally,xu2020adversarial,wang2020heterogeneous}. For instance, invariant risk minimization \cite{arjovsky2019invariant} propose a regularization such that the classifier is optimal in all the environments. Representation Self-Challenging \cite{huang2020self} propose to suppress the dominant features that get activated on the training data, which forces the network to use other features that correlate with labels. Risk extrapolation \cite{krueger2021out} propose a regularization that minimizes the variance between domain-wise loss, in the hope that it is representative of the variance including unseen test domains. See \cite{zhou2021domain} for a survey on DG methods. 

Our investigation in this work is complementary to all these domain generalization methods. Additionally, one of our main focus is to also study and improve performance instability on out-domain data during training, which results in more reliable model selection. This aspect has not received much attention. 

\section{Conclusion}
\vspace{-5pt}
We investigated a hyperparameter-free and efficient protocol for model averaging in the ERM framework, and showed that it provides a significant boost to out-domain performance compared to un-averaged models. Building on this observation, we showed that an ensemble of moving average models performs better compared to an ensemble of un-averaged models. Importantly, we showed that in both cases, model averaging significantly improves the rank correlation between in-domain validation accuracy and out-domain test accuracy, which is crucial for reliable model selection using in-domain validation data. We experimented with three pre-trained models with increasing pre-training dataset and model size, and found that EoA provides a proportionally larger gain compared to the corresponding ERM baseline, and lies in the range of $4\%-6\%$. Finally, we explain the performance boost of EoA by adapting the Bias-Variance trade-off perspective to the domain generalization setting. Further discussions along with limitations of our work are provided in Appendix \ref{sec_discussion}.

\bibliography{ref}
\bibliographystyle{plain}

\section*{Checklist}
\begin{enumerate}

\item For all authors...
\begin{enumerate}
  \item Do the main claims made in the abstract and introduction accurately reflect the paper's contributions and scope?
    \answerYes{}
  \item Did you describe the limitations of your work?
    \answerYes{In Appendix \ref{sec_discussion}, we have discussed various aspects including the limitations of our and existing methods in addressing domain generalization, functional diversity of ensembles vs model averaging strategy, and more.}
  \item Did you discuss any potential negative societal impacts of your work?
    \answerYes{See Appendix \ref{sec_broader_impact}.}
  \item Have you read the ethics review guidelines and ensured that your paper conforms to them?
    \answerYes{}
\end{enumerate}

\item If you are including theoretical results...
\begin{enumerate}
  \item Did you state the full set of assumptions of all theoretical results?
    \answerYes{See section \ref{sec_histogram_ablation}.}
        \item Did you include complete proofs of all theoretical results?
    \answerNA{}
\end{enumerate}

\item If you ran experiments...
\begin{enumerate}
  \item Did you include the code, data, and instructions needed to reproduce the main experimental results (either in the supplemental material or as a URL)?
    \answerYes{}
  \item Did you specify all the training details (e.g., data splits, hyperparameters, how they were chosen)?
    \answerYes{}
        \item Did you report error bars (e.g., with respect to the random seed after running experiments multiple times)?
    \answerYes{}
        \item Did you include the total amount of compute and the type of resources used (e.g., type of GPUs, internal cluster, or cloud provider)?
    \answerYes{}
\end{enumerate}

\item If you are using existing assets (e.g., code, data, models) or curating/releasing new assets...
\begin{enumerate}
  \item If your work uses existing assets, did you cite the creators?
    \answerYes{}
  \item Did you mention the license of the assets?
    \answerNo{}
  \item Did you include any new assets either in the supplemental material or as a URL?
    \answerNA{}
  \item Did you discuss whether and how consent was obtained from people whose data you're using/curating?
    \answerNA{}
  \item Did you discuss whether the data you are using/curating contains personally identifiable information or offensive content?
    \answerNA{}
\end{enumerate}

\item If you used crowdsourcing or conducted research with human subjects...
\begin{enumerate}
  \item Did you include the full text of instructions given to participants and screenshots, if applicable?
    \answerNA{}
  \item Did you describe any potential participant risks, with links to Institutional Review Board (IRB) approvals, if applicable?
    \answerNA{}
  \item Did you include the estimated hourly wage paid to participants and the total amount spent on participant compensation?
    \answerNA{}
\end{enumerate}

\end{enumerate}

\newpage
\appendix
\onecolumn
\section*{Appendix}

\begin{table*}[]
\centering
\caption{Hyper-parameter search space for all experiments.}
\begin{tabular}{cccc}
\hline
\textbf{Hyper-parameter}   & \textbf{Default value}           & \multicolumn{2}{c}{\textbf{Random distribution}}   \\ 
&& \cite{gulrajani2020search} & Ours\\
\hline
Learning rate & $5e-5$     & $10^{\text{Uniform}(-5, -3.5)}$ & $5e-5$\\
Batch size & 32 & $2^{\text{Uniform}(3, 5.5)}$ & 32\\
ResNet dropout & 0 & RandomChoice([0, 0.1, 0.5]) & RandomChoice([0, 0.1, 0.5])\\
Weight decay & 0 & $10^{\text{Uniform}(-6, -2)}$ & $10^{\text{Uniform}(-6, -4)}$\\
\hline
\end{tabular}
\label{tab:hyper_param}
\end{table*}

\section{Broader Impact}
\label{sec_broader_impact}
Our work aims at improving out-domain performance of models. Thus it is naturally geared towards mitigating the effects of training dataset biases on the hypothesis learned by the model, which we believe has a positive societal impact.

\section{Training and Evaluation Protocols for DomainBed Benchmarking}
\label{sec_training_protocol}

We use the training protocol described in \cite{gulrajani2020search} with minor changes: we use a smaller hyper-parameter search space (shown in Table \ref{tab:hyper_param}) and smaller number of random trials for computational reasons, and train on DomainNet dataset for $15,000$ iterations instead of $5,000$ similar to \cite{cha2021swad}, because its training loss is quite high. For a dataset with $D$ domains, we run a total of $6D$ random trials. This results in $6$ experiments per domain, in which this domain is used as the test set, while the remaining domains are used as training/validation set (randomly split). This is also the reason why we use a smaller hyper-parameter search space, because otherwise the search space would be under-sampled. For moving average models, the iteration $t_0$ at which averaging is started (Eq. \ref{eq_my_sma}) is set to be $100$ in all experiments unless specified otherwise. For ensembles (both with and without averaged models), the $6$ models corresponding to the $6$ experiments per domain, in which this domain is used as the test set (as described above), are used for ensembling as described in section \ref{sec_eoa}.

All models are trained using the ERM objective and optimized using the Adam optimizer \cite{kingma2014adam}. We use ResNet-50 \cite{he2016deep} pre-trained on Imagenet as our initialization for training in all the experiments. In the final benchmarking experiment, we also use ResNeXt-50 32x4d, that is trained using semi-weakly supervised objective on IG-1B targeted (containing 1 billion weakly labeled images) and ImageNet labeled data \cite{yalniz2019billion}. This model was downloaded from Pytorch hub. For all models, the batch normalization \cite{ioffe2015batch} statistics are kept frozen throughout training and inference. Validation accuracy is calculated every 300 iterations for all datasets except DomainNet where it is calculated every 1000 iterations. Unless specified otherwise, we use the said protocol in all the experiments. For model selection, we use the \textit{training-domain validation set} protocol in \cite{gulrajani2020search} with $80\%-20\%$ training-validation split, and the average out-domain test performance is reported across all runs for each domain. 

All experiments were performed on Google Could Platform (GCP) using 24 NVIDIA A100 GPUs.


\begin{figure*}
    \centering
        \includegraphics[width=0.45\columnwidth]{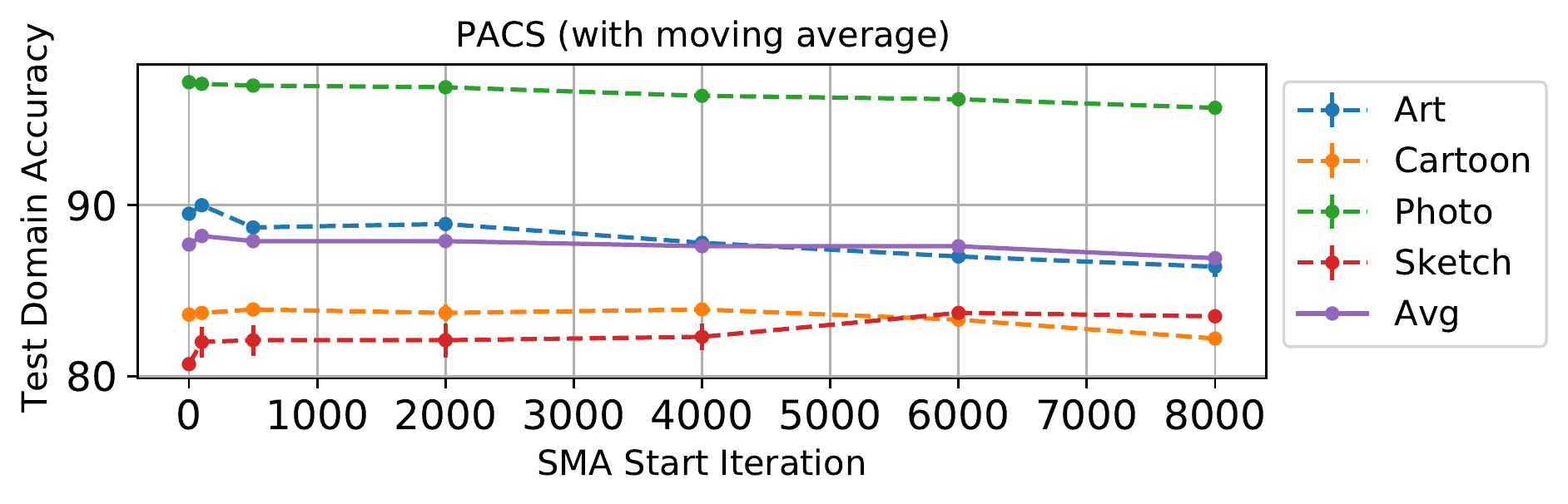}
      \includegraphics[width=0.45\columnwidth]{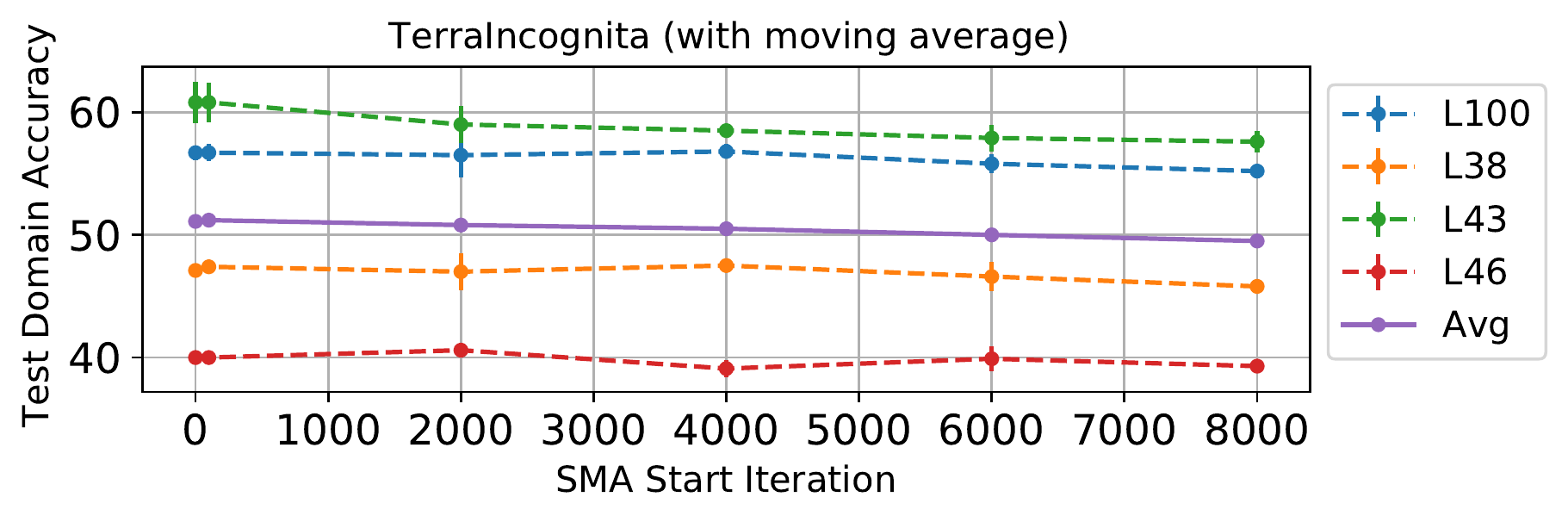}
    \caption{The impact of iteration $t_0$ at which we start simple moving averaging as described in Eq. \ref{eq_my_sma}, on the domain generalization performance for PACS and TerraIncognita datasets. The dominant pattern across all the experiments suggests that starting averaging earlier yields a stronger boost in performance.}
    \label{fig:mo_start_iter}
\end{figure*}

\begin{figure*}
    \centering
        \includegraphics[width=0.45\columnwidth]{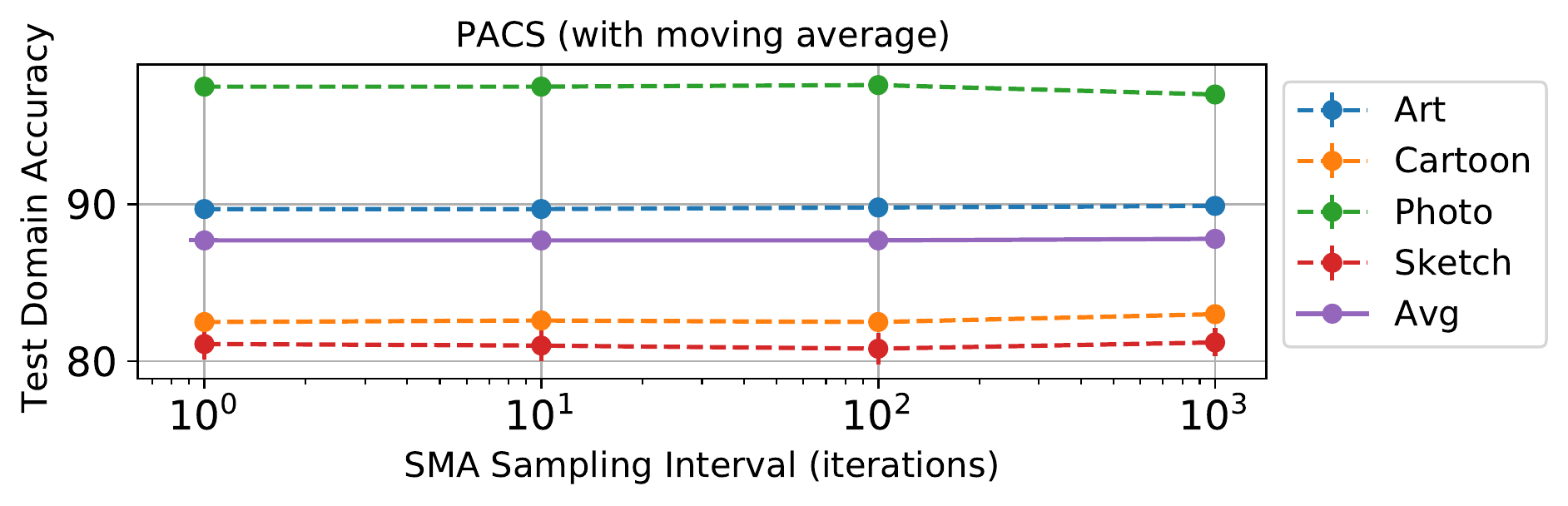}
      \includegraphics[width=0.45\columnwidth]{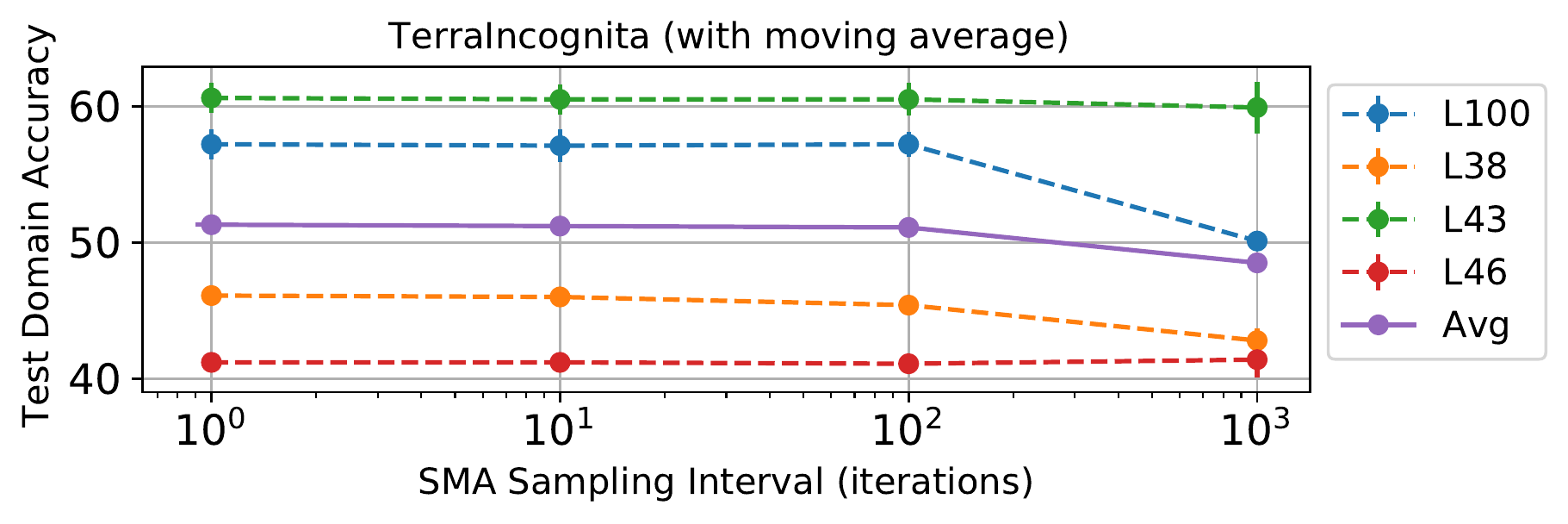}
    \caption{The impact of the frequency (number of iterations), at which model states are sampled for computing the simple moving average (SMA), on the domain generalization performance for PACS and TerraIncognita datasets. Broadly, we find that the frequency of sampling does not have a major influence on performance unless the sampling interval is too large: performance drops significantly on TerraIncognita only when the frequency is set to sampling every 1000 iterations.}
    \label{fig:sma_freq}
\vspace{-11pt}
\end{figure*}

\section{Additional Ablation Analysis of Our Model Averaging Protocol (Eq. \ref{eq_my_sma})}
\label{sec_ablation_sm}

\subsection{Dataset and Experiment Details for SMA Ablation Analysis in Section \ref{sec_ma_analysis}, \ref{sec_start_iter}, and \ref{sec_freq_analysis}}

\textbf{Experimentation Details}: We use the training protocol described in \cite{gulrajani2020search} with minor changes: we use a smaller hyper-parameter search space for feasibility, and train on DomainNet dataset for $15,000$ iterations instead of $5,000$ similar to \cite{cha2021swad}, because its training loss is quite high. Unless specified otherwise, we use the said protocol in all the experiments. For model selection, we use the \textit{training-domain validation set} protocol in \cite{gulrajani2020search}, where the average out-domain test performance is reported across all runs. For more details, see section \ref{sec_training_protocol} in the appendix. 

\textbf{Dataset Details}: We use a subset of the DomainBed benchmark: PACS dataset (4 domains, 7 classes, and 9,991 images), TerraIncognita dataset (4 domains, 10 classes, and 24,788 images), VLCS dataset (4 domains, 5 classes, and 10,729 images), OfficeHome dataset (4 domains, 65 classes, and 15,588 images), and DomainNet dataset (6 domains, 345 classes, and 586,575 images).

\subsection{Start Iteration}
\label{sec_start_iter}

We investigate how domain generalization performance is impacted by the choice of iteration $t_0$ when we start model averaging in Eq. \ref{eq_my_sma}. In this section, we simply refer to it as start iteration, which should not be confused with the start of the training process. 
For experiments, we use the PACS and TerraIncognita datasets. To investigate a wide range of start and end iterations, for all experiments in this section, we train models for $10,000$ iterations. 

We consider starting model averaging from iterations in $\{0, 100, 500, 2000, 4000, 6000, 8000\}$. We plot the performance in Figure \ref{fig:mo_start_iter} for each value. We find that the test performance, averaged across all the domains for both datasets, decreases if we start model averaging later during the training, and choosing $t_0$ close to initialization yields the best performance. We believe using the initialization state in model averaging causes a slight dip in performance because loss is initially high. Based on these experiments, instead of tuning start iteration as a hyper-parameter, \textit{we arbitrarily choose 100 as the start iteration for the remaining experiments in this paper}. This choice of starting averaging later during training is called \textit{tail averaging}, and the theoretical motivation for this choice are discussed in more detail in section \ref{sec_relatedw_ma_models}.

\subsection{Averaging Frequency}
\label{sec_freq_analysis}

When performing simple model averaging described in Eq. \ref{eq_my_sma}, instead of averaging iterates from every iteration, we can alternatively sample iterates at a larger interval. We study the impact of averaging frequency on out-domain test performance. Once again, we use the PACS and TerraIncognita datasets. We train models for $10,000$ iterations, and sample iterates at intervals in $\{10^0, 10^1, 10^2, 10^3\}$ iterations. Test accuracy is once again computed using the protocol of \cite{gulrajani2020search} for each case. The performance as a function of the iterate sampling interval used in SMA is shown in Figure \ref{fig:sma_freq}. Broadly, we find that the frequency of sampling does not have a major impact on performance unless the sampling interval is too large, which happens in the case of TerraIncognita, where performance drops significantly when the sampling interval is set to 1000.

\subsection{Rank Correlation}
\label{sec_rank_corr_intro}
Rank correlation metrics aim to quantify the degree to which an increase in one random variable's value is consistent with an increase in the other random variable's value. Therefore, instead of Pearson's correlation, they are better suited for studying the relationship between the in-domain and out-domain performance for the purpose of model selection because we select the best model during an optimization based on ranking the validation performance (early stopping). We consider Spearman correlation in our experiments. Its value vary between $-1$ and $+1$, where $-1$ implies the ranking of the two random variables are exactly the reverse of each other, and $+1$ implies the ranking of the two random variables are exactly the same as each other. A value of $0$ implies there is no relationship between the two variables.

\subsection{Instability Reduction: Qualitative Analysis}
\label{sec_qualitative_analysis}
Here we try to qualitatively study the robustness of  model selection using in-domain validation set, on out-domain performance. To do so, consider the ideal scenario where the in-domain validation performance correlates well with the out-domain performance. In this case, training longer should not be a problem in general, because if the model starts overfitting beyond a certain point, model selection can take care of it. In such a situation, we would expect the out-domain performance to either improve with longer training, or remain stable.

\begin{figure}[t]
    \centering
      \includegraphics[width=0.45\columnwidth]{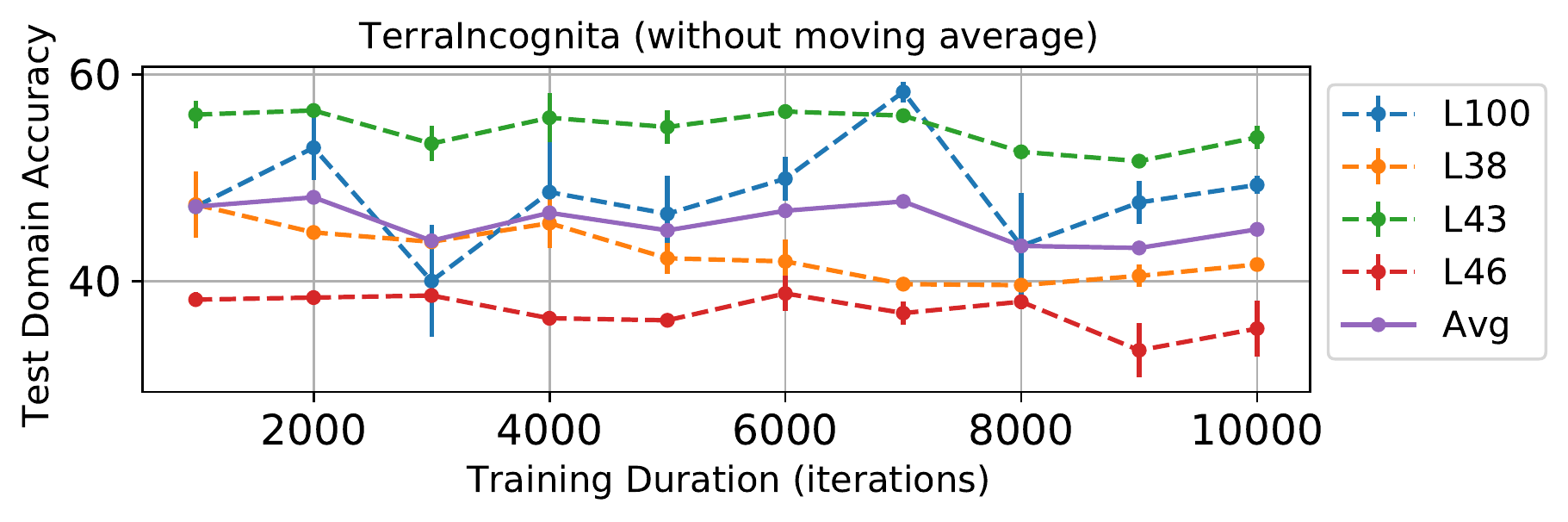}
      \includegraphics[width=0.45\columnwidth]{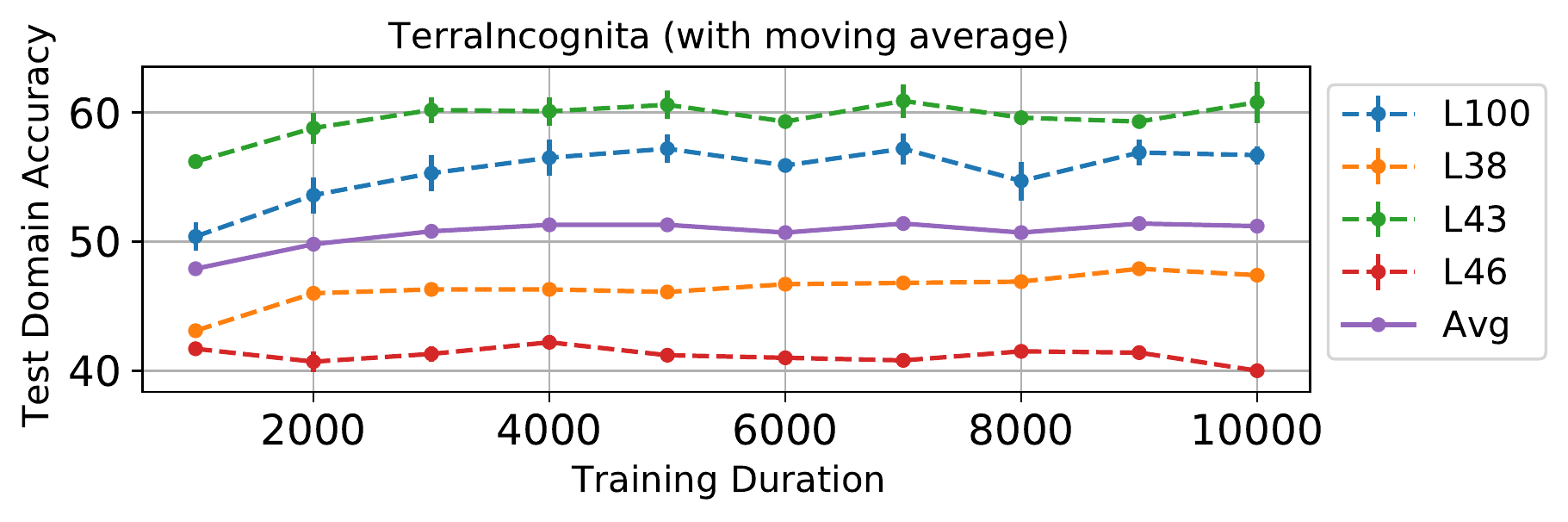}
    \caption{Qualitatively accessing the reliability of model selection while varying the training duration. For each training duration, the out-domain test accuracy is calculated using model selection over the in-domain validation data. Not using model averaging leads to unreliable model selection as evident in the instability of the out-domain performance. Model averaging is able to reduce this instability.}
    \label{fig:end_iter}
\vspace{41pt}
\end{figure}

We use TerraIncognita dataset for this experiment. We consider training duration to be $1,000$ to $10,000$ iterations, at intervals of $1,000$. We plot the performance in Figure \ref{fig:end_iter} for online model (left) and moving average model (right). We find that the performance of moving average models is more stable compared to online models, suggesting that model selection is more reliable when using moving average models. Figure \ref{fig:terra_evolution_all} in the appendix shows the training loss, in-domain validation accuracy and out-domain test accuracy for all the runs used in the above experiment. It shows that the out-domain test performance is unstable during optimization without model averaging, which causes problem for model selection using the in-domain validation performance, as is evident in the above experiment. 

\subsection{Cross-run rank correlation}
\label{sec_cross_run_rank_corr}
In addition to the experiments in \ref{sec_rank_corr}, there is another way in which it makes sense to study the rank correlation between validation and test performance. Suppose we set one of the domains of PACS as our test domain, and the remaining as training/validation data, and perform multiple independent runs with different seeds/hyper-parameter values. At each iteration during training, we can gather the tuple (validation, test) accuracy for each of these runs, and then study the rank correlation between them. The utility of this perspective is to assess the reliability of model selection in terms of selecting a single model across multiple independently trained models, based on their validation performance. We study this rank correlation for PACS and TerraIncognita datasets. The results are shown in Figure \ref{fig:cross_run_spearman} in the appendix. We find that the cross-run rank correlations are poor (not consistently close to 1) for both online model (without averaging) and moving average model. This implies that in-domain validation performance based model selection is not a reliable approach for selecting a model from a pool of multiple independently trained models.

\begin{table}[t]
\caption{Spearman correlation (closer to 1 is better) between within-run in-domain validation accuracy and out-domain test accuracy on the multiple datasets in the DomainBed benchmark for individual models (left) and ensembles (right). In most cases, using model averaging results in a significantly better rank correlation, which makes model selection more reliable.}

\begin{minipage}{.5\linewidth}
      \caption{Individual Models}
      \centering
        \begin{tabular}{lll}
        \hhline{===}
        PACS    & w/o avg & with avg   \\ \hline
        Art     & 0.31 $\pm$ 0.04   & \textbf{0.62 $\pm$ 0.04}   \\
        Cartoon & 0.25 $\pm$ 0.10   & \textbf{0.52 $\pm$ 0.03}  \\
        Photo   & \textbf{0.09 $\pm$ 0.07}   & -0.38 $\pm$ 0.15 \\
        Sketch  & 0.24 $\pm$ 0.06   & \textbf{0.53 $\pm$ 0.06}  \\ \hhline{===}
        TerraIncognita    & w/o avg & with avg   \\ \hline
        L100     & 0.21 $\pm$ 0.07   & \textbf{0.90 $\pm$ 0.05}   \\
        L38 & 0.12 $\pm$ 0.13   & \textbf{0.83 $\pm$ 0.05}  \\
        L43   & 0.30 $\pm$ 0.06   & \textbf{0.67 $\pm$ 0.18} \\
        L46  & 0.03 $\pm$ 0.11   & \textbf{0.52 $\pm$ 0.14}  \\ 
        \hhline{===}
        VLCS    & w/o avg & with avg   \\ \hline
        Caltech101     & \textbf{0.21 $\pm$ 0.10}   & 0.16 $\pm$ 0.15   \\
        LabelMe & \textbf{0.30 $\pm$ 0.08}   & 0.02 $\pm$ 0.14  \\
        Sun09   & 0.27 $\pm$ 0.12   & \textbf{0.32 $\pm$ 0.11} \\
        VOC2007  & 0.17 $\pm$ 0.11   & \textbf{0.38 $\pm$ 0.05}  \\ \hhline{===}
        OfficeHome    & w/o avg & with avg   \\ \hline
        Art     & 0.05 $\pm$ 0.11   & \textbf{0.80 $\pm$ 0.04}   \\
        Clipart & 0.33 $\pm$ 0.04   & \textbf{0.84 $\pm$ 0.04} \\
        Product   & 0.61 $\pm$ 0.04   & \textbf{0.80 $\pm$ 0.04} \\
        RealWorld  & 0.41 $\pm$ 0.06   & \textbf{0.74 $\pm$ 0.04}  \\ \hhline{===}
        DomainNet    & w/o avg & with avg   \\ \hline
        Clip     & 0.96 $\pm$ 0.01   & \textbf{1 $\pm$ 0}   \\
        Info & 0.80 $\pm$ 0.05   & \textbf{1 $\pm$ 0}  \\
        Paint   & 0.87 $\pm$ 0.02   & \textbf{1 $\pm$ 0} \\
        Quick  & 0.65 $\pm$ 0.04   & \textbf{1 $\pm$ 0}  \\ 
        Real   & 0.91 $\pm$ 0.01   & \textbf{1 $\pm$ 0} \\
        Sketch  & 0.82 $\pm$ 0.04   & \textbf{1 $\pm$ 0}  \\ 
        \hline
        \end{tabular}
        \label{tab:rank_correlation}
    \end{minipage}%
    \begin{minipage}{.5\linewidth}
      \centering
        \caption{Ensembles}
        \begin{tabular}{lll}
        \hhline{===}
        PACS    & w/o avg & with avg   \\ \hline
        Art     & 0.06   & \textbf{0.78}   \\
        Cartoon & 0.33   & \textbf{0.81}  \\
        Photo   & \textbf{-0.12}   & -0.52 \\
        Sketch  & 0.43   & \textbf{0.70}  \\ \hhline{===}
        TerraIncognita    & w/o avg & with avg   \\ \hline
        L100     & 0.48   & \textbf{1}   \\
        L38 & 0.17   & \textbf{0.95}  \\
        L43   & \textbf{0.59}   & 0.38 \\
        L46  & 0.08   & \textbf{0.61}  \\ 
        \hhline{===}
        VLCS    & w/o avg & with avg   \\ \hline
        Caltech101     & 0.52   & \textbf{0.81}   \\
        LabelMe & 0.05   &\textbf{ 0.38}  \\
        Sun09   & 0.63   & \textbf{0.82} \\
        VOC2007  & 0.55   & \textbf{0.65}  \\ \hhline{===}
        OfficeHome    & w/o avg & with avg   \\ \hline
        Art     & 0.27   & \textbf{0.92}   \\
        Clipart & 0.66   & \textbf{0.95} \\
        Product   & 0.20   & \textbf{0.95} \\
        RealWorld  & 0.09   & \textbf{0.78}  \\ \hhline{===}
        DomainNet    & w/o avg & with avg   \\ \hline
        Clip     & \textbf{1}   & \textbf{1}   \\
        Info & 0.88   & \textbf{1 }  \\
        Paint   & 0.98   & \textbf{1 } \\
        Quick  & 0.95   & \textbf{1 }  \\ 
        Real   & 0.97   & \textbf{1 } \\
        Sketch  & 0.97   & \textbf{1 }  \\ 
        \hline
        \end{tabular}
        \label{tab:rank_correlation_ensemble}
    \end{minipage} 
\end{table}

\begin{figure}[t]
    \centering
        \includegraphics[width=0.23\columnwidth]{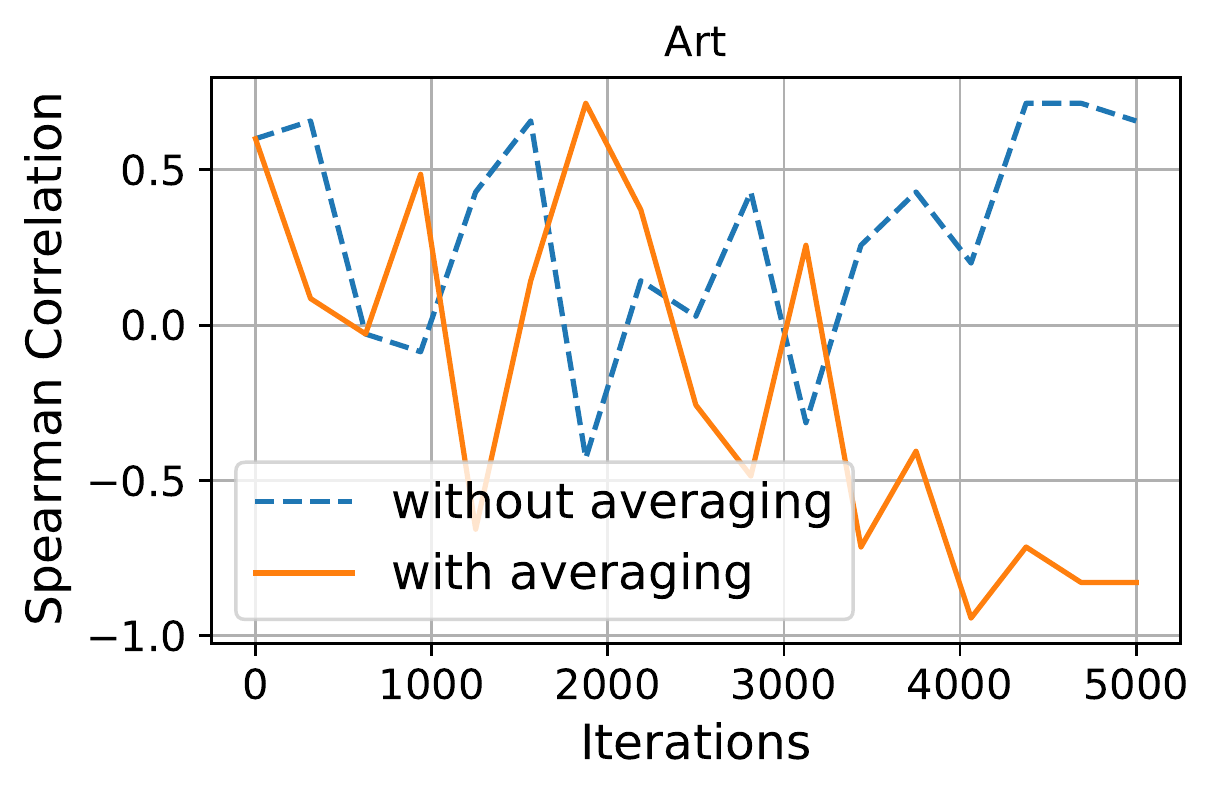}
       \includegraphics[width=0.23\columnwidth]{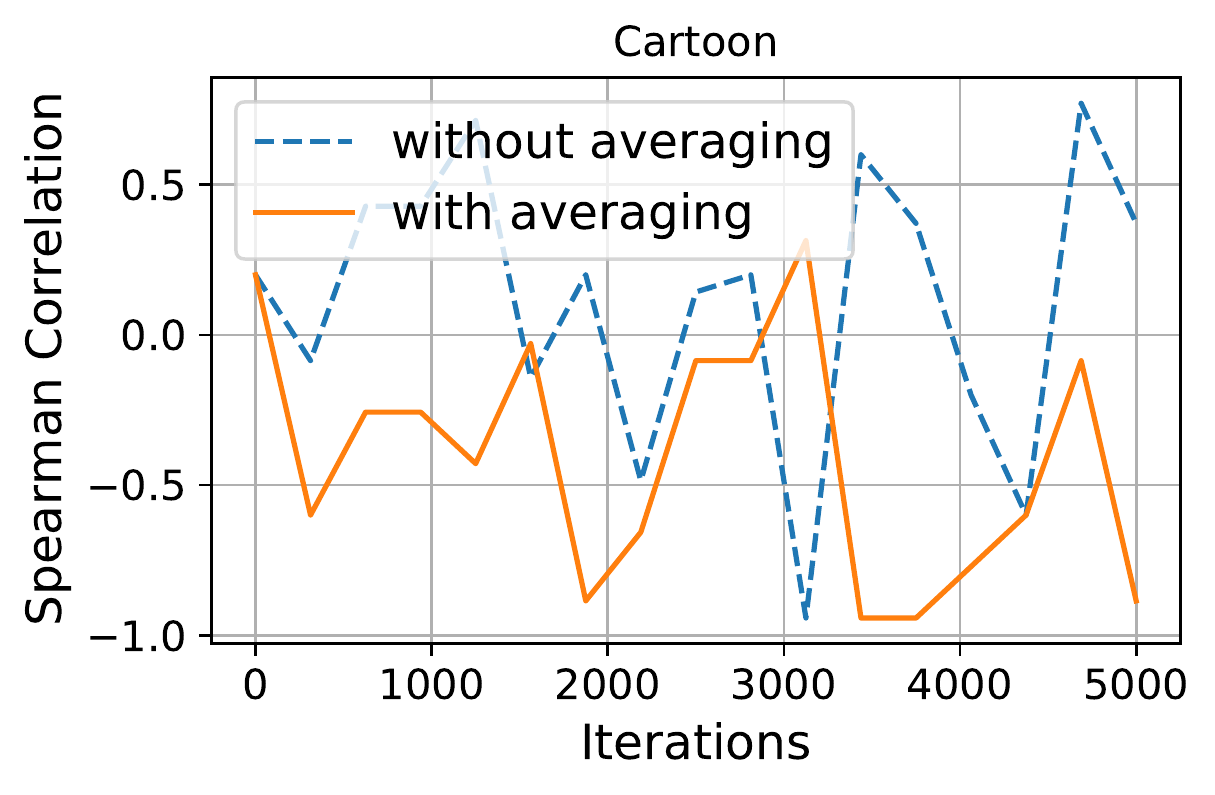}
       \includegraphics[width=0.23\columnwidth]{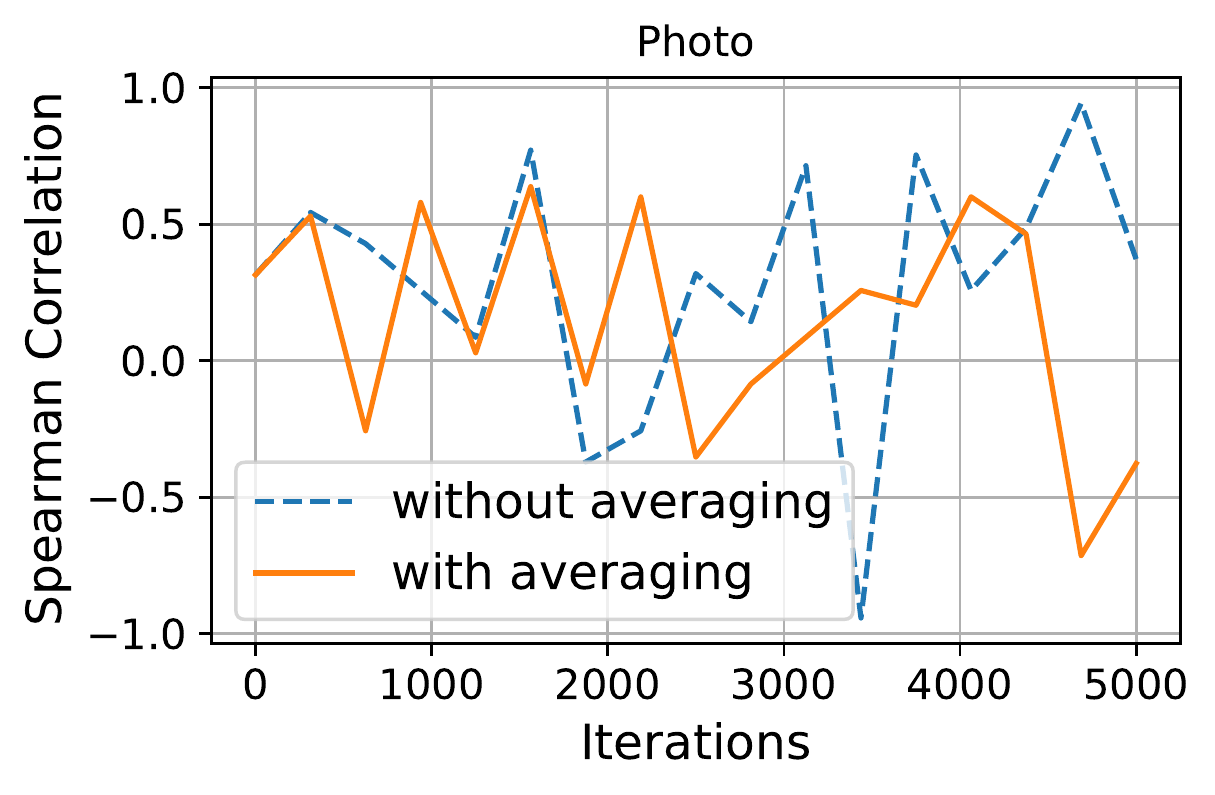}
       \includegraphics[width=0.23\columnwidth]{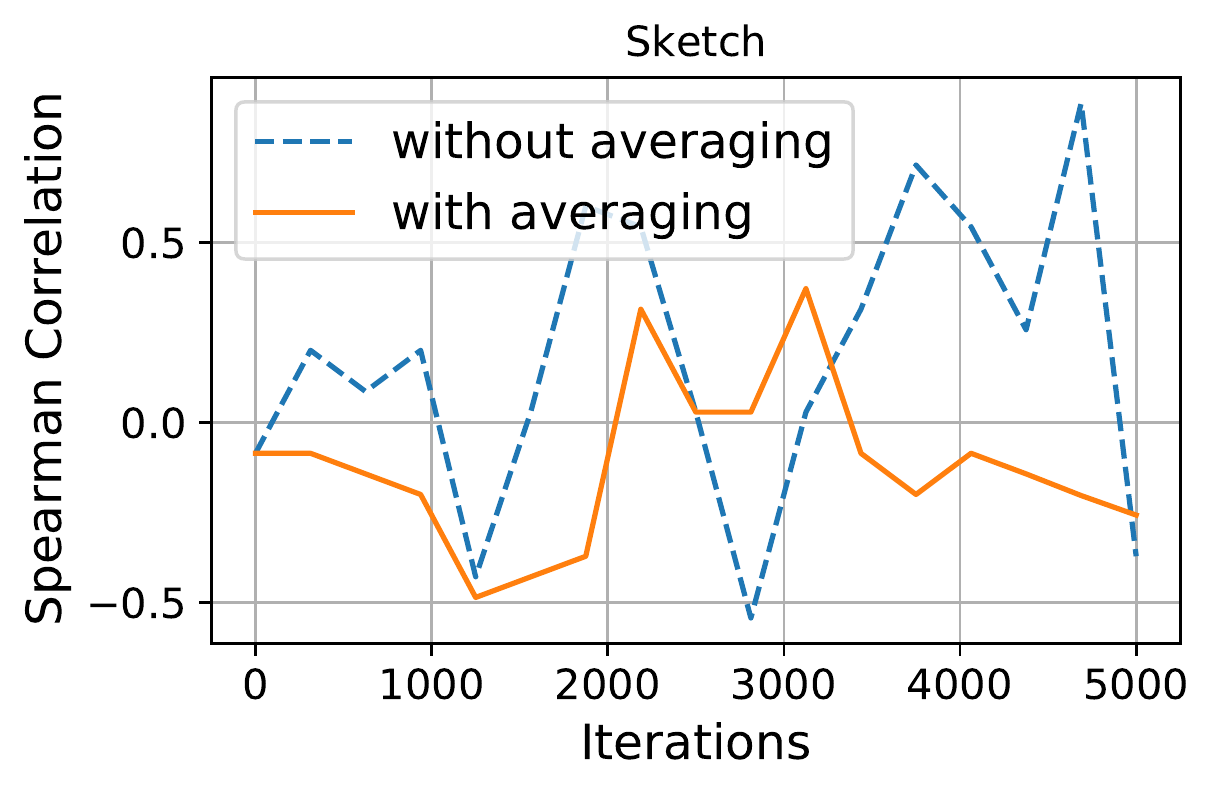}
       \includegraphics[width=0.23\columnwidth]{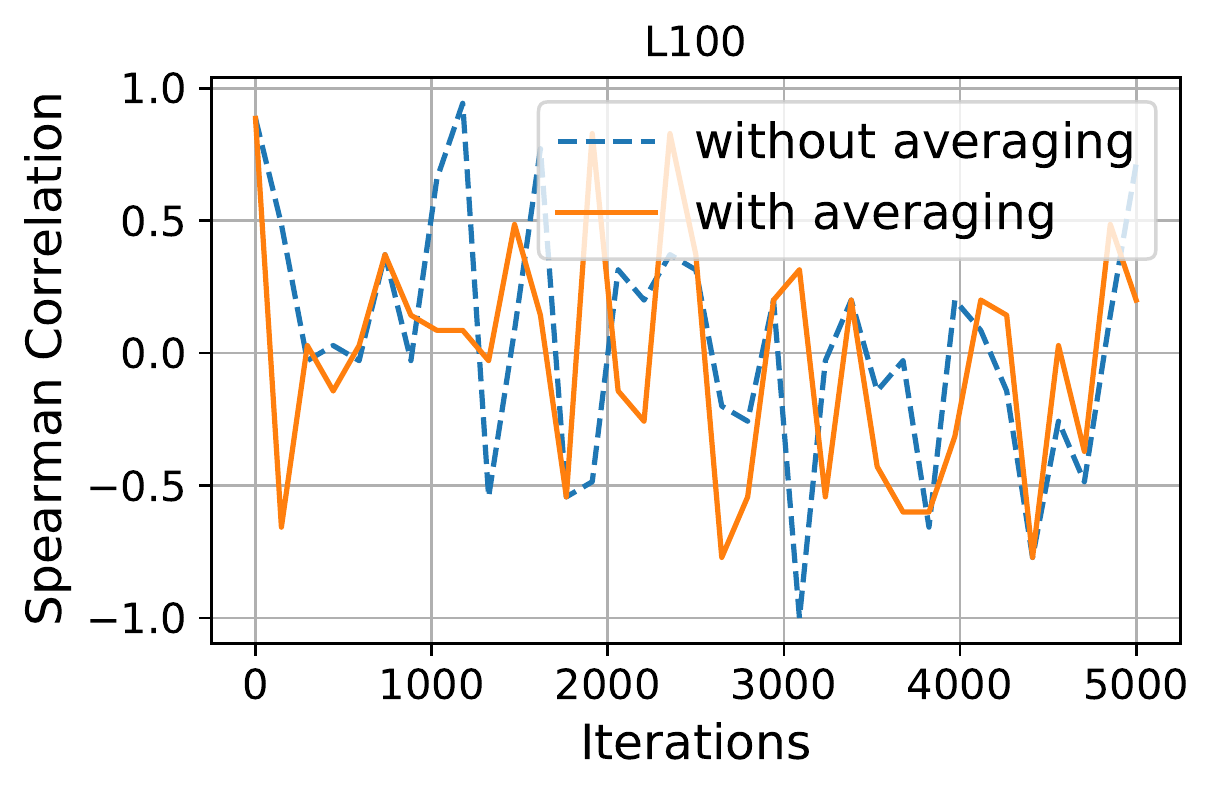}
       \includegraphics[width=0.23\columnwidth]{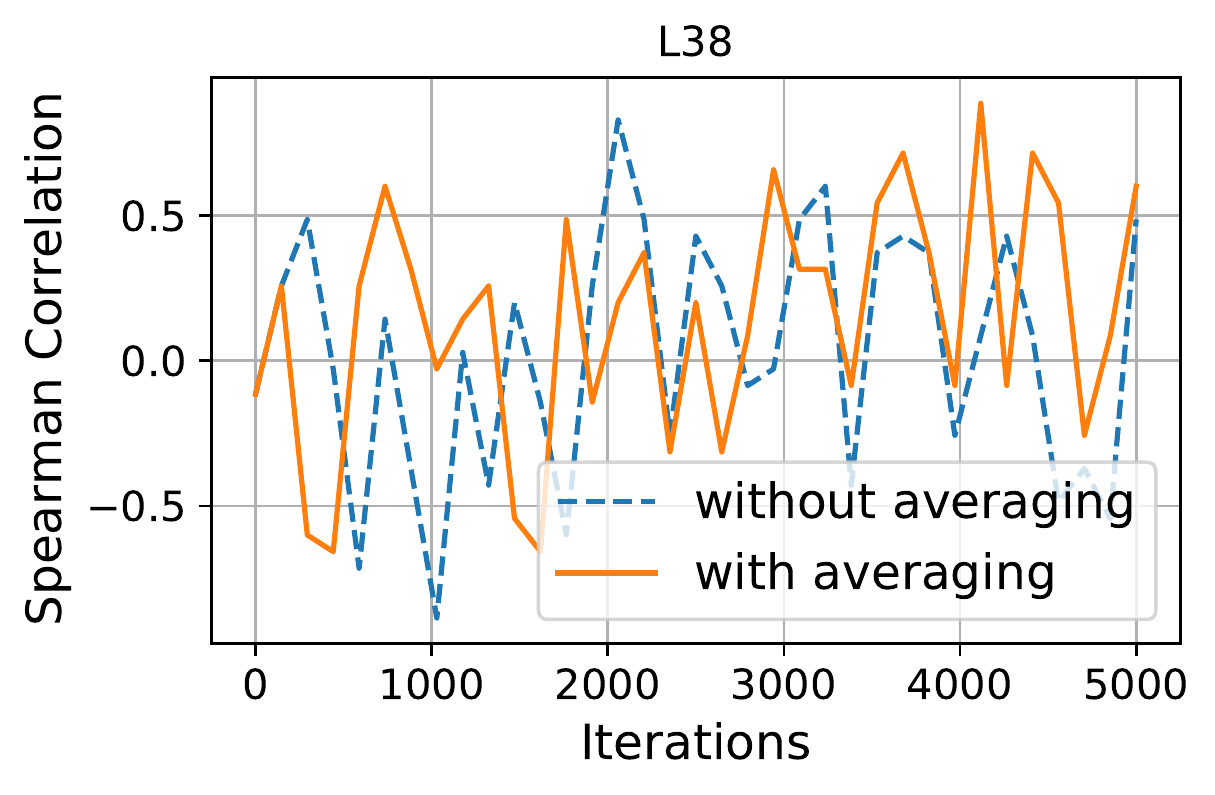}
       \includegraphics[width=0.23\columnwidth]{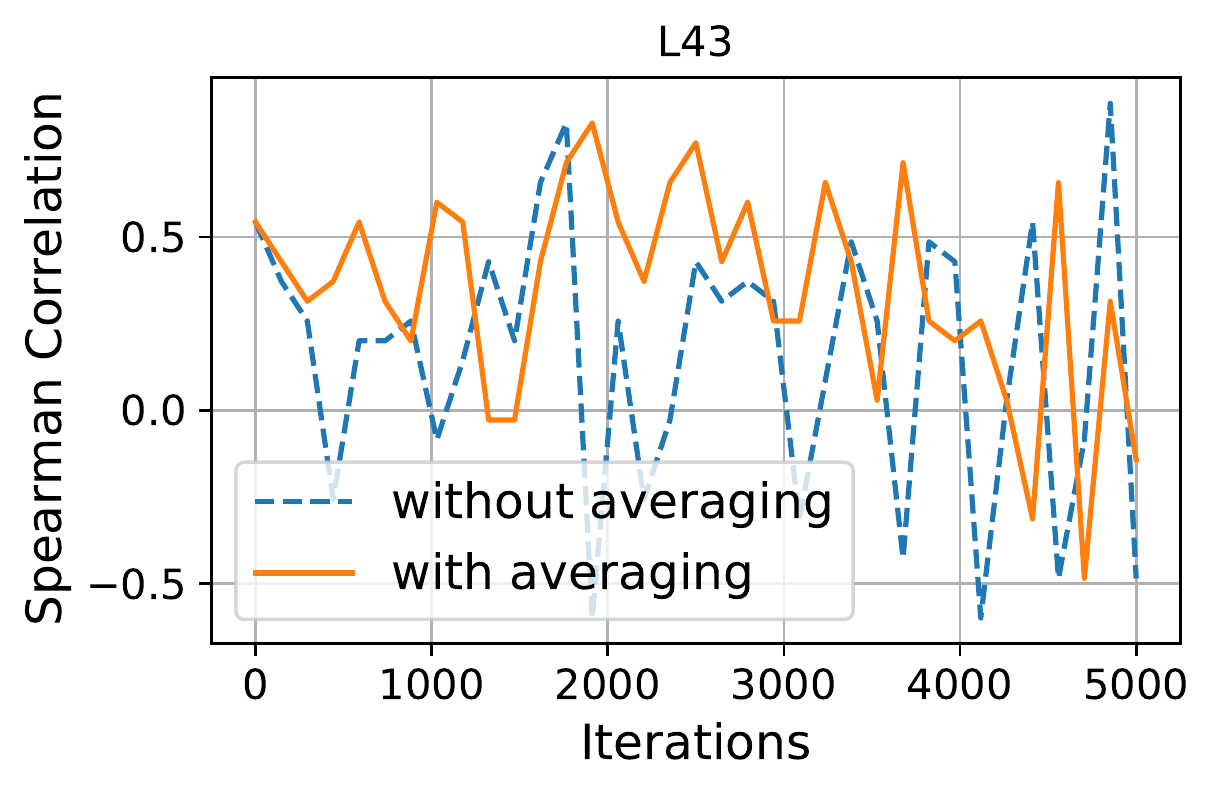}
       \includegraphics[width=0.23\columnwidth]{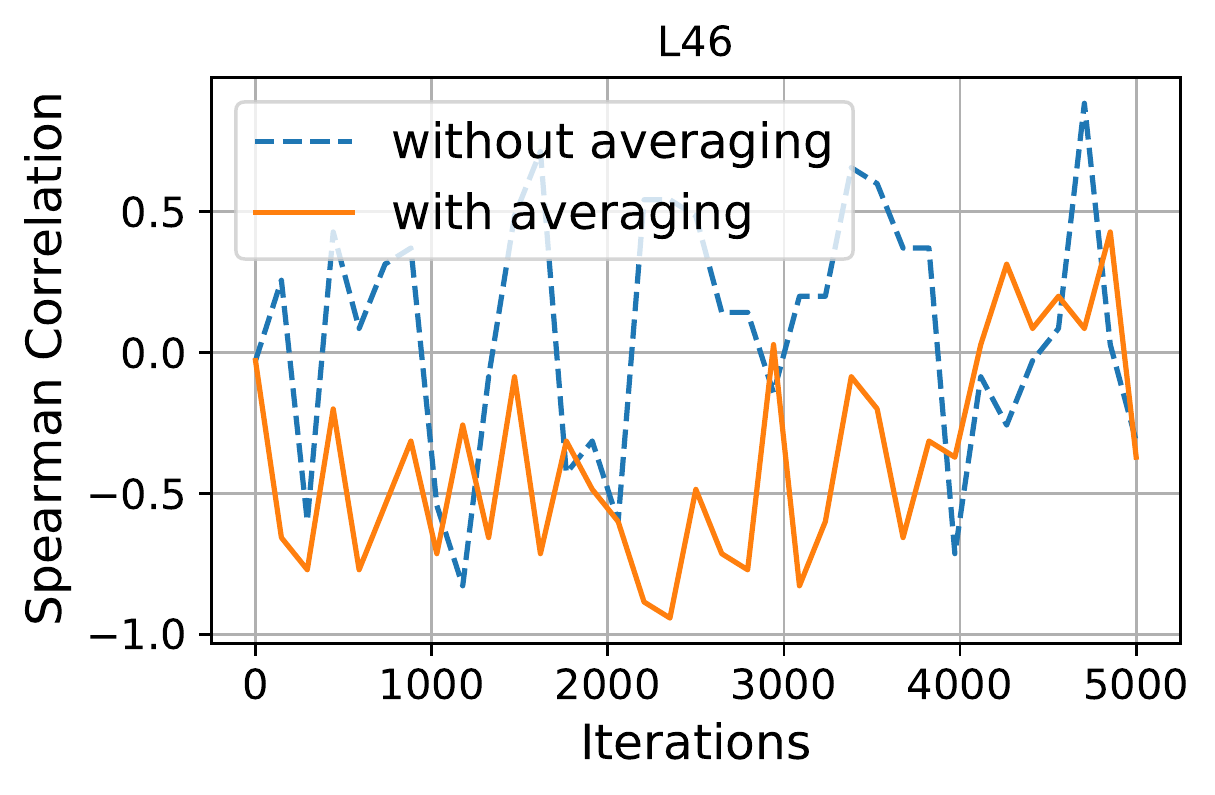}
    \caption{Spearman correlation between \textit{cross-run} in-domain validation accuracy and out-domain test accuracy for PACS dataset (top) and TerraIncognita dataset (bottom). The cross-run rank correlations are poor (not consistently close to 1) for both online model (without avg) and moving average model. This implies that in-domain validation performance based model selection is not a reliable approach for selecting a single model from a pool of multiple independently trained models. See section \ref{sec_rank_corr} for details.}
    \label{fig:cross_run_spearman}
\end{figure}

\section{Why Does Ensembling Improve Performance?}
\label{sec_ablation_details}

\begin{figure*}
    \centering
        \includegraphics[width=0.45\columnwidth,trim={0cm 0cm 0cm 0},clip]{figs/histogram_ablation/ablation_histogram_per_dim-0_terra.pdf}
        \includegraphics[width=0.45\columnwidth,trim={0cm 0cm 0cm 0},clip]{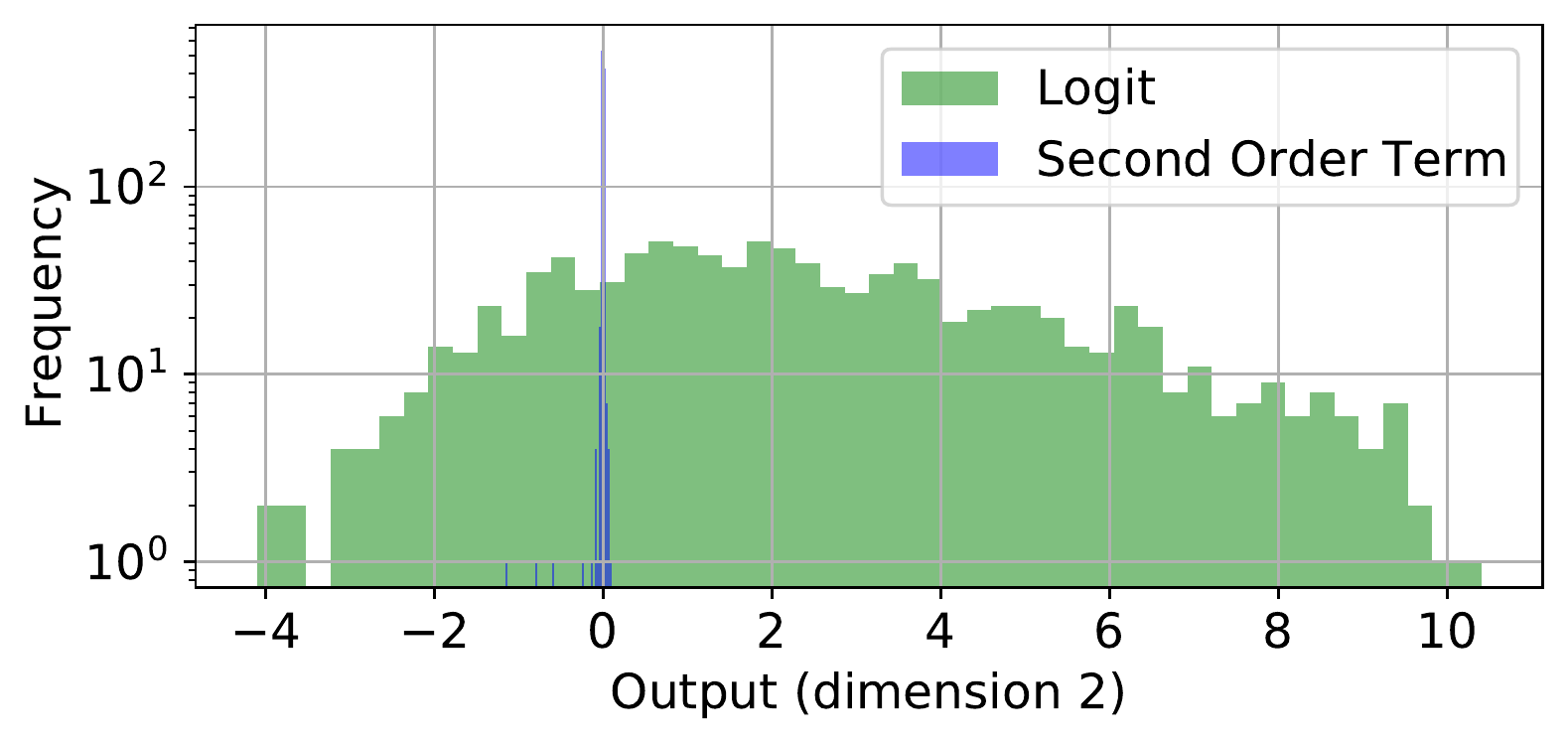}
        \vfil
        \includegraphics[width=0.45\columnwidth,trim={0cm 0cm 0cm 0},clip]{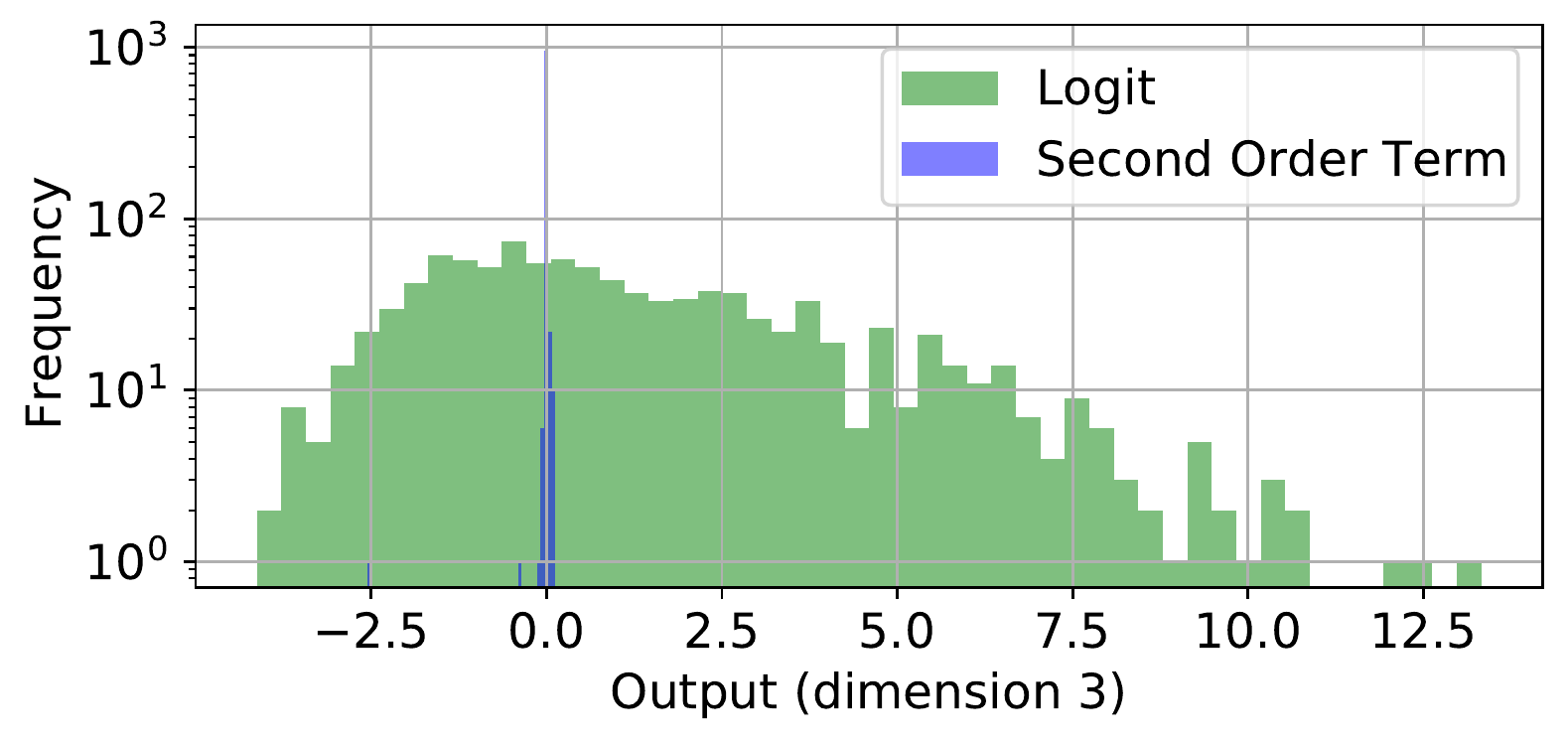}
        \includegraphics[width=0.45\columnwidth,trim={0cm 0cm 0cm 0},clip]{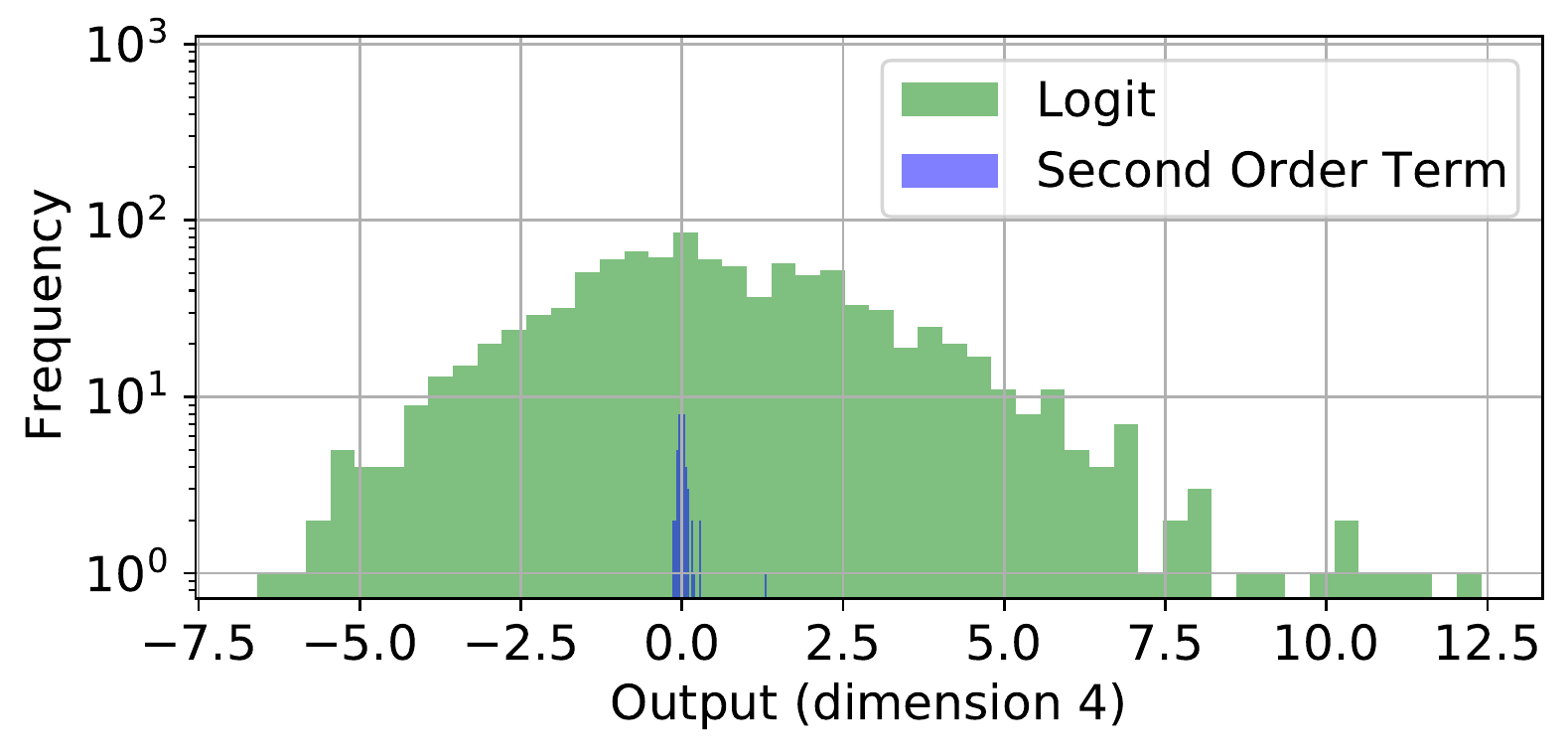}
        \vfil
        \includegraphics[width=0.45\columnwidth,trim={0cm 0cm 0cm 0},clip]{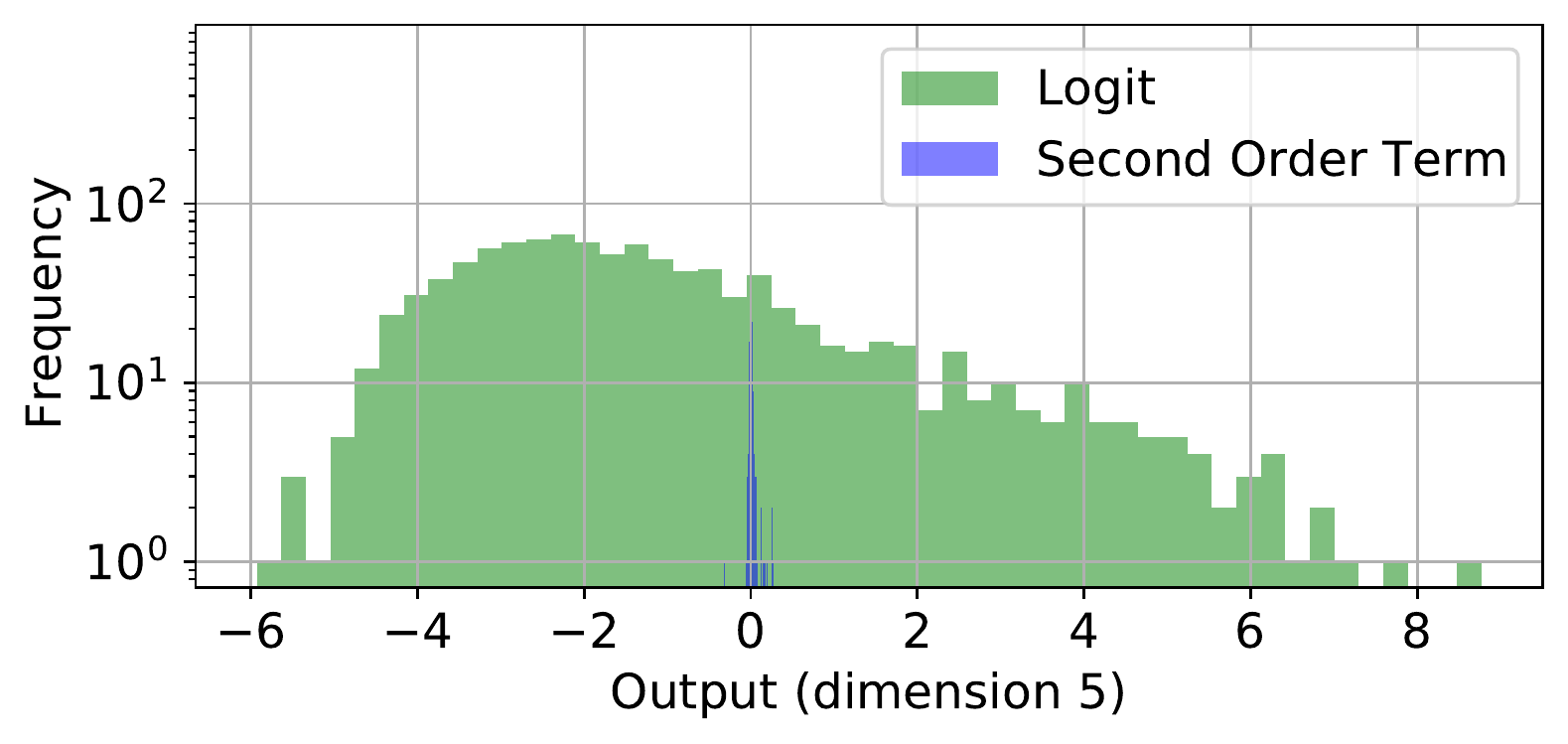}
        \includegraphics[width=0.45\columnwidth,trim={0cm 0cm 0cm 0},clip]{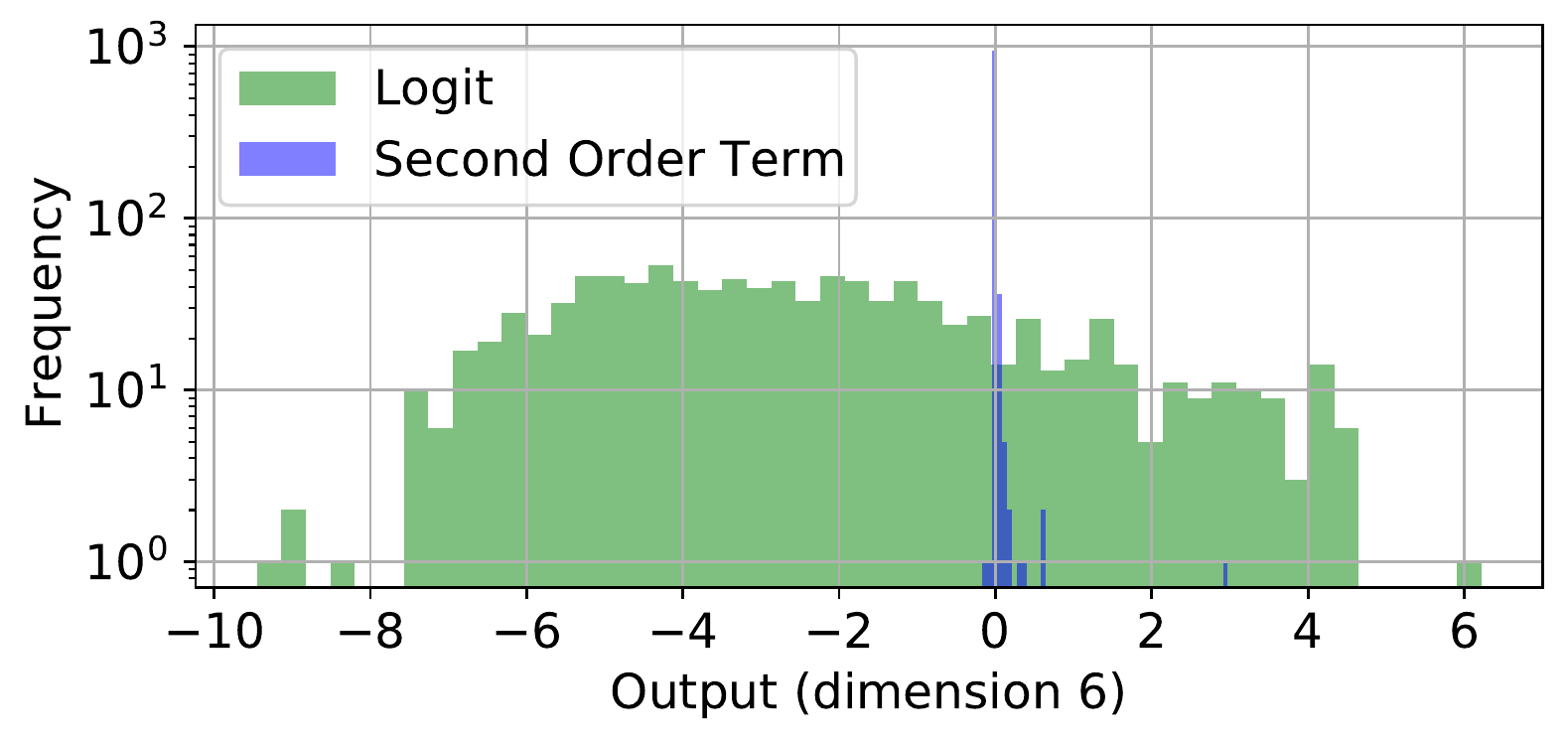}
        \vfil
        \includegraphics[width=0.45\columnwidth,trim={0cm 0cm 0cm 0},clip]{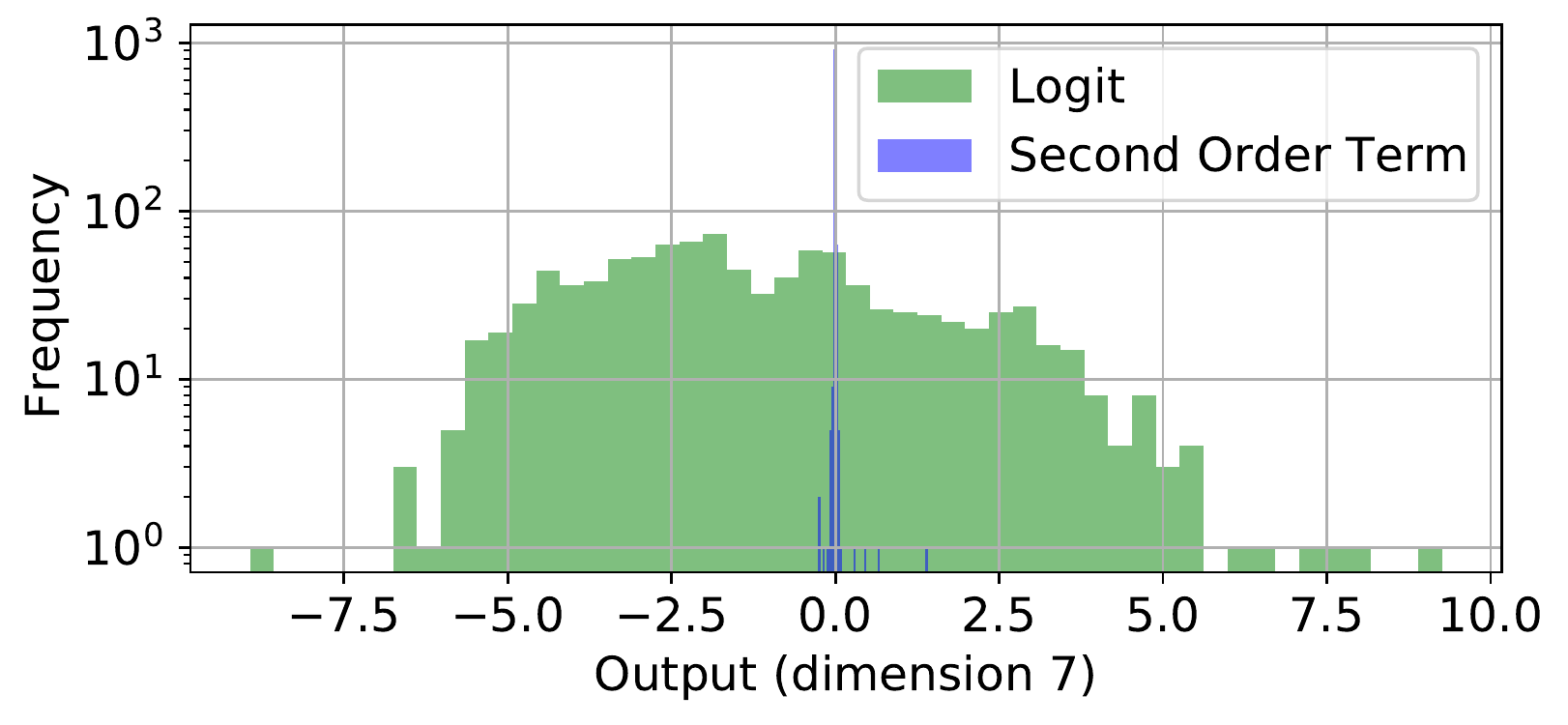}
        \includegraphics[width=0.45\columnwidth,trim={0cm 0cm 0cm 0},clip]{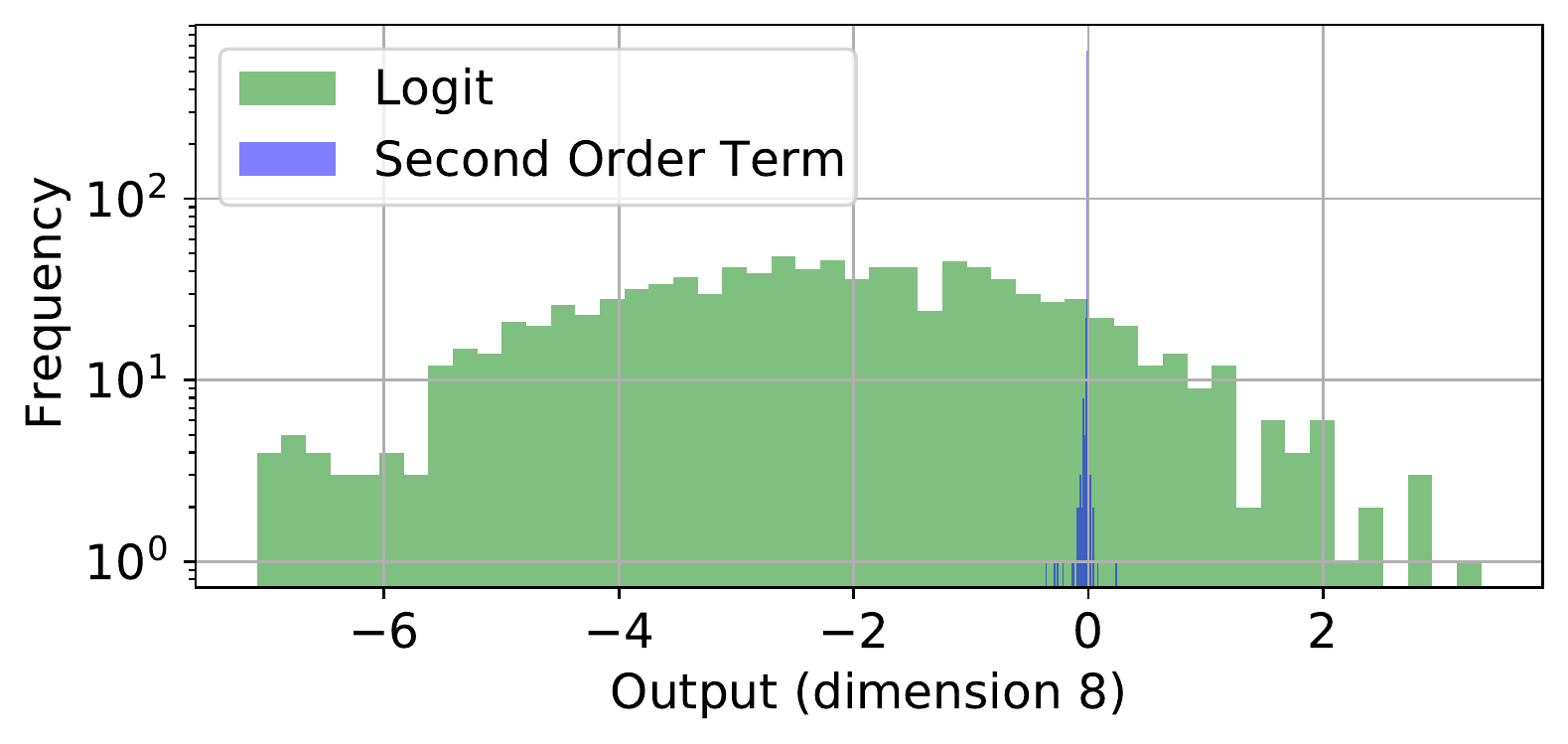}
        \vfil
        \includegraphics[width=0.45\columnwidth,trim={0cm 0cm 0cm 0},clip]{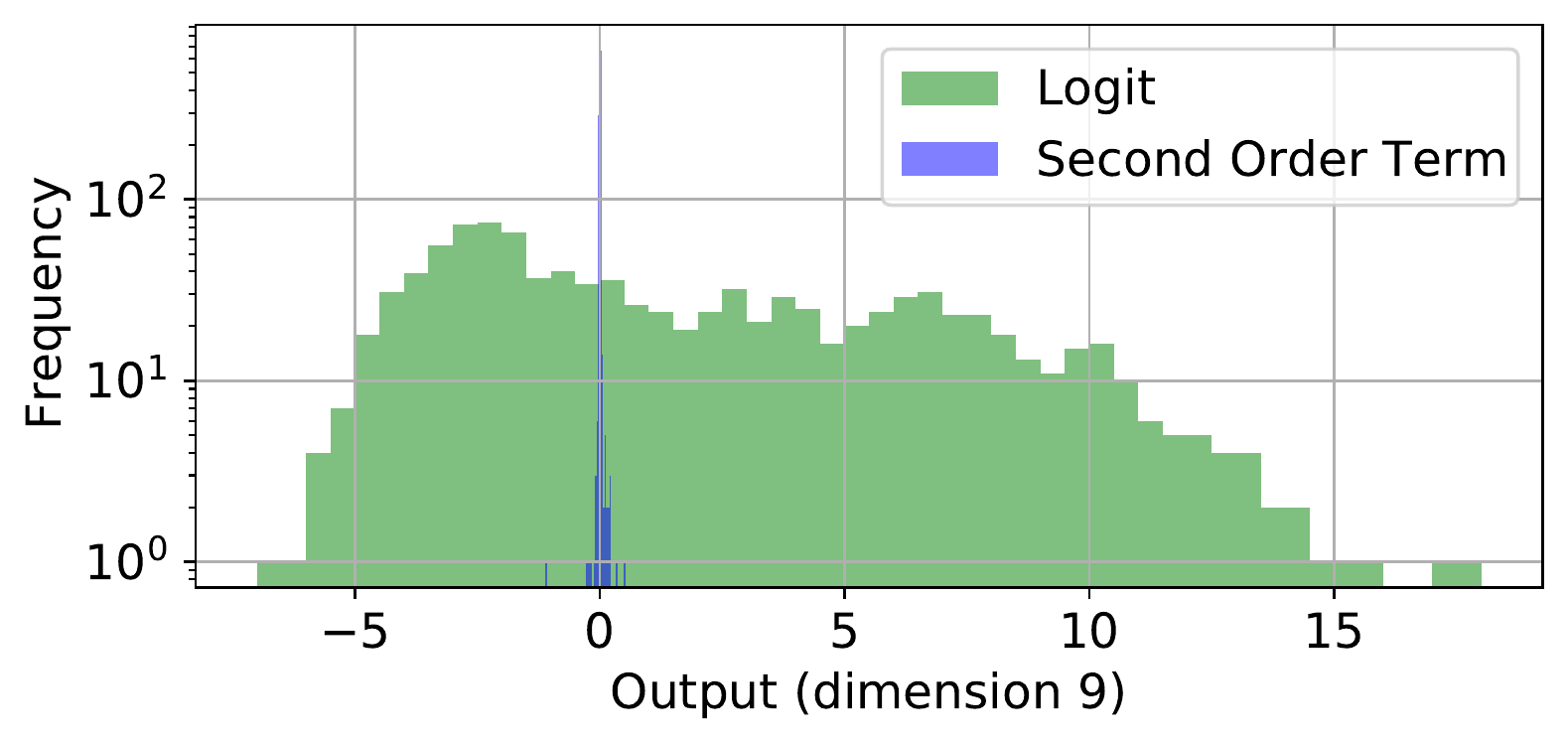}
        \includegraphics[width=0.45\columnwidth,trim={0cm 0cm 0cm 0},clip]{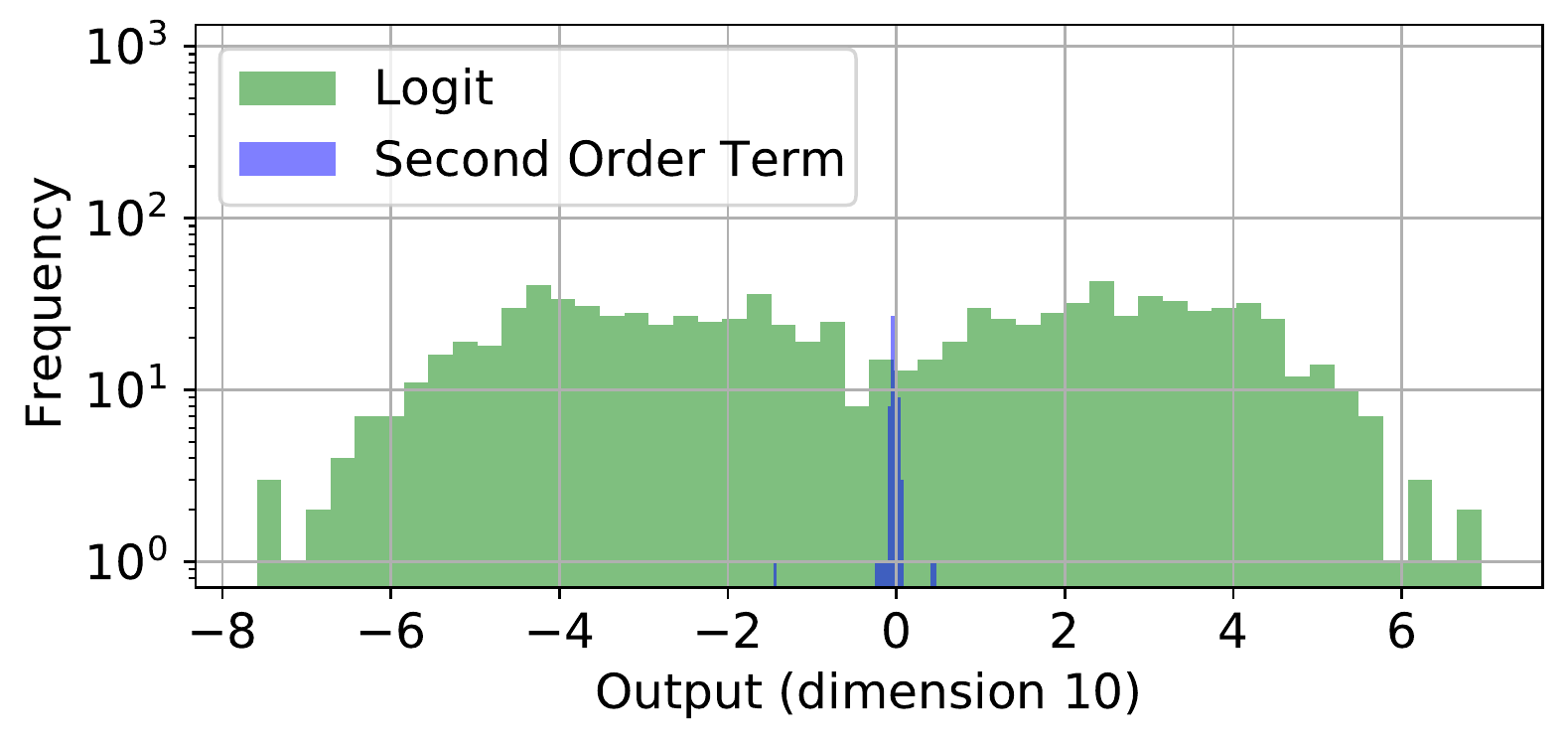}
    \caption{The scale of terms-- moving average model's logit and the second order term in Eq. \ref{eq_taylor_expansion}. The latter concentrates around 0, suggesting our model averaging protocol approximates ensembles.}
    \label{fig:ablation_dim_all}
\vspace{-15pt}
\end{figure*}

We describe the histogram experiment presented in section \ref{sec_histogram_ablation}. For this experiment, we use the TerraIncognita dataset with L46 as the test domain. We use one of the runs with model averaging from the experiments done in section \ref{sec_benchmarking}. We begin by re-writing the Taylor's expansion more precisely for our model averaging protocol,
\begin{align}
    \frac{1}{T-t_0+1}\cdot \sum_{t=t_0}^{T} f(\mathbf{x}; {\theta_t})_k &\approx f(\mathbf{x}; \hat{\theta}_{T})_k + 0.5 \cdot \frac{1}{T-t_0+1}\cdot \sum_{t=t_0}^{T} (\hat{\theta}_{T} - {\theta}_t)^T  \frac{\partial^2 f(\mathbf{x}; {\hat{\theta}_{T}})_k }{\partial \hat{\theta}_{T}^2} (\hat{\theta}_{T} - {\theta}_t)
\end{align}
where $\hat{\theta}_T := \frac{1}{T-t_0+1}\cdot \sum_{t=t_0}^{T} {\theta}_t$. Notice the first order term has been omitted since it is zero. As an important detail, instead of computing the second order term in the above equation for all integer valued $t$ between $t_0$ and $T$, we only use $\theta_t$ for $t = 300*i$ for $i \in \{1,2,\hdots, 16\}$ due to computational constraints. This should not affect our conclusion because as shown in section \ref{sec_freq_analysis}, averaging frequency does not have a significant impact on performance. Finally, we record the logit values and the second order term for 1000 randomly selected samples $\mathbf{x}$ from the test domain data. Note that TerraIncognita has 10 classes and so we have the above equation for $k \in \{1,2, \hdots, 10\}$. We plot the histogram of the logit values and the second order term corresponding to each class's output separately. All the plots are shown in figure \ref{fig:ablation_dim_all}. The conclusion is the same as described in the main text.

\section{Discussions and Limitations}
\label{sec_discussion}

\textbf{Domain Generalization Limitations}: The Bias-Variance trade-off analysis of EoA in section \ref{sec_histogram_ablation} shows that the expected loss of individual models constitutes both the bias and variance term, while that of ensembles is dominated by the bias term alone, and is thus strictly lower. Thus ensembles (either explicitly or implicitly by model averaging) is guaranteed to improve performance on OOD data compared to the corresponding individual unaveraged models. However, this strategy cannot go beyond getting rid of the variance term, and other strategies need to be used to reduce the bias term, which will further improve the OOD performance. 

Additionally, our proposal does not make use of the environment ID of samples. A popular strategy to utilize this information is to increase some form of alignment between the latent representations of different domains, which has been shown to be one of the terms in the upper bound of the test domain generalization error \cite{ben2006analysis}. While a number of prior work has proposed variants of this strategy (eg. CORAL, DANN, see section \ref{sec_related_work_dg} for a discussion), \cite{gulrajani2020search} has showed that ERM with appropriate training-validation protocol performs at least as well as these methods. Further, \cite{vedantam2021empirical} has recently argued supported by empirical evidence that domain alignment is neither necessary nor sufficient for domain generalization. Thus it remains an open question what other strategies can be used to utilize the environment ID to boost domain generalization

\textbf{Functional Diversity}: Model averaging mitigates instability within a run, which makes model selection more reliable. However, we note that the gap in performance between different runs still exists, though it is smaller on average compared with online models (see training evolution plots in appendix for reference). On another note, while our analysis in section \ref{sec_histogram_ablation} suggests that model averaging behaves as an ensemble, we believe that it does not offer as much of functional diversity as independently trained models. This is because if it did, model averaging should have had a much better performance compared with a traditional ensembles (since there are $T-t_0+1$ models in model averaging while only 6 models in the traditional ensemble in our experiments), but this is not the case (see their performance in table \ref{tab:benchmark_full} for a comparison). This implies that there is still diversity among the independently trained models with model averaging. This is also inline with \cite{fort2019deep} which shows that ensembling methods such as model averaging and Monte Carlo dropout \cite{gal2016dropout} do not provide diversity in function space as much as ensembles of independently trained models. Perhaps this is why EoA performs better compared to both individual moving average models and traditional ensembles, by better approximating the expected ensemble behavior. 

\textbf{Scalability}: Following the protocol of \cite{gulrajani2020search}, we used samples from all the training domains in each mini-batch update. However, in settings where the number of domains is very large, this approach can be prohibitive. As an alternative, we also performed preliminary experiments in which we stochastically picked one of the training domains at every iteration, and sampled a mini-batch from that domain to update parameters. We found that this protocol resulted in a similar performance as that achieved by the protocol used in our work.

\textbf{Computational Complexity}: The computational overhead due to SMA is practically negligible (compared to back-propagation) since it merely involves a running average estimate of the parameters. So its complexity is the same as that of training a vanilla supervised deep network. On the other hand, since EoA trains an ensemble of models, the complexity scales linearly with the number of models used in the ensemble compared with SMA on a single model, if these models are trained sequentially. Of course, these different models can be trained in parallel if resources are available, given each model is trained independently of one another, in which case the complexity of EoA remains the same as that of vanilla supervised training.

\textbf{Information leaking considerations}: We present many experiments where validation and test performances are studied. It is therefore natural to wonder if there was any information leak from the test set while performing this analysis. We note that in the model averaging protocol we investigated, there were two moving parts: iteration $t_0$ at which model averaging is started, and averaging frequency. We studied them in section \ref{sec_start_iter} and \ref{sec_freq_analysis} respectively. For both of them, we proposed to fix their values universally instead of tuning them on each dataset. Specifically, \cite{jain2018parallelizing} propose tail averaging in which iterates from every iteration are used for computing the simple moving average. We found this proposal to work well empirically in our analysis, and therefore set averaging frequency to 1. For start iteration $t_0$ on the other hand, \cite{jain2018parallelizing} theoretically show that the initial bias term in the excess error upper bound decays exponentially with $t_0$. In line with this theory, our analysis showed that an iteration close to but not at initialization worked well. So we arbitrarily set $t_0=100$. Note that these are not their optimal values, but are rather arbitrary choices guided by our investigation and existing theory. Aside from these two objects (which we fix in all experiments except the aforementioned ablation), there are no hyper-parameters introduced by the averaging protocol or the ensemble of averages studied in this paper, and all other experiments are purely observational. Finally, we followed the protocol of \cite{gulrajani2020search} for training and evaluation.

\textbf{Smaller HP search space}: We use a smaller hyper-parameter search space compared with that in \cite{gulrajani2020search}. Nonetheless, we find that on average, our runs of the ERM baseline performance (without model averaging) yield $64\%$ test accuracy on average compared with $63.8\%$ reported in \cite{gulrajani2020search} on the 5 datasets we used. We also note that model averaging and ensemble of averages, that we study in our work, are not competing with ERM baseline, in the sense that these techniques essentially rely on the quality of the baseline model to further boost performance. Therefore any boost in the ERM baseline performance is likely to improve the model averaging and EoA performance. This is also evident in our benchmarking experiments in Table \ref{tab:benchmark}, where using ResNeXt-50 32x4d pre-trained on a larger dataset \cite{yalniz2019billion} has a better ERM baseline performance compared to ResNet-50 pre-trained on ImageNet. This results in a further boost of $3.9\%$ and $5\%$ test accuracy on average when using model averaging and EoA respectively.

\begin{table*}[]
\small
\centering
\caption{Performance benchmarking on 5 datasets of the DomainBed benchmark using two different pre-trained models. SWAD is the previous SOTA. Note that ensembles do not have confidence interval because an ensemble uses all the models to make a prediction. Gray background shows our proposal. \textit{Our runs} implies we ran experiments, but we did not propose it.}
\begin{tabular}{llllll|l}
\hline
{Algorithm}          & {PACS}           & {VLCS}           & {OfficeHome}     & {TerraIncognita} & {DomainNet}      & {Avg}. \\ \hhline{======|=}
\multicolumn{7}{c}{{ResNet-50 (25M Parameters, pre-trained on ImageNet)}}\\
\hline
ERM (our runs) & 84.4 $\pm$ 0.8 & 77.1 $\pm$ 0.5 & 66.6 $\pm$ 0.2 & 48.3 $\pm$ 0.2 & 43.6 $\pm$ 0.1 & 64.0 \\
Ensemble (our runs) & 87.6             & 78.5           &       70.8         & 49.2           &          \textbf{47.7}      &   66.8   \\
ERM \cite{gulrajani2020search} & 85.7 $\pm$ 0.5 & 77.4 $\pm$ 0.3 & 67.5 $\pm$ 0.5 & 47.2 $\pm$ 0.4 & 41.2 $\pm$ 0.2 & 63.8\\ 
IRM \cite{arjovsky2019invariant}  & 84.4 $\pm$ 1.1            & 78.1 $\pm$ 0.0            & 66.6 $\pm$ 1.0            & 47.9 $\pm$ 0.7            & 35.7 $\pm$ 1.9 & 62.5\\
Group DRO \cite{sagawa2019distributionally}   & 84.1 $\pm$ 0.4            & 77.2 $\pm$ 0.6            & 66.9 $\pm$ 0.3            & 47.0 $\pm$ 0.3            & 33.7 $\pm$ 0.2 & 61.8\\
Mixup \cite{xu2020adversarial,wang2020heterogeneous}   & 84.3 $\pm$ 0.5            & 77.7 $\pm$ 0.4            & 69.0 $\pm$ 0.1            & 48.9 $\pm$ 0.8            & 39.6 $\pm$ 0.1 & 63.9\\
MLDG  \cite{li2018learning}    & 84.8 $\pm$ 0.6            & 77.1 $\pm$ 0.4            & 68.2 $\pm$ 0.1            & 46.1 $\pm$ 0.8            & 41.8 $\pm$ 0.4 & 63.6\\
CORAL  \cite{sun2016deep}   & 86.0 $\pm$ 0.2            & 77.7 $\pm$ 0.5            & 68.6 $\pm$ 0.4            & 46.4 $\pm$ 0.8            & 41.8 $\pm$ 0.2 & 64.1\\
MMD   \cite{li2018domain}  & 85.0 $\pm$ 0.2            & 76.7 $\pm$ 0.9            & 67.7 $\pm$ 0.1            & 49.3 $\pm$ 1.4            & 39.4 $\pm$ 0.8 & 63.6\\
DANN \cite{ganin2016domain}  & 84.6 $\pm$ 1.1            & 78.7 $\pm$ 0.3            & 65.4 $\pm$ 0.6            & 48.4 $\pm$ 0.5            & 38.4 $\pm$ 0.0 & 63.1\\
C-DANN  \cite{li2018deep}  & 82.8 $\pm$ 1.5            & 78.2 $\pm$ 0.4            & 65.6 $\pm$ 0.5            & 47.6 $\pm$ 0.8            & 38.9 $\pm$ 0.1 & 62.6\\
Fish \cite{shi2021gradient} & 85.5 $\pm$ 0.3 & 77.8 $\pm$ 0.3 & 68.6 $\pm$ 0.4 & 45.1 $\pm$ 1.3 & 42.7 $\pm$ 0.2 & 63.9\\
Fishr \cite{rame2021fishr} & 85.5 $\pm$ 0.4 & 77.8 $\pm$ 0.1 & 67.8 $\pm$ 0.1 & 47.4 $\pm$ 1.6 & 41.7 $\pm$ 0.0 & 65.7\\
SWAD \cite{cha2021swad} & 88.1 $\pm$ 0.4 & \textbf{79.1 $\pm$ 0.4} & 70.6 $\pm$ 0.3 & 50.0 $\pm$ 0.4 & 46.5 $\pm$ 0.2 & 66.9\\
MIRO \cite{cha2022domain} & 85.4 $\pm$ 0.4 & {79.0 $\pm$ 0.} & 70.5 $\pm$ 0.4 & 50.4 $\pm$ 1.1 & 44.3 $\pm$ 0.2 & 65.9\\
\hline
\rowcolor{Gray}
SMA (ours)    & 87.5 $\pm$ 0.2 & 78.2 $\pm$ 0.2   & 70.6 $\pm$ 0.1   & 50.3 $\pm$ 0.5 & 46 $\pm$ 0.1 & 66.5 \\
\rowcolor{Gray}
EoA (ours) & \textbf{88.6}           & \textbf{79.1}           & \textbf{72.5}          & \textbf{52.3}           &        {47.4}        & \textbf{68.0}\\
\hhline{======|=}
\multicolumn{7}{c}{{ResNeXt-50 32x4d  \cite{yalniz2019billion} (25M Parameters, Pre-trained 1B Images)}}\\
\hline
ERM (our runs) & 88.9 $\pm$ 0.3 & 79.0 $\pm$ 0.1 & 70.9 $\pm$ 0.5 & 51.4 $\pm$ 1.2 & 48.1 $\pm$ 0.2 & 67.7 \\ 
Ensemble (our runs) & 91.2             & 80.3           &       77.8         & 53.5           &          52.8      &   71.1   \\
 \hline
 \rowcolor{Gray}
SMA (ours)    & {92.7 $\pm$ 0.3} & {79.7 $\pm$ 0.3}   & {78.6 $\pm$ 0.1}   & {53.3 $\pm$ 0.1} & {53.5 $\pm$ 0.1} & {71.6} \\ 
\rowcolor{Gray}
EoA (ours) & \textbf{93.2}           & \textbf{80.4}           &  \textbf{80.2}  & \textbf{55.2}   &  \textbf{54.6}   & \textbf{72.7} \\
\hhline{======|=}
\multicolumn{7}{c}{{RegNetY-16GF \cite{singh2022revisiting} (81M Parameters, Pre-trained on 3.6B Images)}}\\
\hline
ERM (our runs) & 92.0 $\pm$ 0.4 & 78.6 $\pm$ 0.6 &  $73.8 \pm$ 0.5 & 55.6 $\pm$ 0.9  & 53.2 $\pm$ 0.2 & 70.6 \\ 
Ensemble (our runs) &     95.1        &     80.6       &        80.5       &      59.5     &     57.8       &   74.7 \\
ERM \cite{cha2022domain} & 89.6 $\pm$ 0.4 & 78.6 $\pm$ 0.3 & 71.9 $\pm$ 0.6 & 51.4 $\pm$ 1.8 & 48.5 $\pm$ 0.6 & 68.0 \\
SWAD \cite{cha2022domain}& 94.7 $\pm$ 0.2 & 79.7 $\pm$ 0.2 & 80.0 $\pm$ 0.1 & 57.9 $\pm$ 0.7 & 53.6 $\pm$ 0.6 & 73.2 \\
MIRO \cite{cha2022domain}& \textbf{97.4 $\pm$ 0.2} & 79.9 $\pm$ 0.6 & 80.4 $\pm$ 0.2 & 58.9 $\pm$ 1.3 & 53.8 $\pm$ 0.1 & 74.1\\
 \hline
 \rowcolor{Gray}
SMA (ours)    & {95.5$\pm$ 0.0} & {80.7 $\pm$ 0.1}   & {82.0 $\pm$ 0.0}   & {59.7 $\pm$ 0.0} & {60.0 $\pm$ 0.0} & {75.6} \\ 
\rowcolor{Gray}
EoA (ours) & {95.8}           & \textbf{81.1}           &  \textbf{83.9}  & \textbf{61.1}   &  \textbf{60.9}   & \textbf{76.6} \\
\hline
\end{tabular}
\label{tab:benchmark_full}
\end{table*}

\begin{table*}[]
\centering
\caption{Out-domain accuracy for PACS dataset.}
\begin{tabular}{lllll|l}
\hline
\textbf{Algorithm}          & \textbf{A}           & \textbf{C}           & \textbf{P}     & \textbf{S}    & \textbf{Avg}. \\ \hhline{=====|=}
\multicolumn{6}{c}{{ResNet-50}}\\
\hline
ERM & 86.4 $\pm$ 1.0           & 80.4 $\pm$ 0.6           & 94.8 $\pm$ 0.1          & 76.2 $\pm$ 1.7         & 84.4\\
Ensemble & 88.3             & \textbf{83.6}           &       96.5         & 81.9 &   87.6  \\
\hline
\rowcolor{Gray}
SMA & 89.1 $\pm$ 0.1           & 82.6 $\pm$ 0.2           & 97.6 $\pm$ 0.0          & 80.5 $\pm$ 0.9         & 87.5\\
 \rowcolor{Gray}
Ensemble of Averages (EoA) & \textbf{90.5}           & 83.4           & \textbf{98.0}           &        \textbf{82.5}        & \textbf{88.6}\\ 
\hhline{=====|=}
\multicolumn{6}{c}{{ResNeXt-50 32x4d  \cite{yalniz2019billion}}}\\
\hline
ERM & 84.7 $\pm$ 1.6 & 87.6 $\pm$ 0.1 & 97.6 $\pm$ 0.4 & 85.7 $\pm$ 0.1 & 88.9 \\ 
Ensemble & 90.2             & 89.2                 & 98.1           &          87.2      &   91.2   \\
\hline
\rowcolor{Gray}
SMA    & 92.6 $\pm$ 0.3 & 90.9 $\pm$ 0.8   & 99.1 $\pm$ 0.3  & 88.3 $\pm$ 0.5 & 92.7 \\ 
 \rowcolor{Gray}
Ensemble of Averages (EoA) & \textbf{93.1}           & \textbf{91.8 }     & \textbf{99.2}   &  \textbf{88.9}   & \textbf{93.2} \\
\hline
\multicolumn{6}{c}{{RegNetY-16GF \cite{singh2022revisiting}}}\\
\hline
ERM & 90.2 $\pm$ 0.6 & 92.6 $\pm$ 0.8 & 97.6 $\pm$ 0.1 & 87.8 $\pm$ 2 & 92 \\ 
Ensemble &      93.75        &        95.35          &       98.02     &       93.38      &   95.1  \\
\hline
\rowcolor{Gray}
SMA    & 93.8 $\pm$ 0.3 & 95.8 $\pm$ 0.2   & 99.2 $\pm$ 0.2  & \textbf{93.4 $\pm$ 0.2} &  95.5\\ 
 \rowcolor{Gray}
Ensemble of Averages (EoA) & \textbf{94.09}           & \textbf{96.33}     & \textbf{99.52}   &  {93.31}   & \textbf{95.8} \\
\hline
\end{tabular}
\label{tab:pacs_perf}
\end{table*}

\begin{table*}[]
\centering
\caption{Out-domain accuracy for VLCS dataset.}
\begin{tabular}{lllll|l}
\hline
\textbf{Algorithm}          & \textbf{C}           & \textbf{L}           & \textbf{S}     & \textbf{V}    & \textbf{Avg}. \\ \hhline{=====|=}
\multicolumn{6}{c}{{ResNet-50}}\\
\hline
ERM & 98.5 $\pm$ 0.5           & 62.4 $\pm$ 1.4           & 72.1 $\pm$ 0.0          & 75.4 $\pm$ 0.1         & 77.1\\
Ensemble & 98.7             & \textbf{64.5}           &       72.1         & \textbf{78.9} &   78.5  \\
\hline
\rowcolor{Gray}
SMA & 99.0 $\pm$ 0.2           & 63.0 $\pm$ 0.2           & 74.5 $\pm$ 0.3          & 76.4 $\pm$ 1.1         & 78.2\\
 \rowcolor{Gray}
Ensemble of Averages (EoA) & \textbf{99.1}           & 63.1           & \textbf{75.9}           &        78.3        & \textbf{79.1}\\ 
\hhline{=====|=}
\multicolumn{6}{c}{{ResNeXt-50 32x4d  \cite{yalniz2019billion}}}\\
\hline
ERM & 97.0 $\pm$ 0.4 & 67.8 $\pm$ 0.7 & 75.7 $\pm$ 0.2 & 75.5 $\pm$ 0.6 & 79.0 \\ 
Ensemble & 98.4             & \textbf{66.1}                 & 76.4           &          {80.5}      &   80.3   \\
\hline
\rowcolor{Gray}
SMA    & \textbf{98.8 $\pm$ 0.2} & 63.3 $\pm$ 0.6   & 77.7 $\pm$ 0.2  & 79.2 $\pm$ 0.8 & 79.7 \\ 
 \rowcolor{Gray}
Ensemble of Averages (EoA) & 98.7           & 64.1      & \textbf{78.2}   &  \textbf{80.6}   & \textbf{80.4} \\
\hline
\multicolumn{6}{c}{{RegNetY-16GF \cite{singh2022revisiting}}}\\
\hline
ERM & 96.0 $\pm$ 0.4 & 66.0 $\pm$ 0.8  & 76.1 $\pm$ 0.7 & 76.2 $\pm$ 2.1 & 78.6 \\ 
Ensemble &       97.88     &       \textbf{67.28}       &    78.46        &       78.55        &   80.6   \\
\hline
\rowcolor{Gray}
SMA    & 98.1 $\pm$ 0.2 & 65.7 $\pm$ 1.1  & 79.2 $\pm$ 1.1 & 79.6 $\pm$ 0.2  & 80.7\\ 
 \rowcolor{Gray}
Ensemble of Averages (EoA) &     \textbf{98.23}     &    66.00 & \textbf{79.49}   &  \textbf{80.63}   & \textbf{81.1} \\
\hline
\end{tabular}
\label{tab:vlcs_perf}
\end{table*}

\begin{table*}[]
\centering
\caption{Out-domain accuracy for OfficeHome dataset.}
\begin{tabular}{lllll|l}
\hline
\textbf{Algorithm}          & \textbf{A}           & \textbf{C}           & \textbf{P}     & \textbf{R}    & \textbf{Avg}. \\ \hhline{=====|=}
\multicolumn{6}{c}{{ResNet-50}}\\
\hline
ERM & 60.5 $\pm$ 0.7           & 54.5 $\pm$ 0.8           & 74.7 $\pm$ 0.8          & 76.6 $\pm$ 0.2         & 66.6\\
Ensemble & 65.6             & 58.5           &       78.7         & 80.5 &   70.8  \\
\hline
\rowcolor{Gray}
SMA & 66.7 $\pm$ 0.5           & 57.1 $\pm$ 0.1           & 78.6 $\pm$ 0.1          & 80.0 $\pm$ 0         & 70.6\\
 \rowcolor{Gray}
Ensemble of Averages (EoA) & \textbf{69.1}           & \textbf{59.8}           & \textbf{79.5}           &        \textbf{81.5}        & \textbf{72.5}\\ 
\hhline{=====|=}
\multicolumn{6}{c}{{ResNeXt-50 32x4d  \cite{yalniz2019billion}}}\\
\hline
ERM & 64.7 $\pm$ 1.0 & 60.6 $\pm$ 0.3 & 77.1 $\pm$ 0.4 & 81.3 $\pm$ 0.2 & 70.9 \\ 
Ensemble & 74.1             & 67.3                 & 83.9           &          86.0      &   77.8  \\
\hline
\rowcolor{Gray}
SMA    & 76.7 $\pm$ 0.4 & 67.8 $\pm$ 0.0   & 84.0 $\pm$ 0.1  & 85.8 $\pm$ 0.1 & 78.6 \\ 
 \rowcolor{Gray}
Ensemble of Averages (EoA) & \textbf{79.0}           & \textbf{70.0}      & \textbf{85.2}   &  \textbf{86.5}   & \textbf{80.2} \\
\hline
\multicolumn{6}{c}{{RegNetY-16GF \cite{singh2022revisiting}}}\\
\hline
ERM &  $ 70.7\pm$ 1.3 & 60.0  $\pm$ 0.5 & 82.4 $\pm$ 0.5 & 82.1 $\pm$ 0.4  & 73.8 \\ 
Ensemble &     79.44     &      68.68    & 86.28 &         87.63     &  80.5  \\
\hline
\rowcolor{Gray}
SMA    & 81.1 $\pm$ 0.4 & 72.3 $\pm$ 0.6  & 86.6 $\pm$ 0.1 & 88.2 $\pm$ 0.1 & 82.0 \\ 
 \rowcolor{Gray}
Ensemble of Averages (EoA) & \textbf{83.89}           & \textbf{73.95}      & \textbf{88.22}   &  \textbf{89.37}   & \textbf{83.9} \\
\end{tabular}
\label{tab:office_perf}
\end{table*}

\begin{table*}[]
\centering
\caption{Out-domain accuracy for TerraIncognita dataset.}
\begin{tabular}{lllll|l}
\hline
\textbf{Algorithm}          & \textbf{L100}           & \textbf{L38}           & \textbf{L43}     & \textbf{L46}    & \textbf{Avg}. \\ \hhline{=====|=}
\multicolumn{6}{c}{{ResNet-50}}\\
\hline
ERM & 52.9 $\pm$ 3.3          & 43.3 $\pm$ 1.7           & 56.9 $\pm$ 0.4         & 40.2 $\pm$ 2.1         & 48.3\\
Ensemble & 53.0             & 42.6           &       60.5         & 40.8 &   49.2  \\
\hline
\rowcolor{Gray}
SMA & 54.9 $\pm$ 0.4           & 45.5 $\pm$ 0.6          & 60.1 $\pm$ 1.5          & 40.5 $\pm$ 0.4         & 50.3\\
 \rowcolor{Gray}
Ensemble of Averages (EoA) & \textbf{57.8}           & \textbf{46.5}           & \textbf{61.3}          &       \textbf{43.5}        & \textbf{52.3} \\ 
\hhline{=====|=}
\multicolumn{6}{c}{{ResNeXt-50 32x4d  \cite{yalniz2019billion}}}\\
\hline
ERM & 64.0 $\pm$ 0.0           & 44.7 $\pm$ 3.2           & 56.1 $\pm$ 3.0          & 40.9 $\pm$ 1.4         & 51.4\\
Ensemble & \textbf{65.7}             & 43.1               & 62.6       &        42.6    & 53.5  \\
\hline
\rowcolor{Gray}
SMA & 60.1 $\pm$ 1.0           & \textbf{47.3 $\pm$ 1.4}           & 61.0 $\pm$ 1.7          & 44.9 $\pm$ 0.8         & 53.3\\
 \rowcolor{Gray}
Ensemble of Averages (EoA) & 63.5        & 46.0   & \textbf{64.3}   &  \textbf{46.9}  & \textbf{55.2} \\
\hline
\multicolumn{6}{c}{{RegNetY-16GF \cite{singh2022revisiting}}}\\
\hline
ERM & 67.1 $\pm$    2.8        & 46.3 $\pm$    2.9        & 61.4 $\pm$  0    & 47.5  $\pm$  1.9      & 55.6 \\
Ensemble & {71.67}     &       50.76       &      64.11  &      \textbf{51.45}    &  59.5 \\
\hline
\rowcolor{Gray}
SMA & 72.4 $\pm$ 0.0          & {52.0 $\pm$ 0.5}           & 66.8  $\pm$ 0.4         & 47.4 $\pm$ 0.2        & 59.7 \\
 \rowcolor{Gray}
Ensemble of Averages (EoA) &   \textbf{73.80}     &  \textbf{52.60}  & \textbf{68.19}   &  {49.75}  & \textbf{61.1} \\
\hline
\end{tabular}
\label{tab:terra_perf}
\end{table*}

\begin{table*}[]
\small
\centering
\caption{Out-domain accuracy for Domainet dataset.}
\begin{tabular}{lllllll|l}
\hline
\textbf{Algorithm}          & \textbf{clip}           & \textbf{info}           & \textbf{paint}     & \textbf{quick}  & \textbf{real}  & \textbf{sketch}    & \textbf{Avg}. \\ \hhline{=======|=}
\multicolumn{7}{c}{{ResNet-50}}\\
\hline
ERM & 63.4 $\pm$ 0.2          & 20.6 $\pm$ 0.1           & 50.0 $\pm$ 0.1         & 13.8 $\pm$ 0.4      &   62.1 $\pm$ 0.2  & 51.9 $\pm$ 0.3 & 43.6\\
Ensemble & \textbf{68.3}             &  23.1           &     54.5  &  16.3  & \textbf{66.9} & 57.0 &   \textbf{47.7}  \\
\hline
\rowcolor{Gray}
SMA & 64.4 $\pm$ 0.3           & 22.4 $\pm$ 0.2          & 53.4 $\pm$ 0.3          & 15.4 $\pm$ 0.1     &  64.7 $\pm$ 0.2  &  55.5 $\pm$ 0.1  & 46.0 \\
 \rowcolor{Gray}
Ensemble of Averages (EoA) & 65.9        & \textbf{23.4}     & \textbf{55.3}  & \textbf{16.5}  & 66.4  &  \textbf{57.1}      & 47.4 \\ 
\hhline{=======|=}
\multicolumn{7}{c}{{ResNeXt-50 32x4d  \cite{yalniz2019billion}}}\\
\hline
ERM & 68.8 $\pm$ 0.1          & 25.5 $\pm$ 0.1     & 55.9 $\pm$ 0.3         & 14.7 $\pm$ 0.7   & 65.8 $\pm$ 0.4 &  58.0 $\pm$ 0.4 &  48.1 \\
Ensemble & 74.3       & 28.7   &  61.1 & 17.0   & 71.9    &     63.5   & 52.8 \\
\hline
\rowcolor{Gray}
SMA & 73.7 $\pm$ 0.1      & 29.9 $\pm$ 0.0     & 62.8 $\pm$ 0.1    & 18.1 $\pm$ 0.1  & 73.0 $\pm$ 0.2 & 63.6 $\pm$ 0.4  & 53.5 \\
 \rowcolor{Gray}
Ensemble of Averages (EoA) & \textbf{74.6}    & \textbf{31.3}  & \textbf{63.7}  & \textbf{19.3}  &   \textbf{73.6}  & \textbf{65.1 }& \textbf{54.6} \\
\hline
\multicolumn{6}{c}{{RegNetY-16GF \cite{singh2022revisiting}}}\\
\hline
ERM & 74.7 $\pm$    0.1    & 34.8$\pm$ 0.9   & 60.3 $\pm$  0      &  15.2 $\pm$ 0.6   & 71.1 $\pm$ 0.3 & 62.1  $\pm$ 0.1 & 53.1  \\
Ensemble &   79.12     &  38.71  & 65.89  &  18.13 & 76.43    &   68.40     & 57.8 \\
\hline
\rowcolor{Gray}
SMA & 78.8 $\pm$ 0.1    &  43.0 $\pm$ 0.0     & 68.6 $\pm$    0.0 & 21.2 $\pm$ 0.0  & 78.5 $\pm$ 0.1 & 69.8  $\pm$ 0.1 & 60 \\
 \rowcolor{Gray}
Ensemble of Averages (EoA) & \textbf{79.63}    & \textbf{44.02}  & \textbf{69.57}  & \textbf{22.46}  &   \textbf{77.95}  & \textbf{71.69}& \textbf{60.9} \\
\hline
\end{tabular}
\label{tab:domainnet_perf}
\end{table*}



\begin{figure*}
    \centering
    \includegraphics[width=0.36\columnwidth]{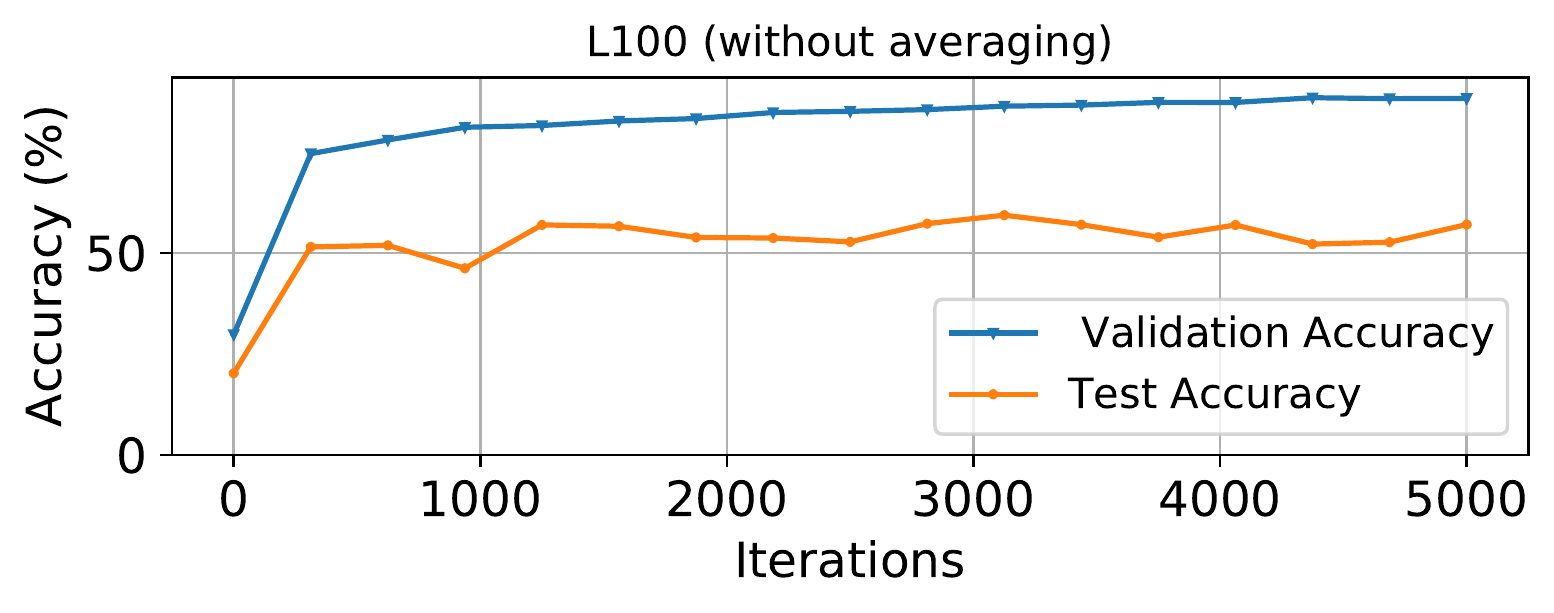}
    \hfil
    \includegraphics[width=0.36\columnwidth]{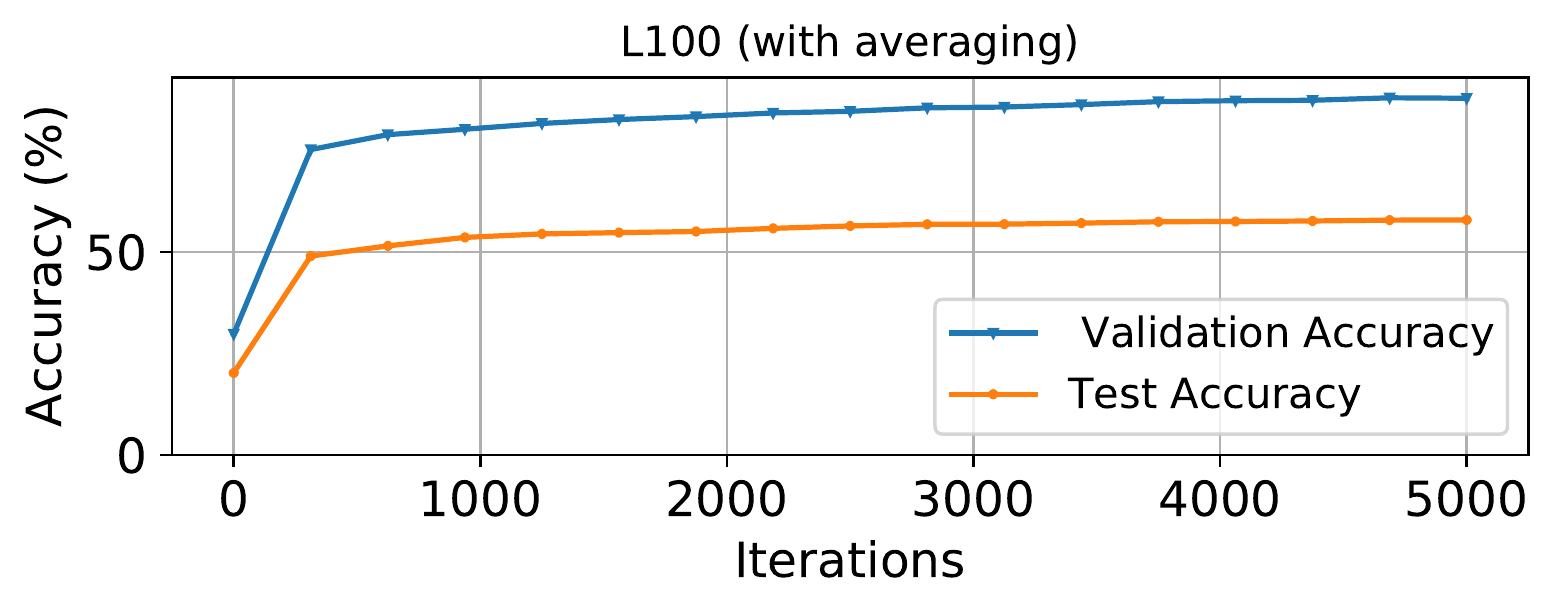}
    \vfil
      \includegraphics[width=0.36\columnwidth]{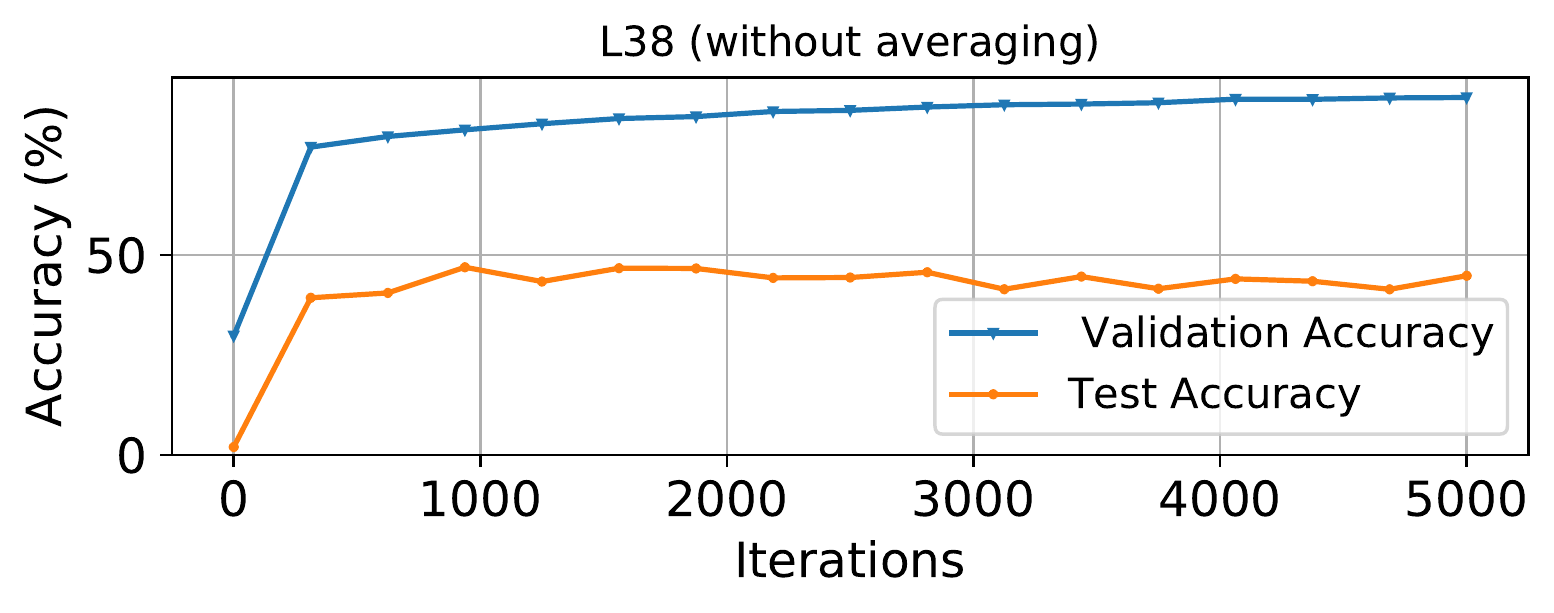}
      \hfil
      \includegraphics[width=0.36\columnwidth]{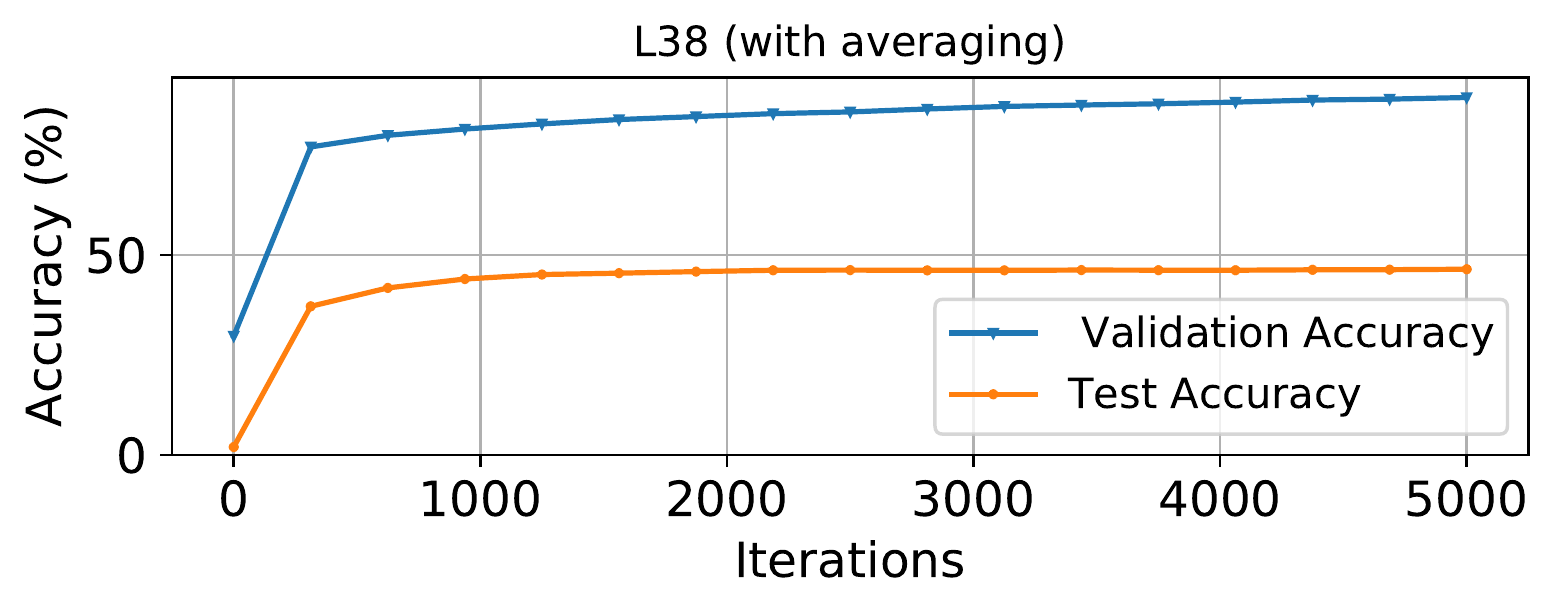}
      \vfil
      \includegraphics[width=0.36\columnwidth]{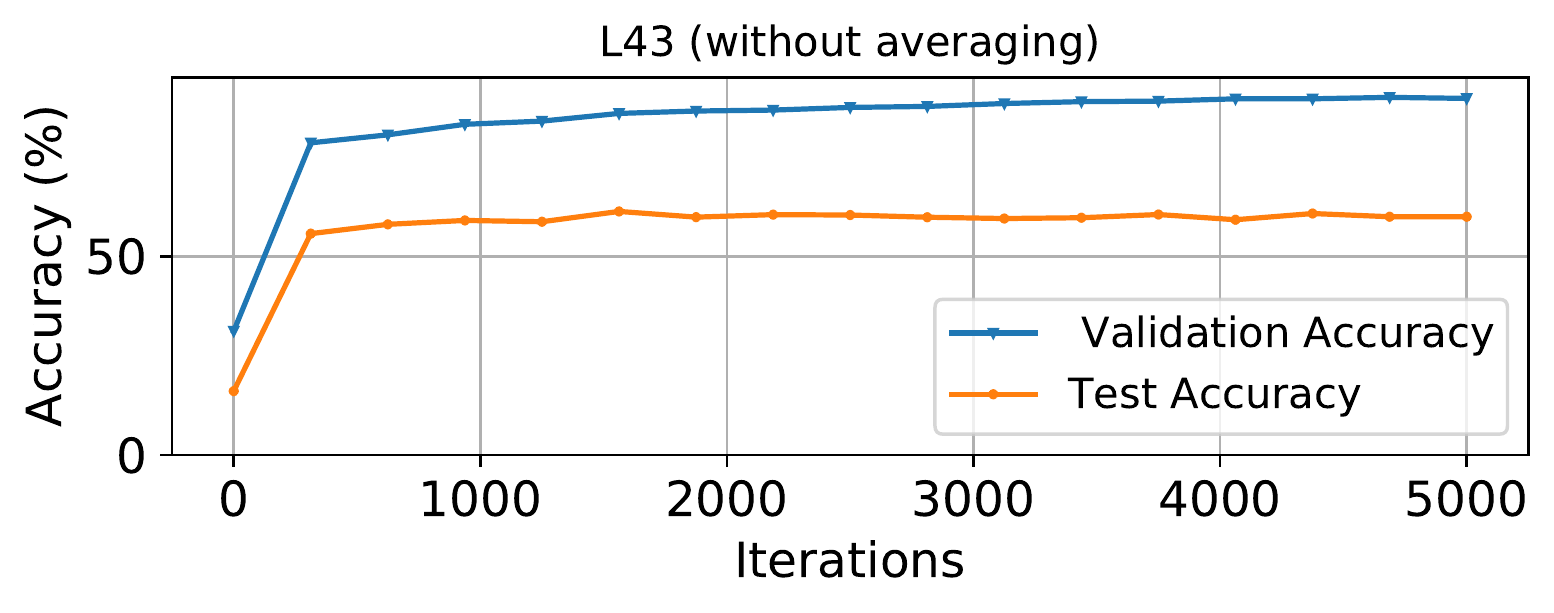}
    \hfil
    \includegraphics[width=0.36\columnwidth]{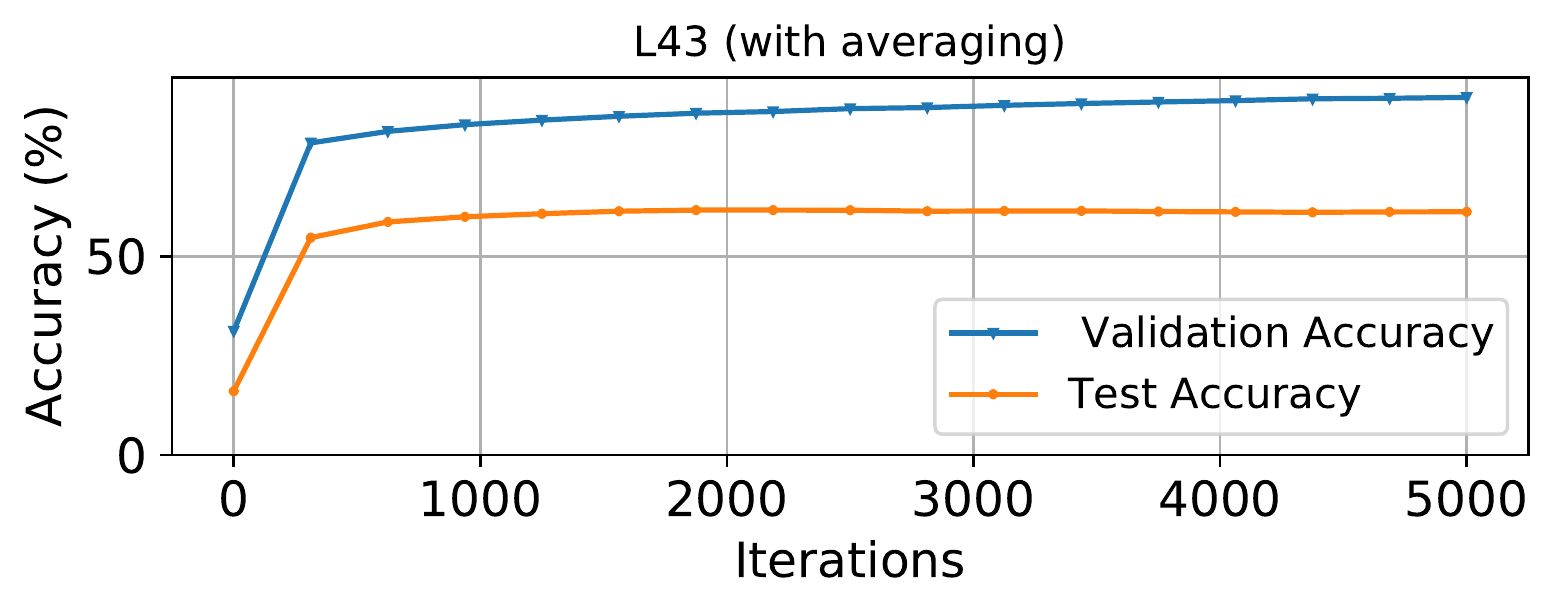}
    \vfil
      \includegraphics[width=0.36\columnwidth]{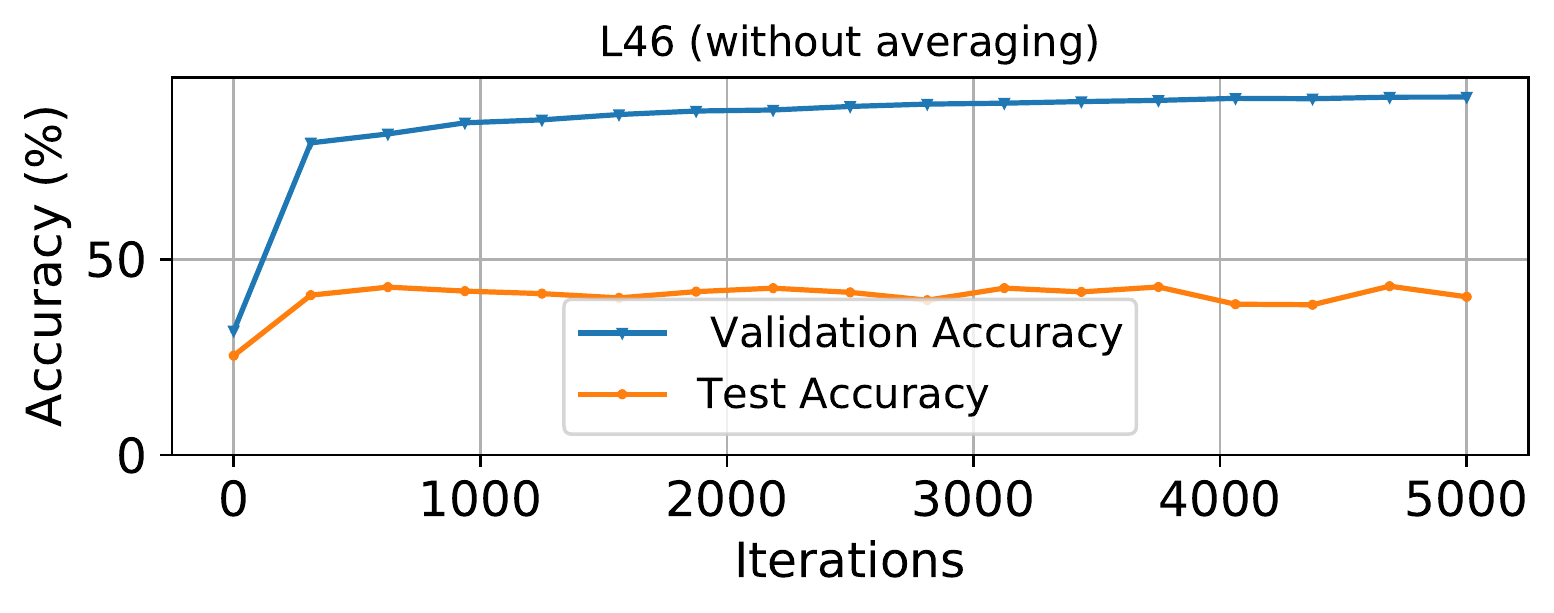}
    \hfil
      \includegraphics[width=0.36\columnwidth]{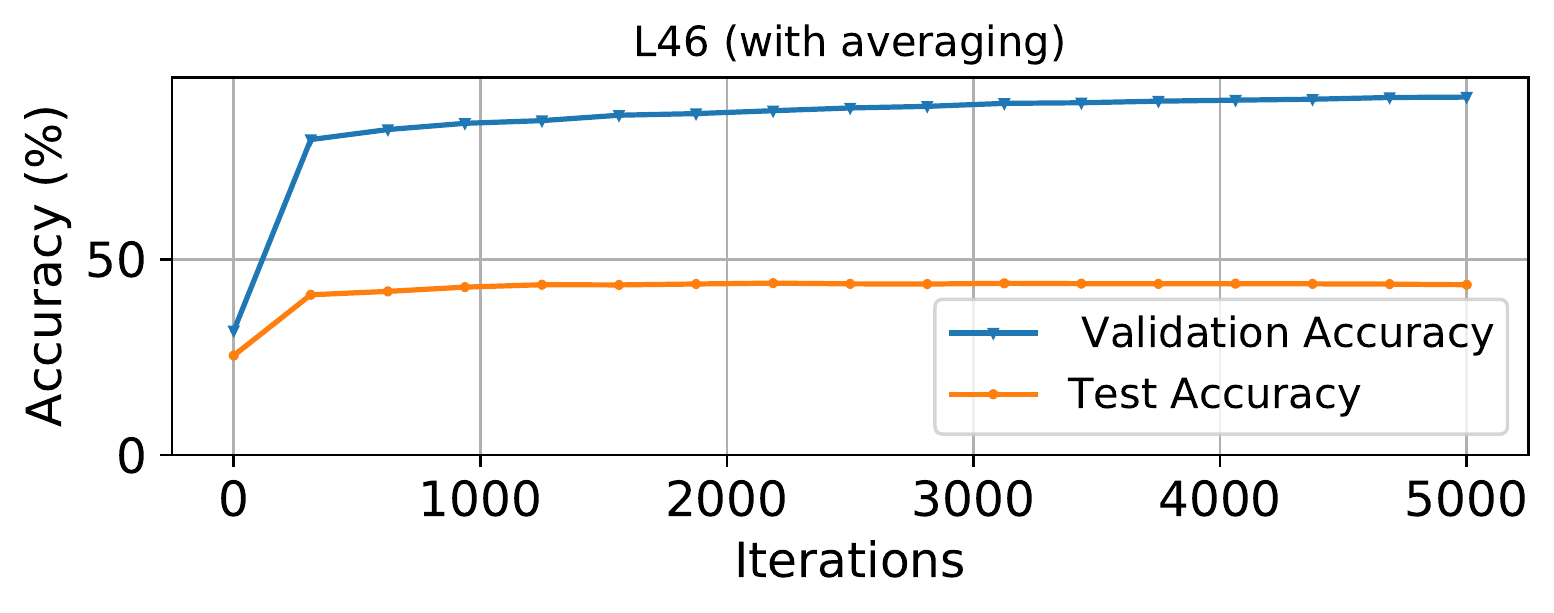}
    \caption{Ensemble of moving average (EOA) models (right) has better out-domain test performance \textit{stability} compared with ensemble of online models (left), w.r.t. in-domain validation accuracy. \textbf{Details}: The plots are for the TerraIncognita dataset with the domain name in title used as the test domain, and others as training/validation domain, and ResNet-50. Each ensemble has 6 different models from independent runs with a different random seed and training/validation split.}
    \label{fig:instability_ensemble_terra_all}
\end{figure*}

\begin{figure*}
    \centering
    \includegraphics[width=0.36\columnwidth]{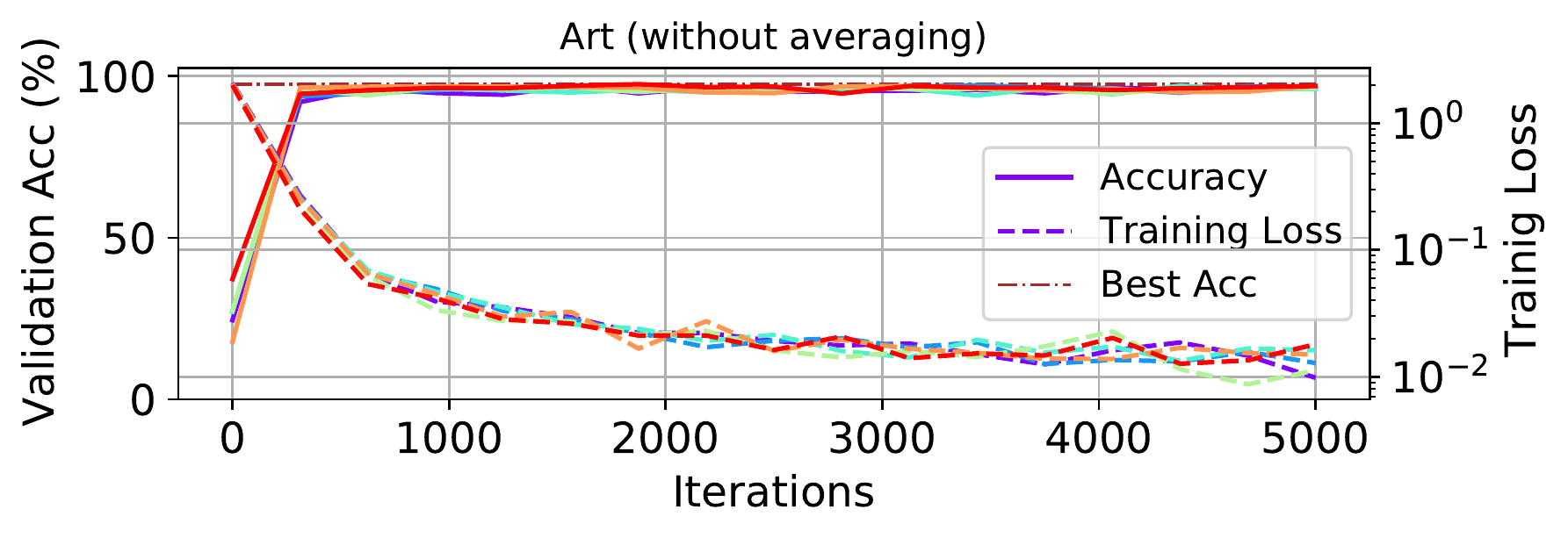}
    \hfil
      \includegraphics[width=0.36\columnwidth]{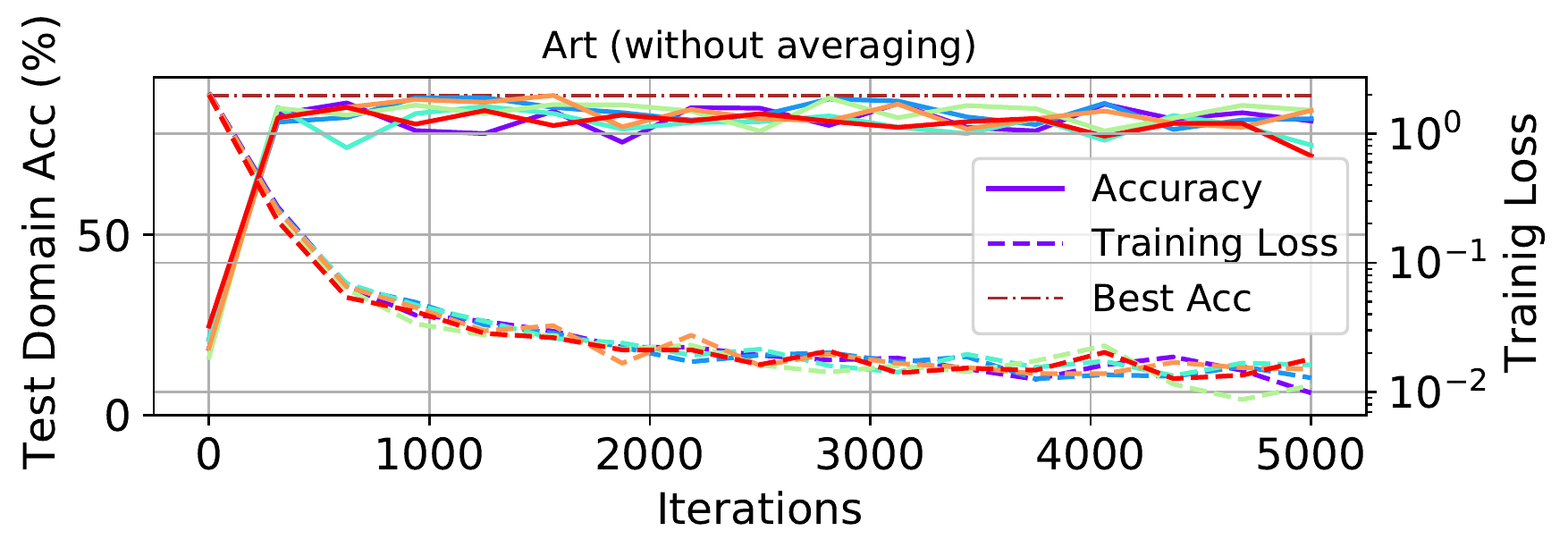}
      \vfil
      \includegraphics[width=0.36\columnwidth]{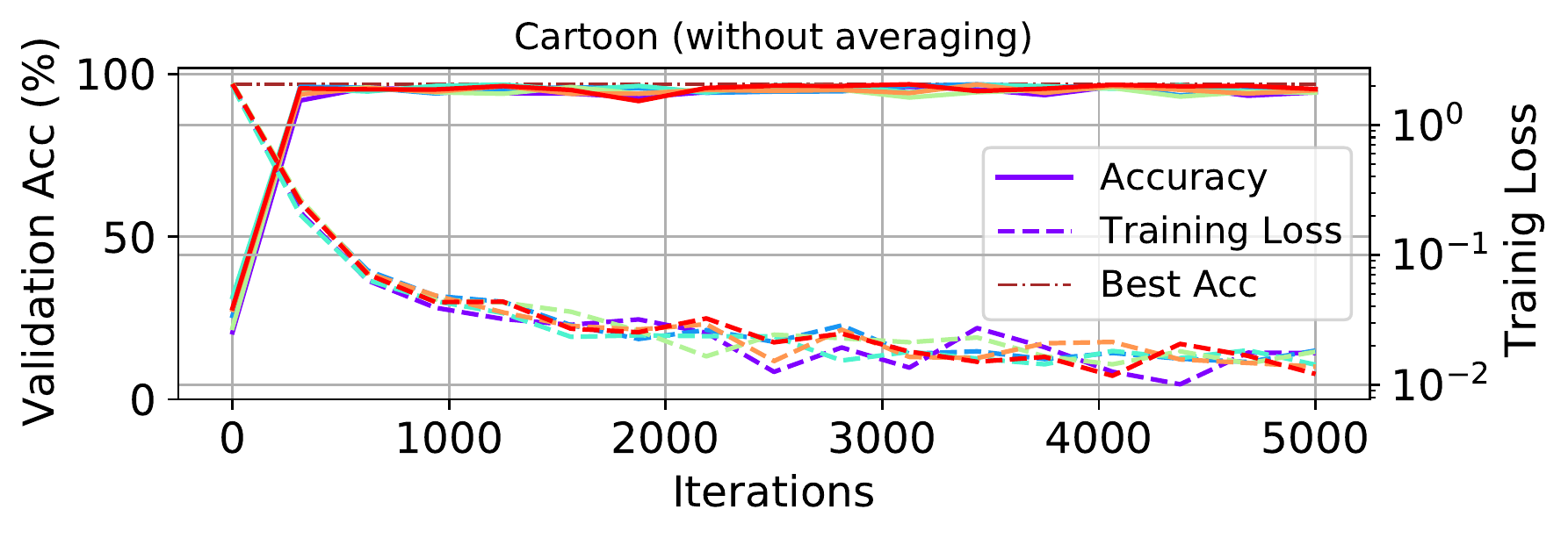}
    \hfil
      \includegraphics[width=0.36\columnwidth]{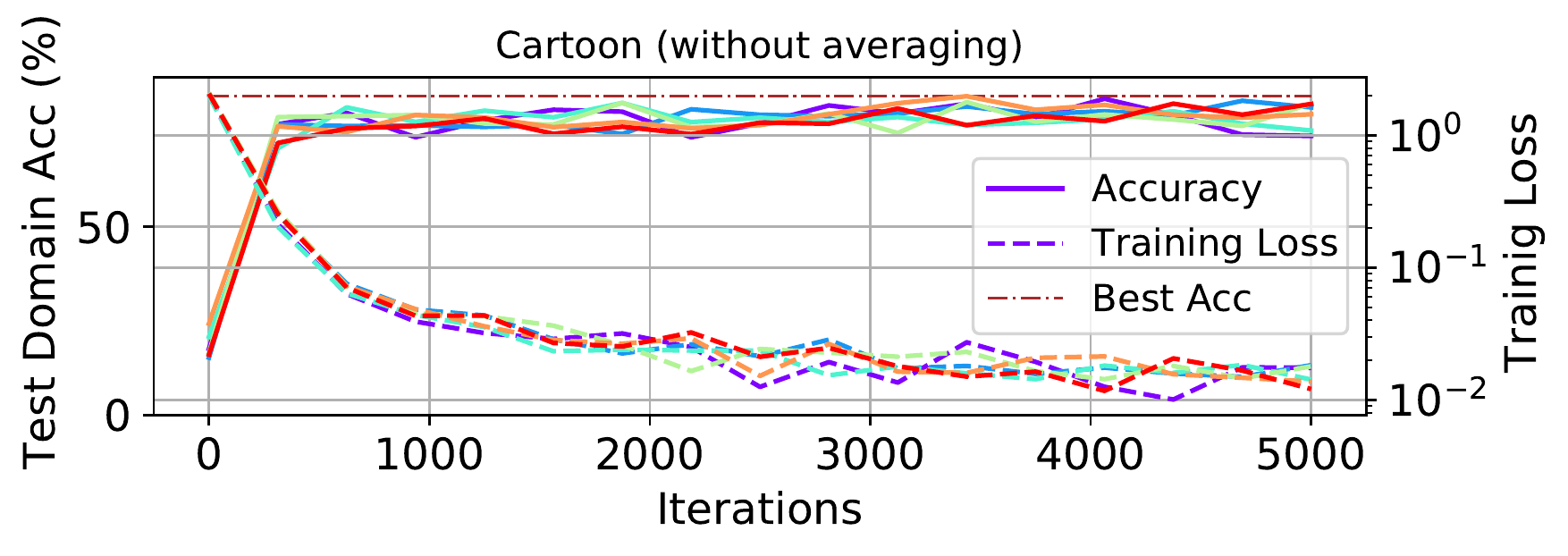}
      \vfil
      \includegraphics[width=0.36\columnwidth]{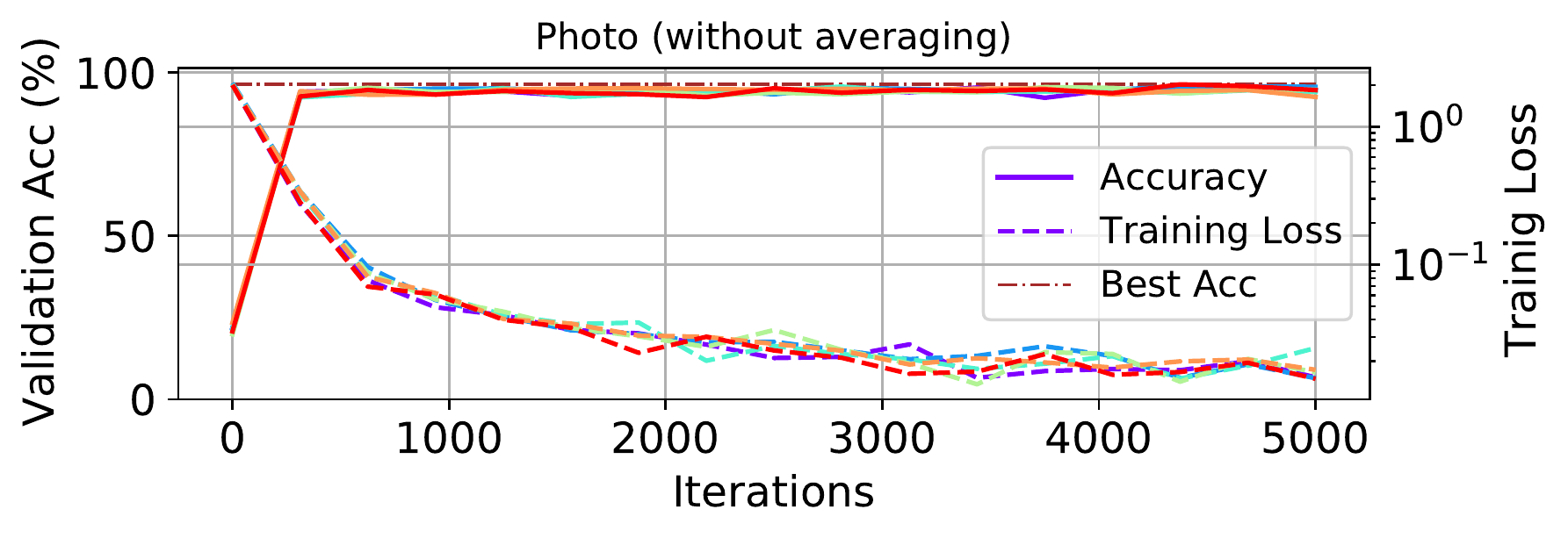}
    \hfil
      \includegraphics[width=0.36\columnwidth]{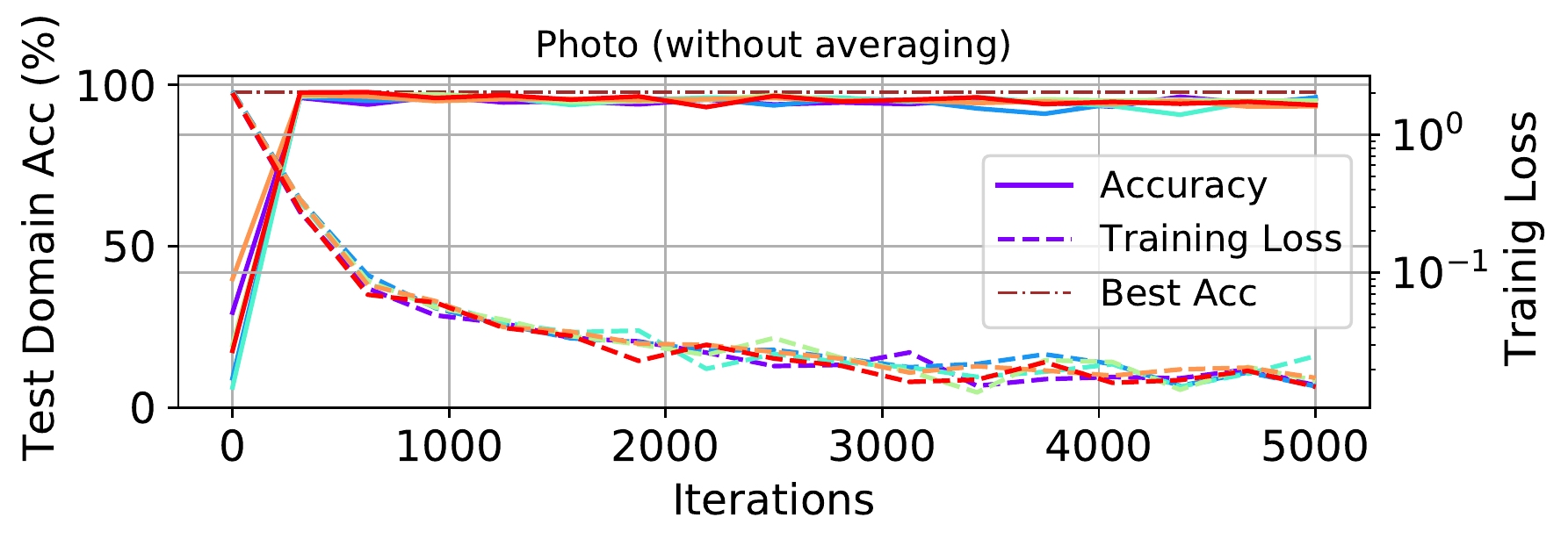}
      \vfil
      \includegraphics[width=0.36\columnwidth]{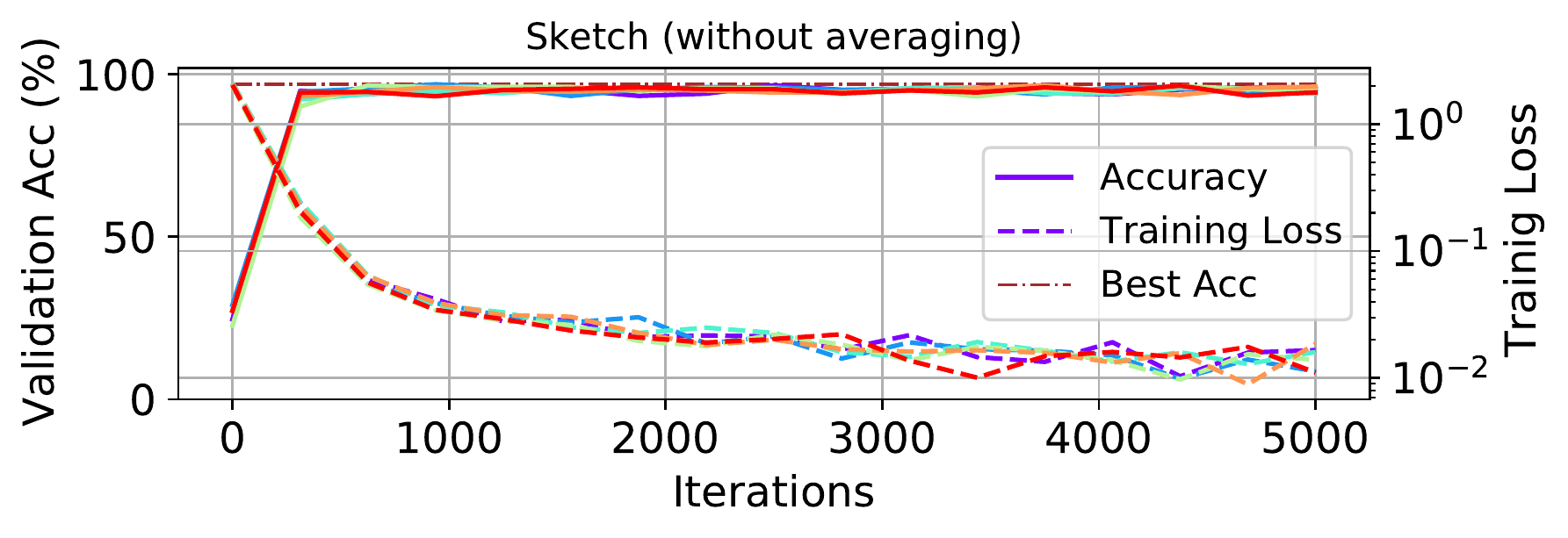}
    \hfil
      \includegraphics[width=0.36\columnwidth]{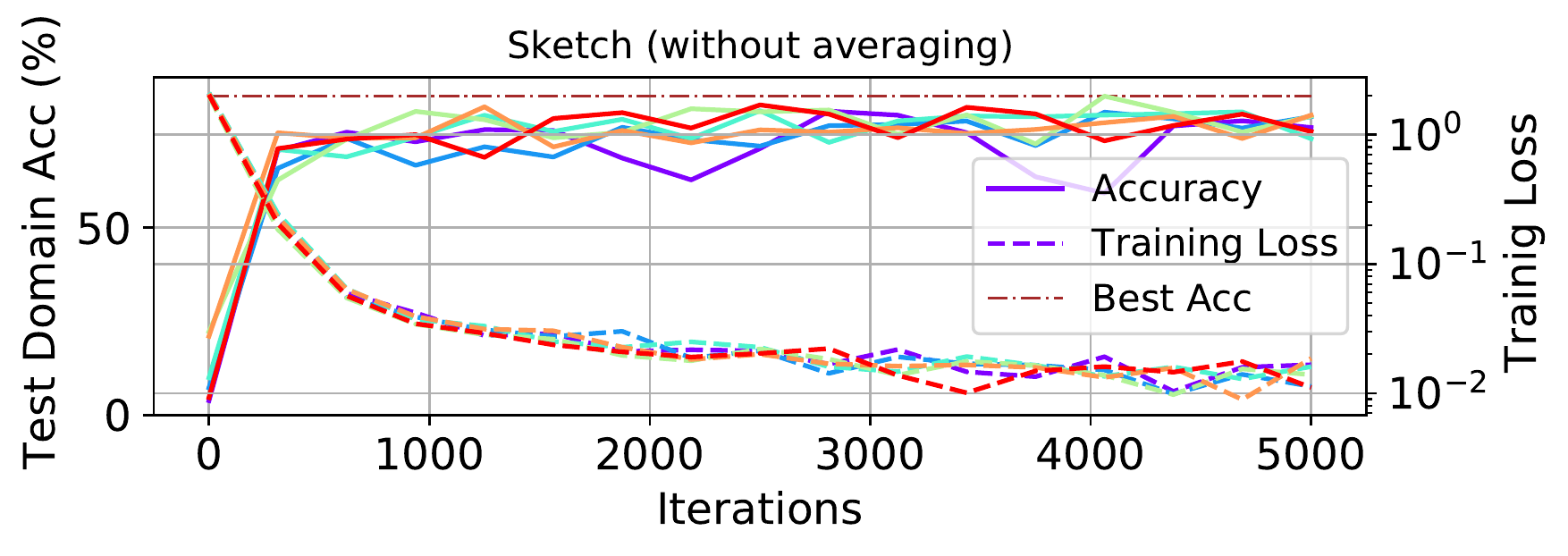}\\
    \hrulefill\vspace{10pt}\par
      \includegraphics[width=0.36\columnwidth]{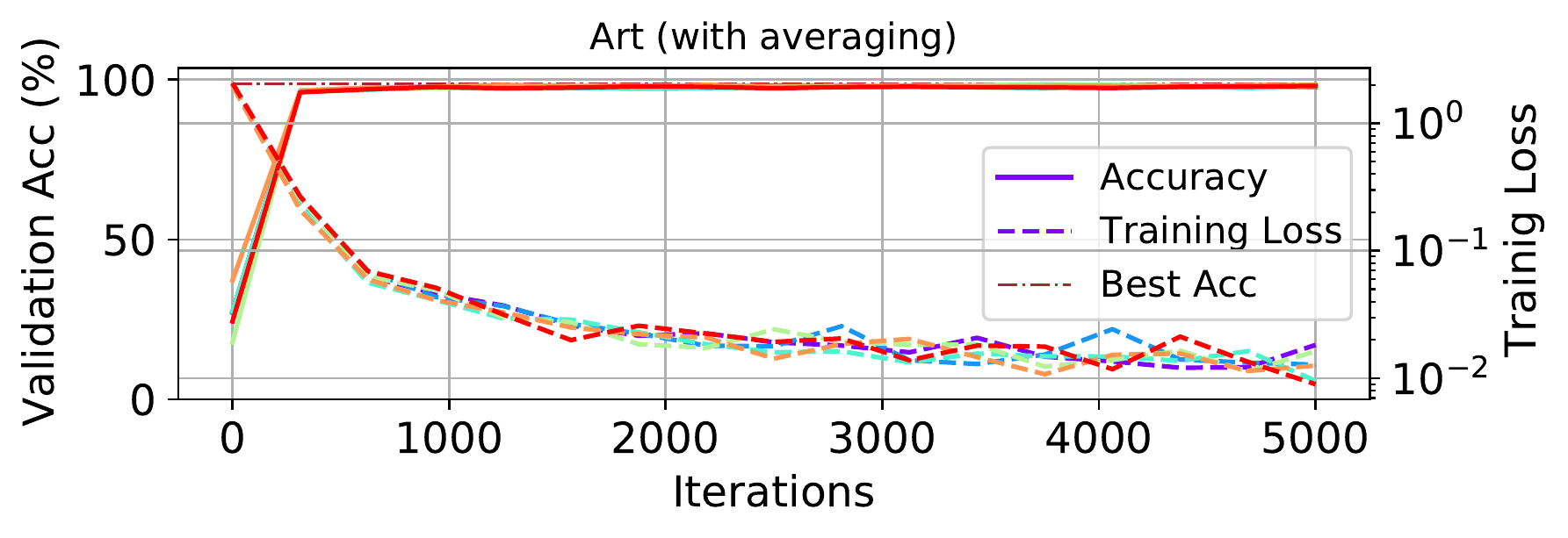}
    \hfil
      \includegraphics[width=0.36\columnwidth]{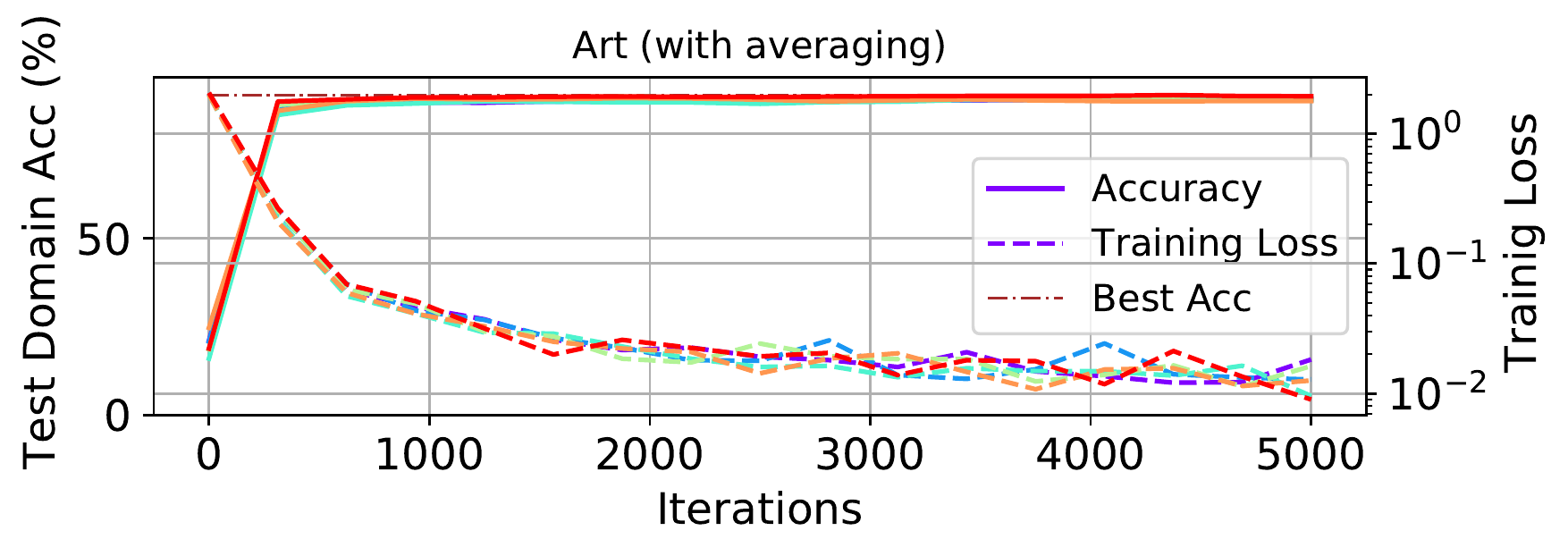}
      \vfil
      \includegraphics[width=0.36\columnwidth]{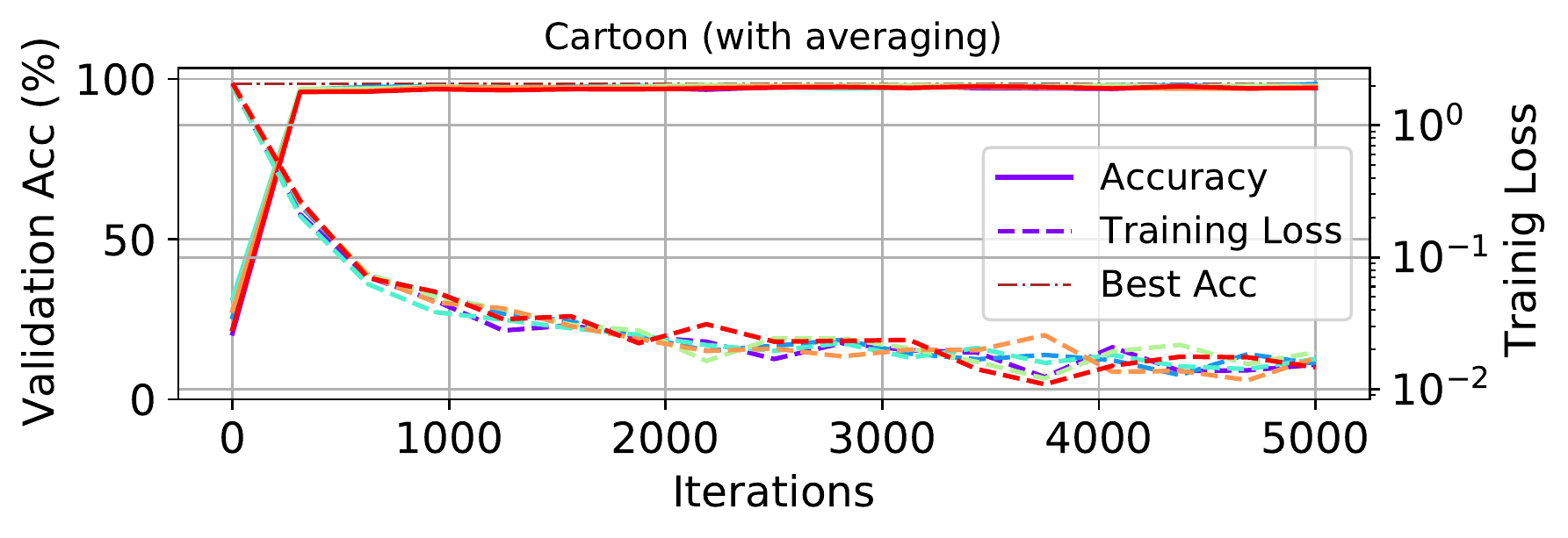}
    \hfil
      \includegraphics[width=0.36\columnwidth]{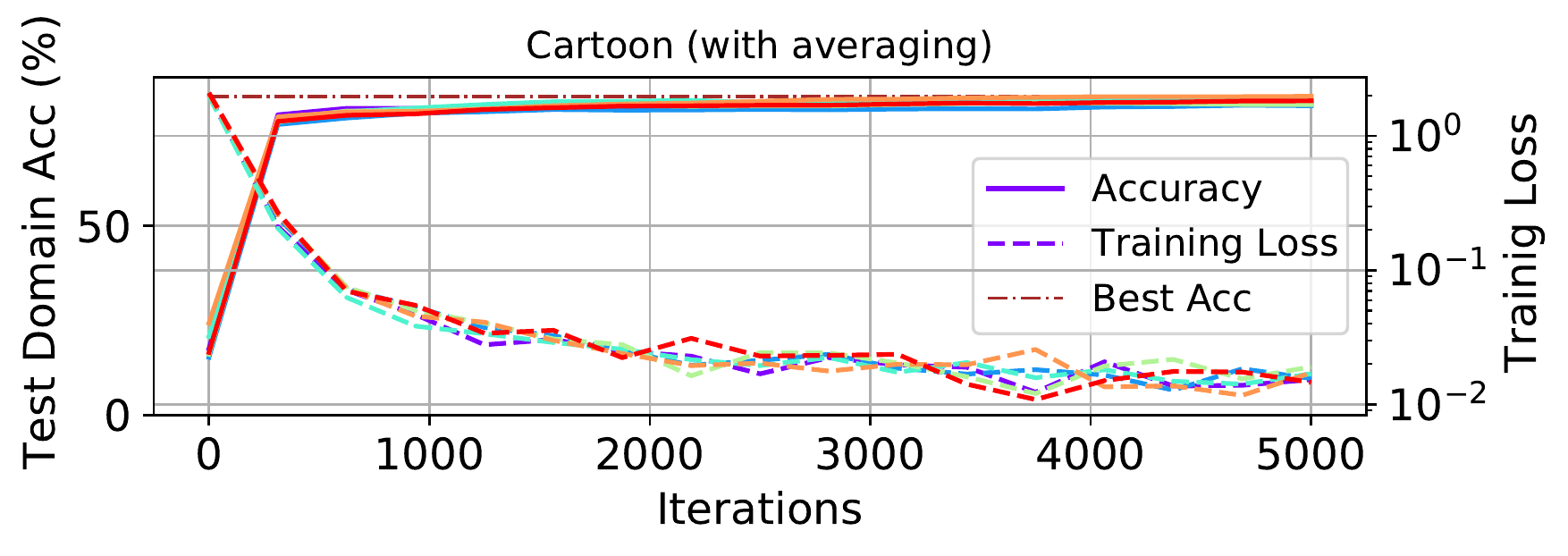}
      \vfil
      \includegraphics[width=0.36\columnwidth]{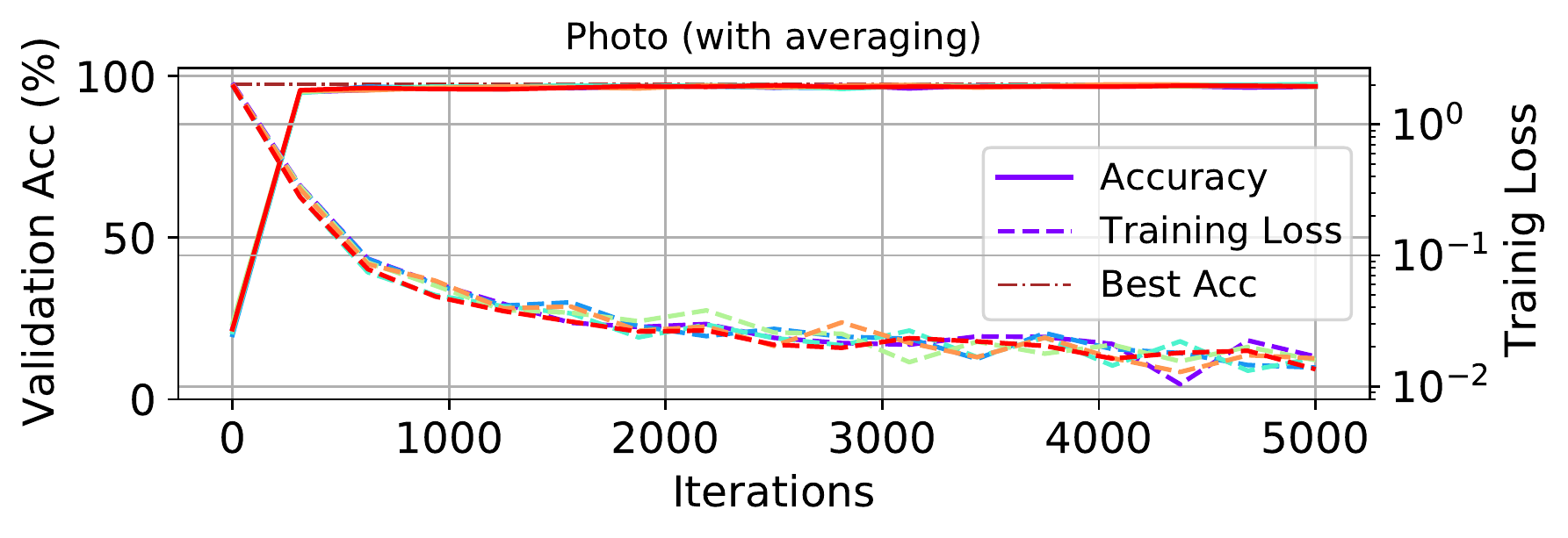}
    \hfil
      \includegraphics[width=0.36\columnwidth]{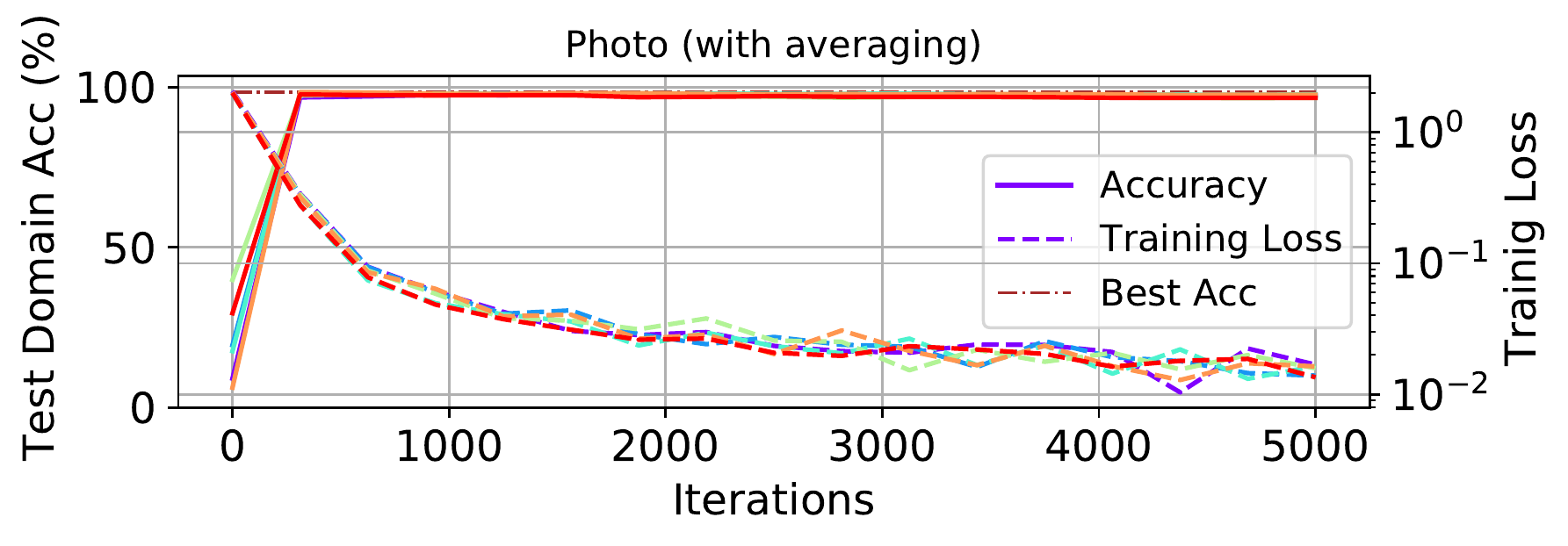}
      \vfil
      \includegraphics[width=0.36\columnwidth]{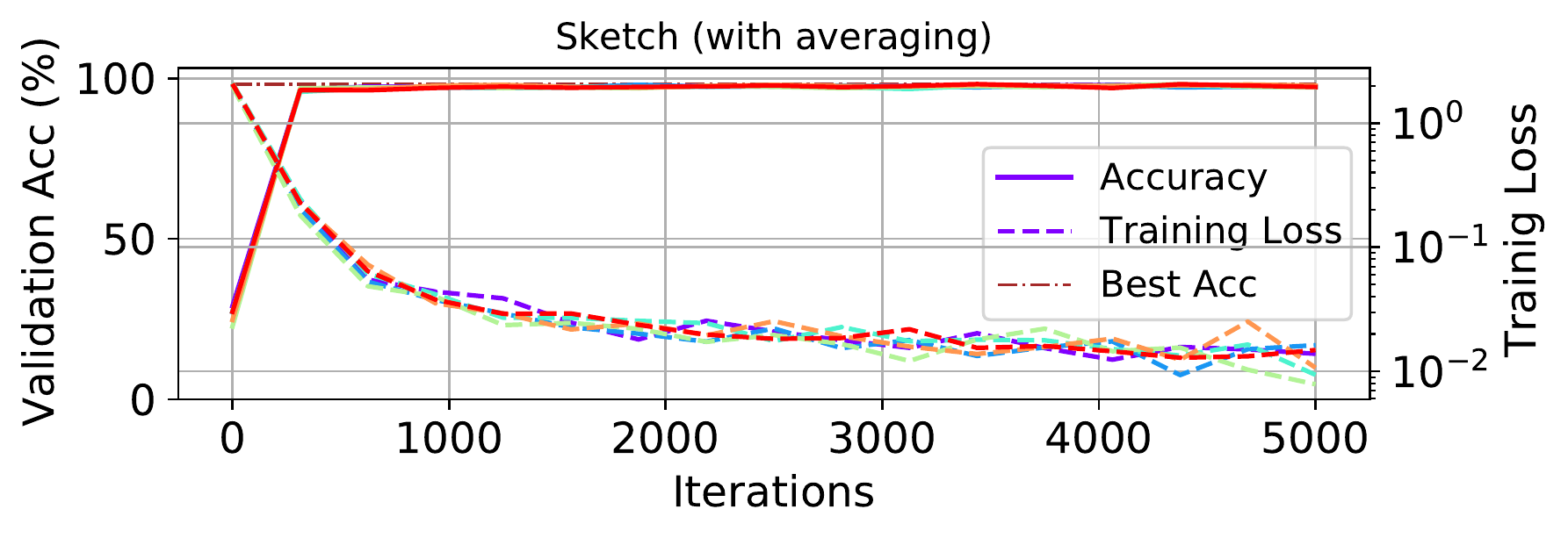}
    \hfil
      \includegraphics[width=0.36\columnwidth]{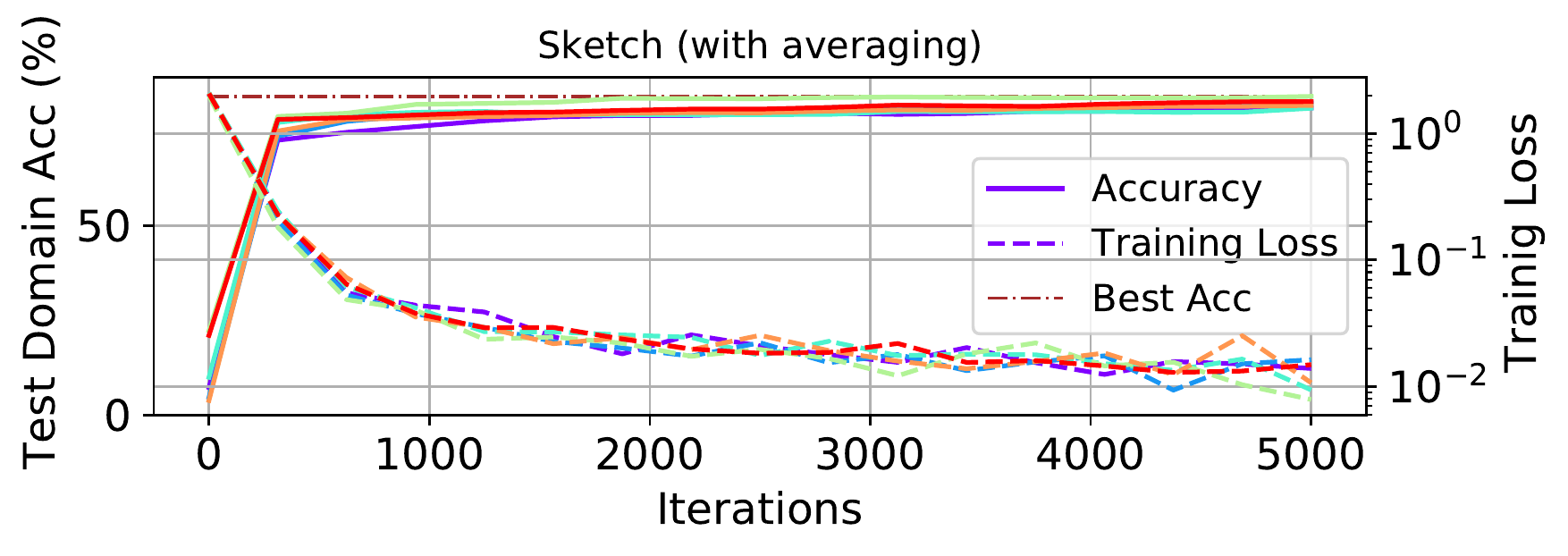}\\
    \caption{Evolution of training loss, in-domain validation accuracy and out-domain test accuracy for ResNet-50 (pre-trained on ImageNet) trained on PACS without model averaging (top 4 rows) and with model averaging (bottom 4 rows) for $5,000$ iterations with the domain mentioned in the title used as test domain and remain domains as training/validation data. Each color represents a different run with randomly chosen seed, hyper-parameters and training-validation split following \cite{gulrajani2020search}.}
    \label{fig:pacs_evolution_all}
\end{figure*}

\begin{figure*}
    \centering
    \includegraphics[width=0.36\columnwidth]{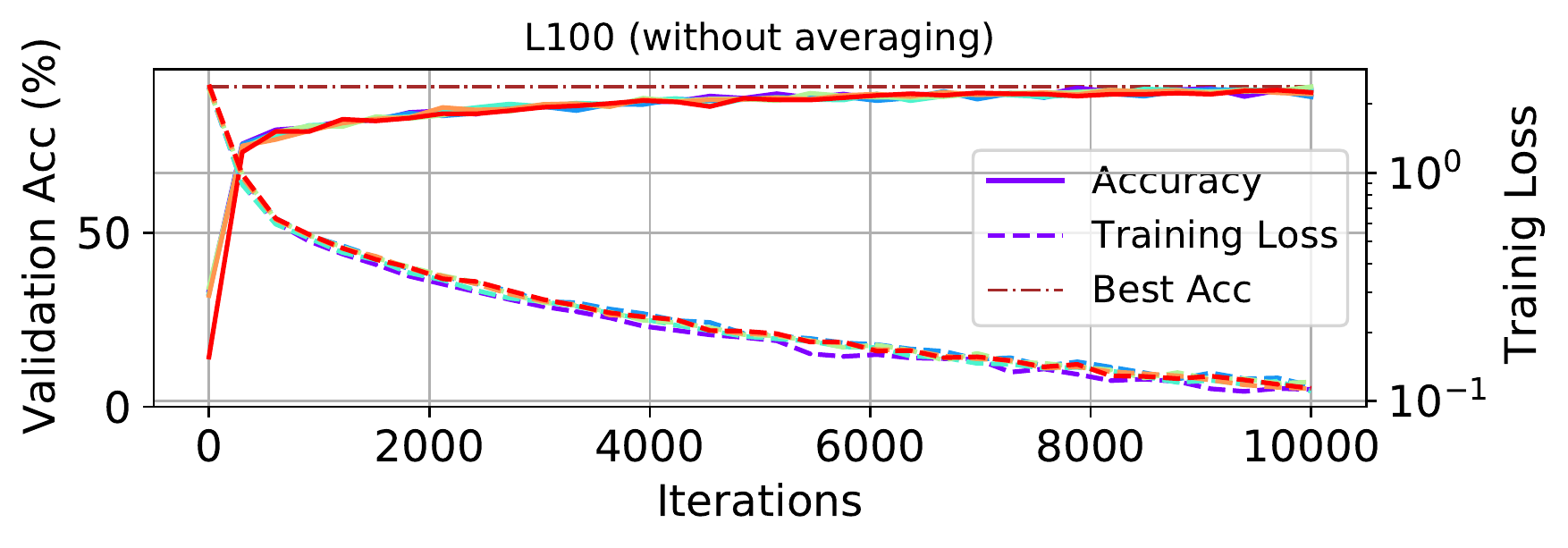}
    \hfil
      \includegraphics[width=0.36\columnwidth]{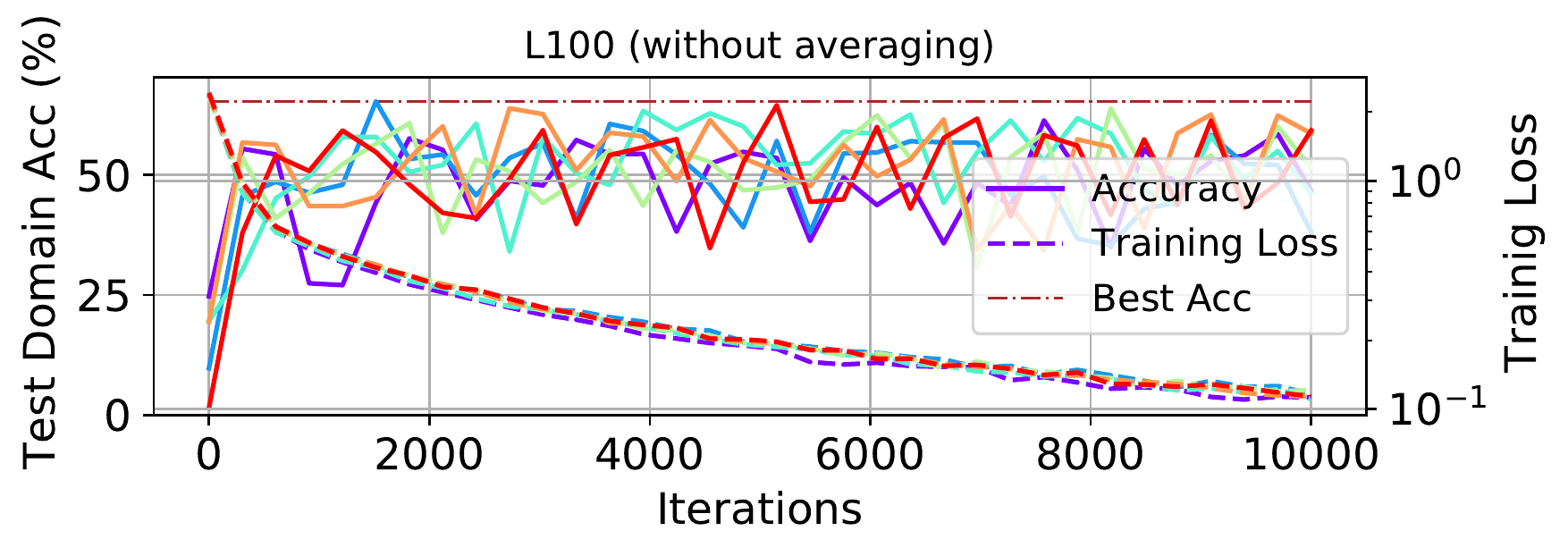}
      \vfil
      \includegraphics[width=0.36\columnwidth]{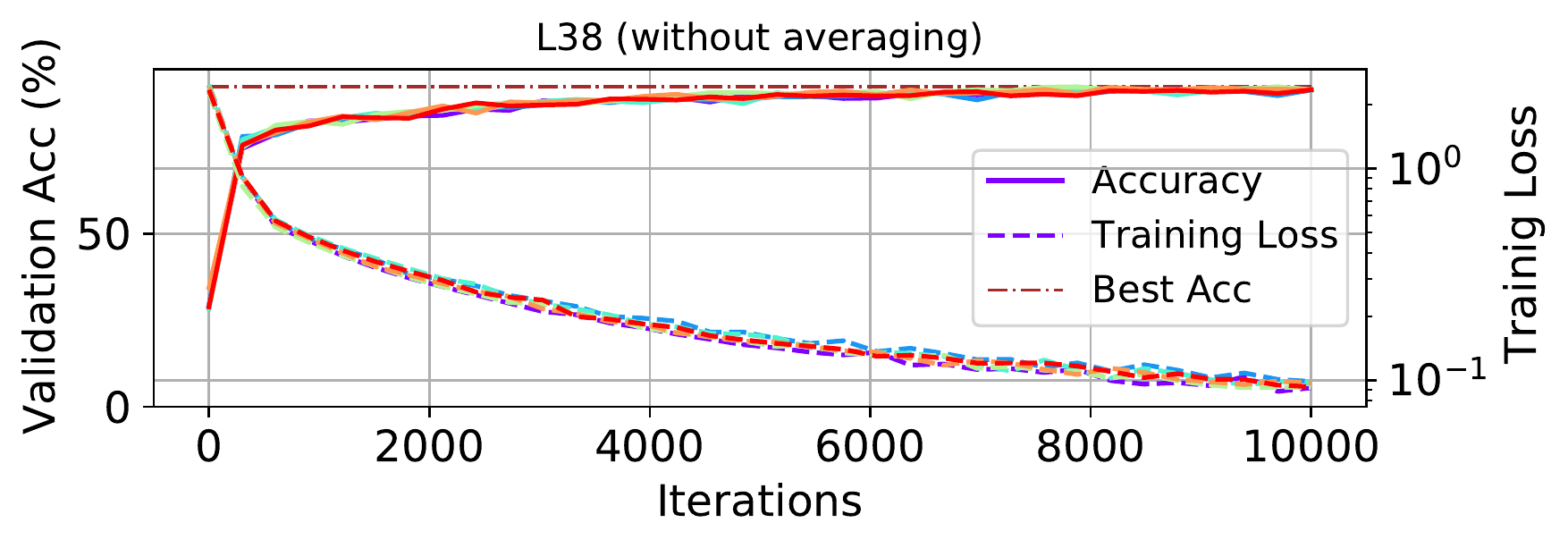}
    \hfil
      \includegraphics[width=0.36\columnwidth]{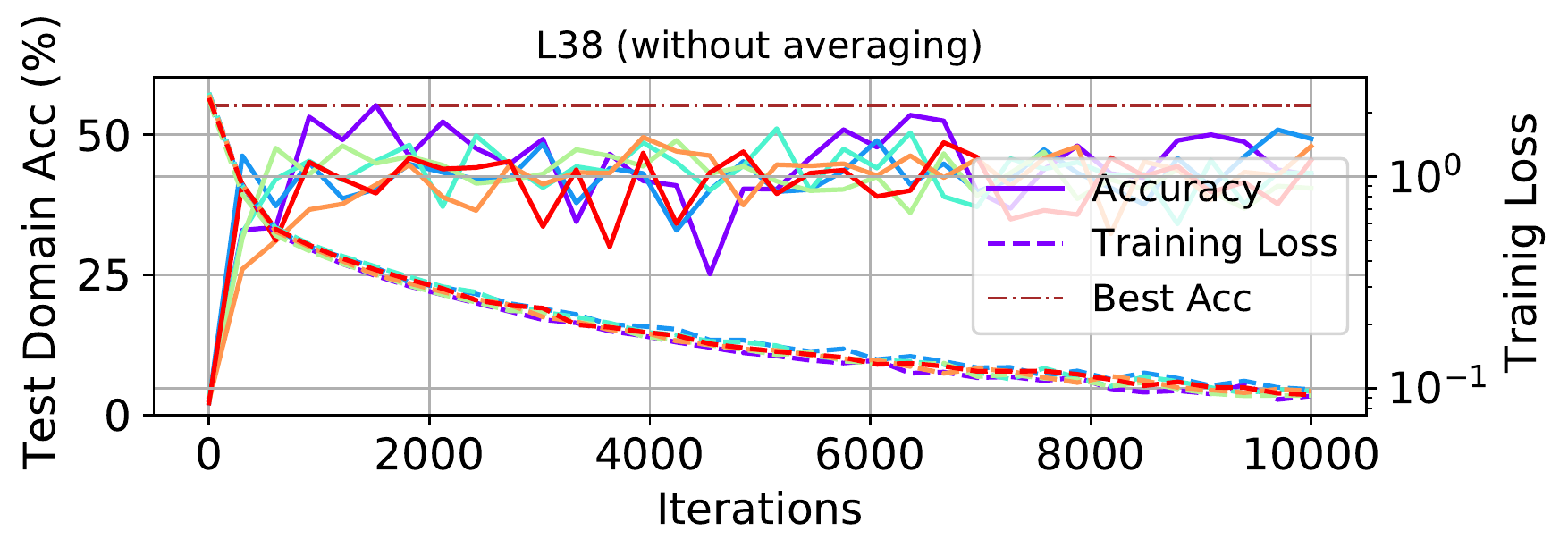}
      \vfil
      \includegraphics[width=0.36\columnwidth]{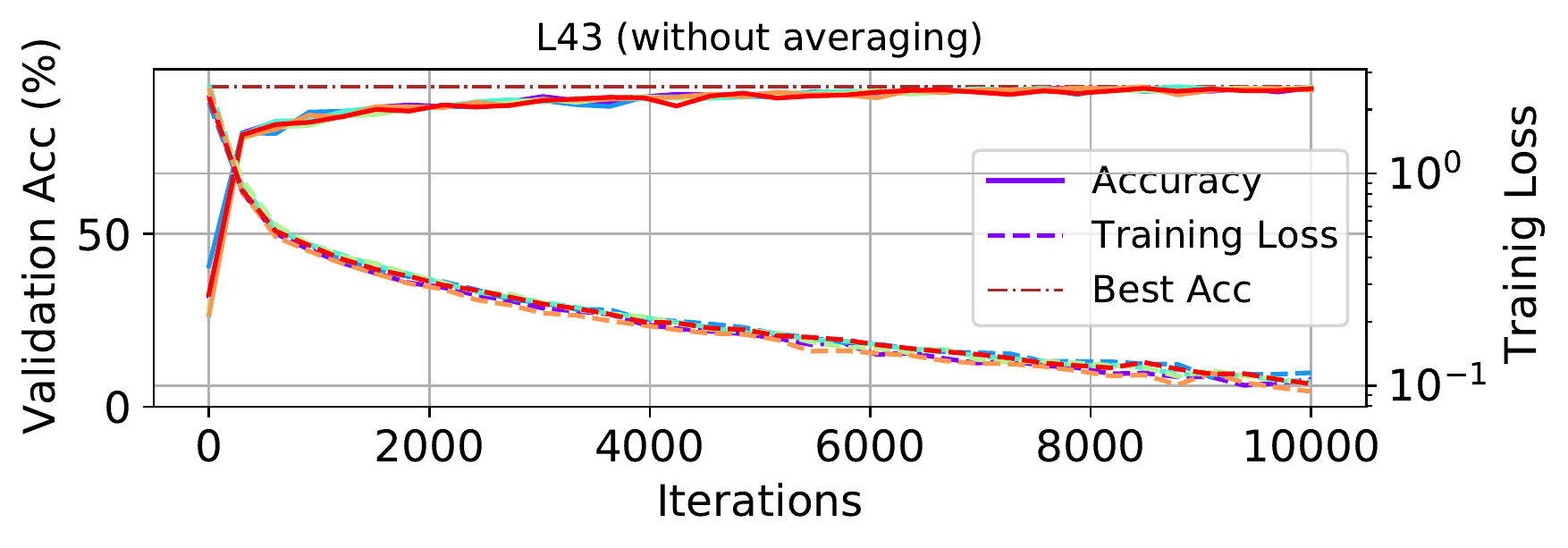}
    \hfil
      \includegraphics[width=0.36\columnwidth]{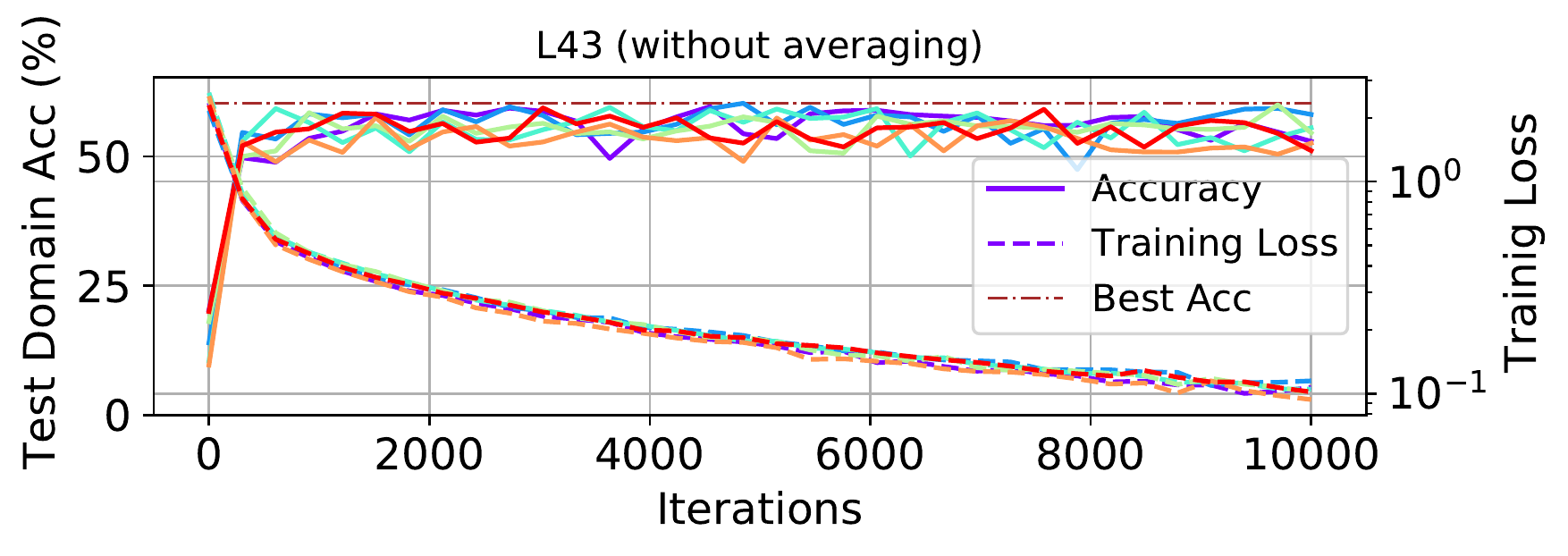}
      \vfil
      \includegraphics[width=0.36\columnwidth]{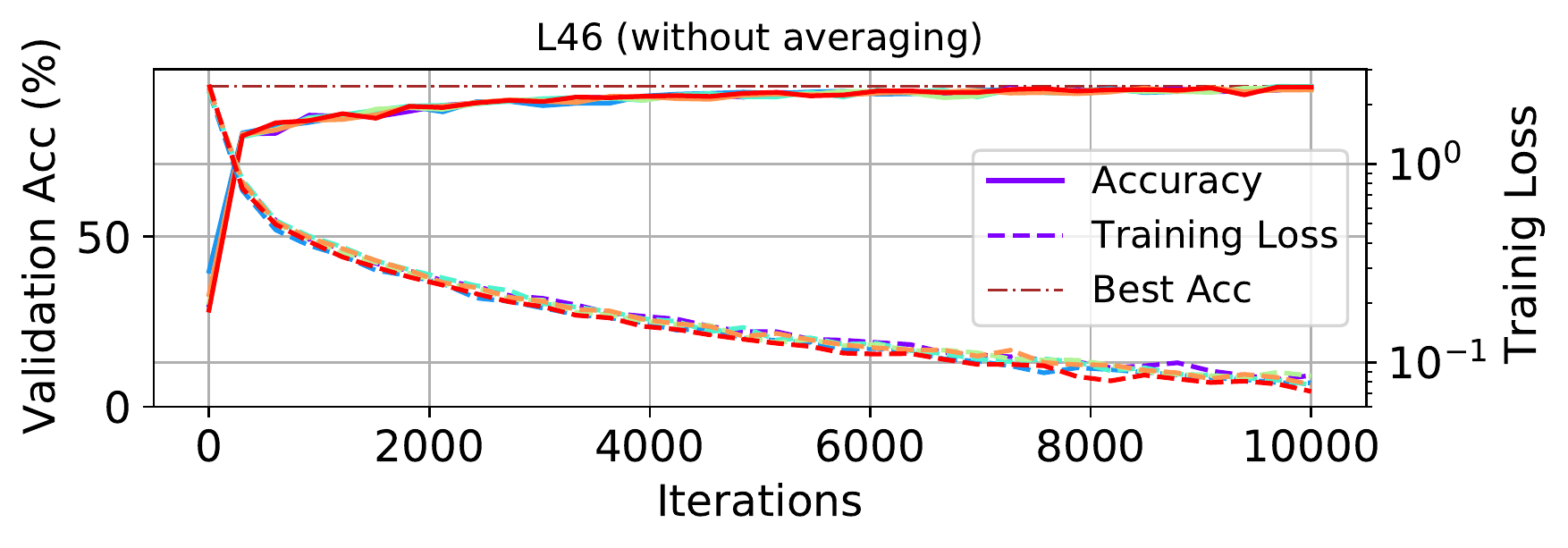}
    \hfil
      \includegraphics[width=0.36\columnwidth]{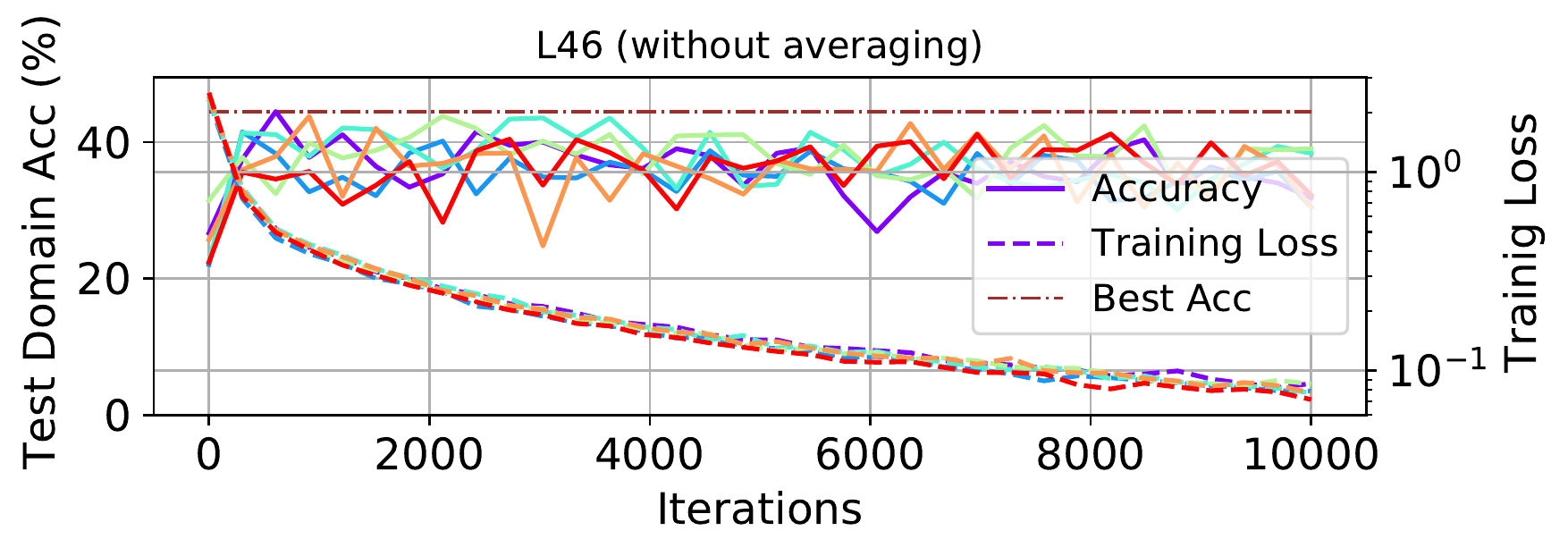}\\
    \hrulefill\vspace{10pt}\par
      \includegraphics[width=0.36\columnwidth]{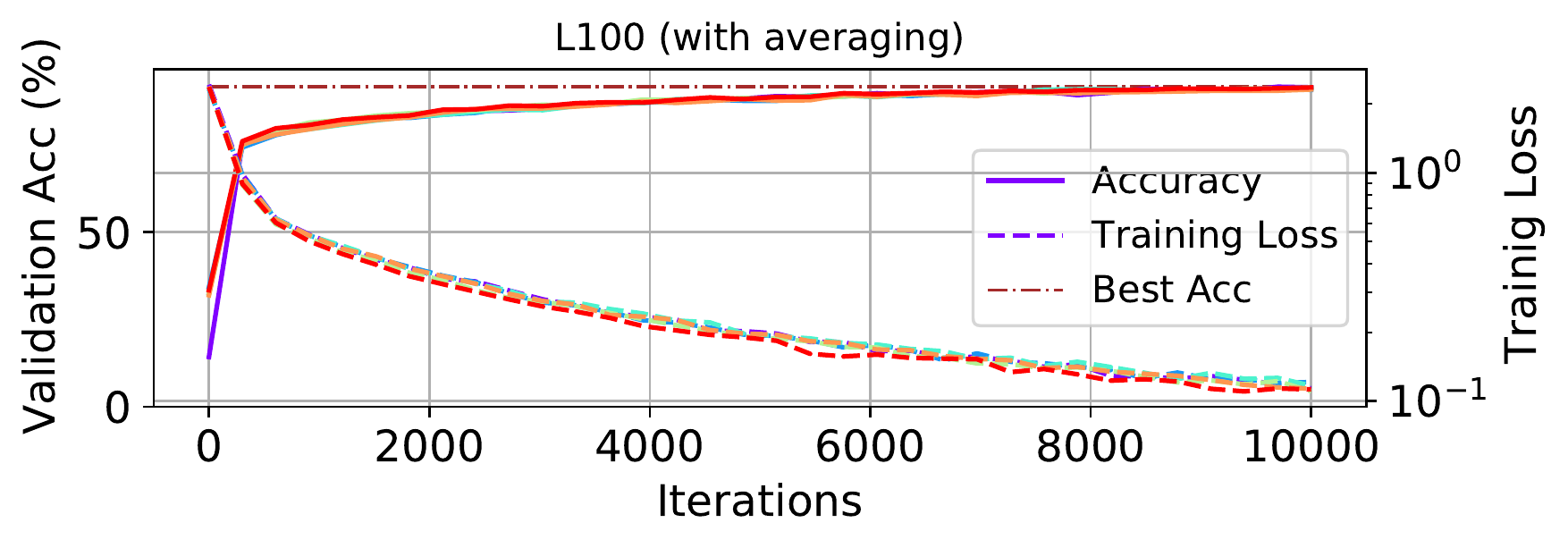}
    \hfil
      \includegraphics[width=0.36\columnwidth]{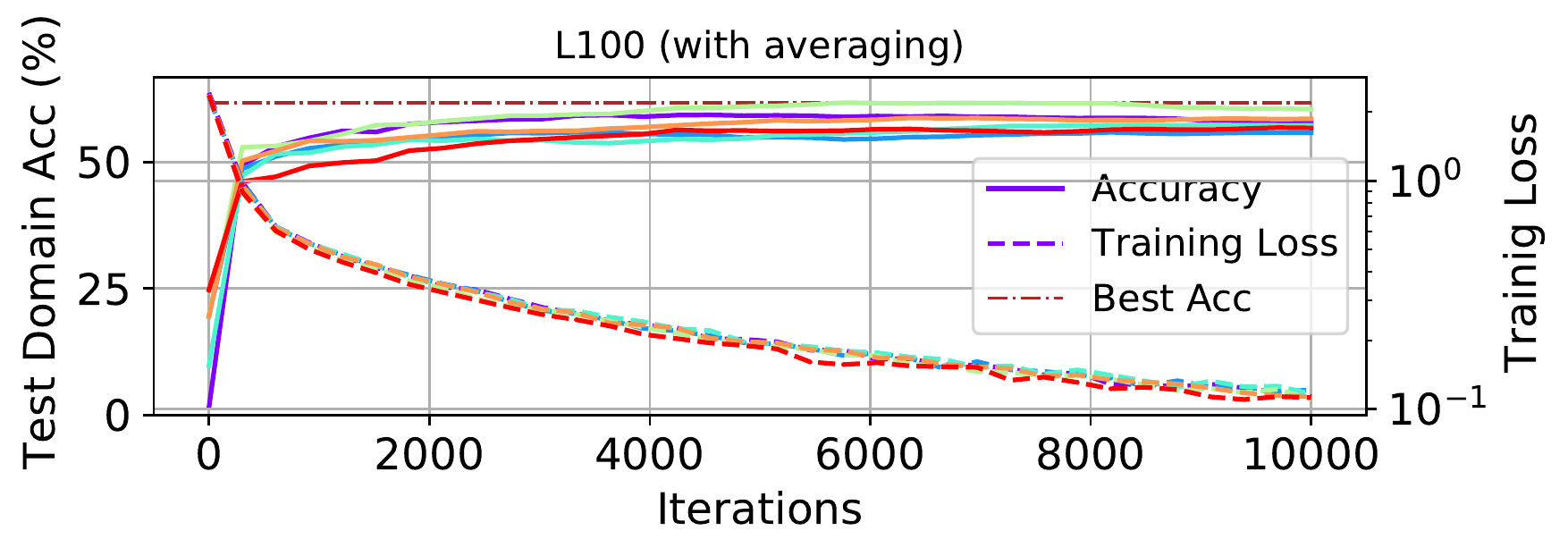}
      \vfil
      \includegraphics[width=0.36\columnwidth]{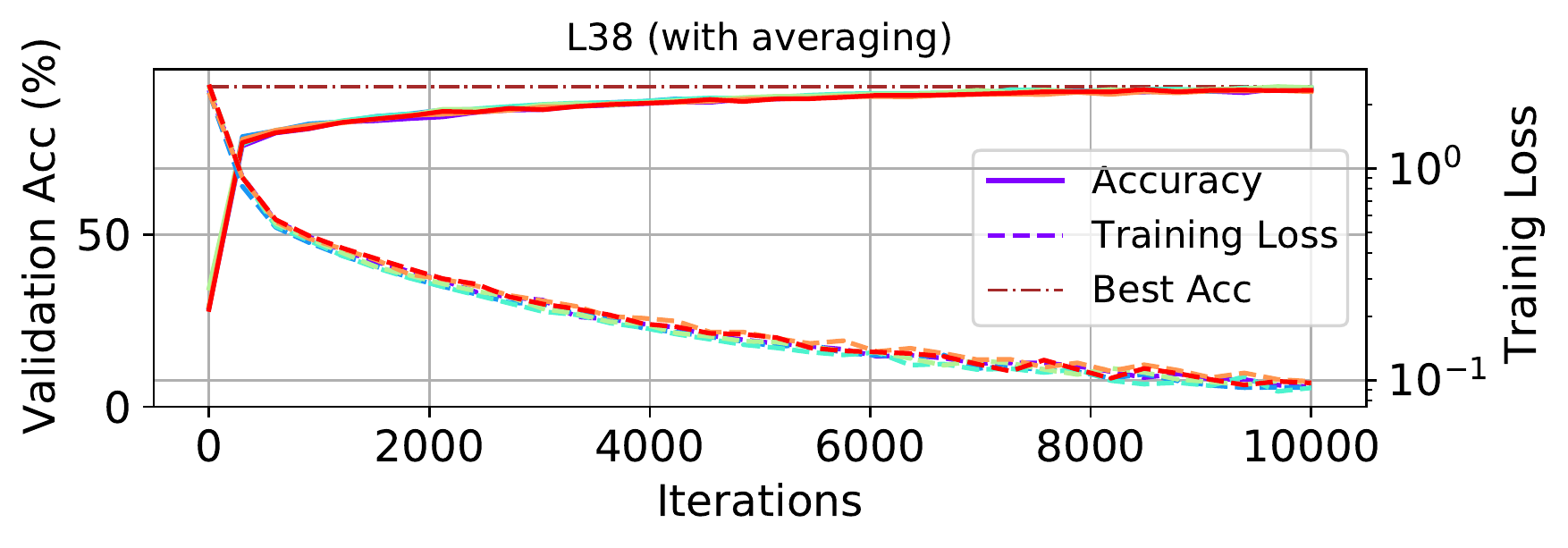}
    \hfil
      \includegraphics[width=0.36\columnwidth]{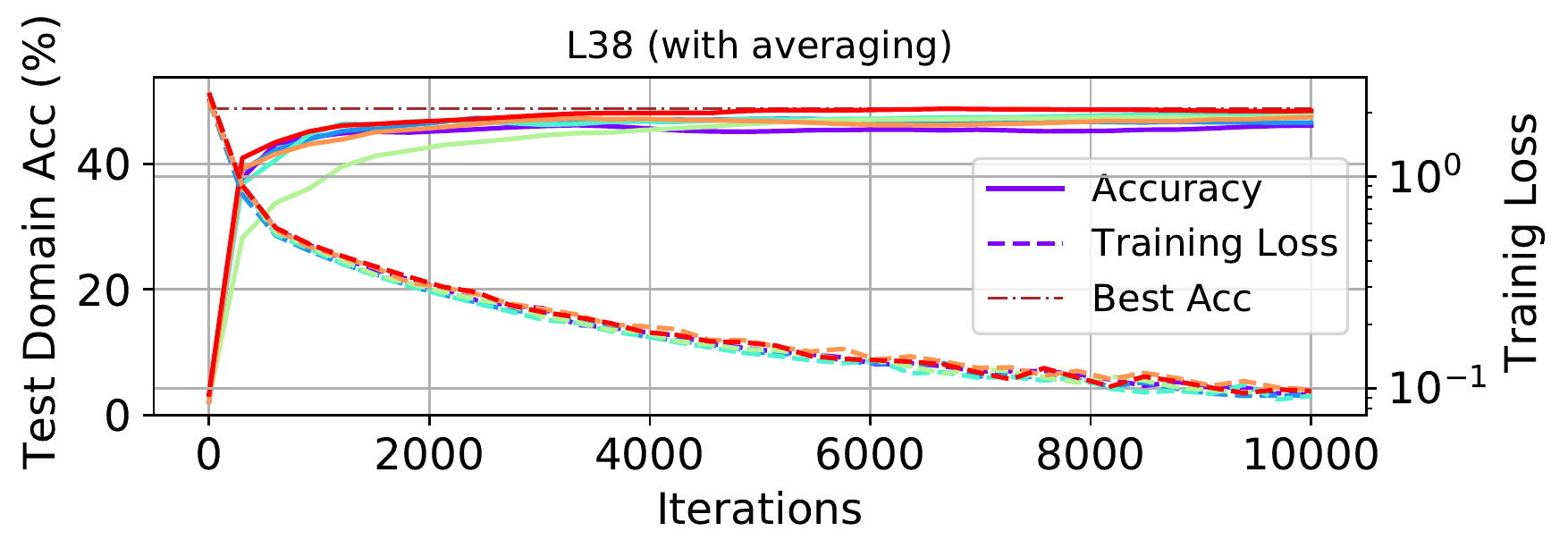}
      \vfil
      \includegraphics[width=0.36\columnwidth]{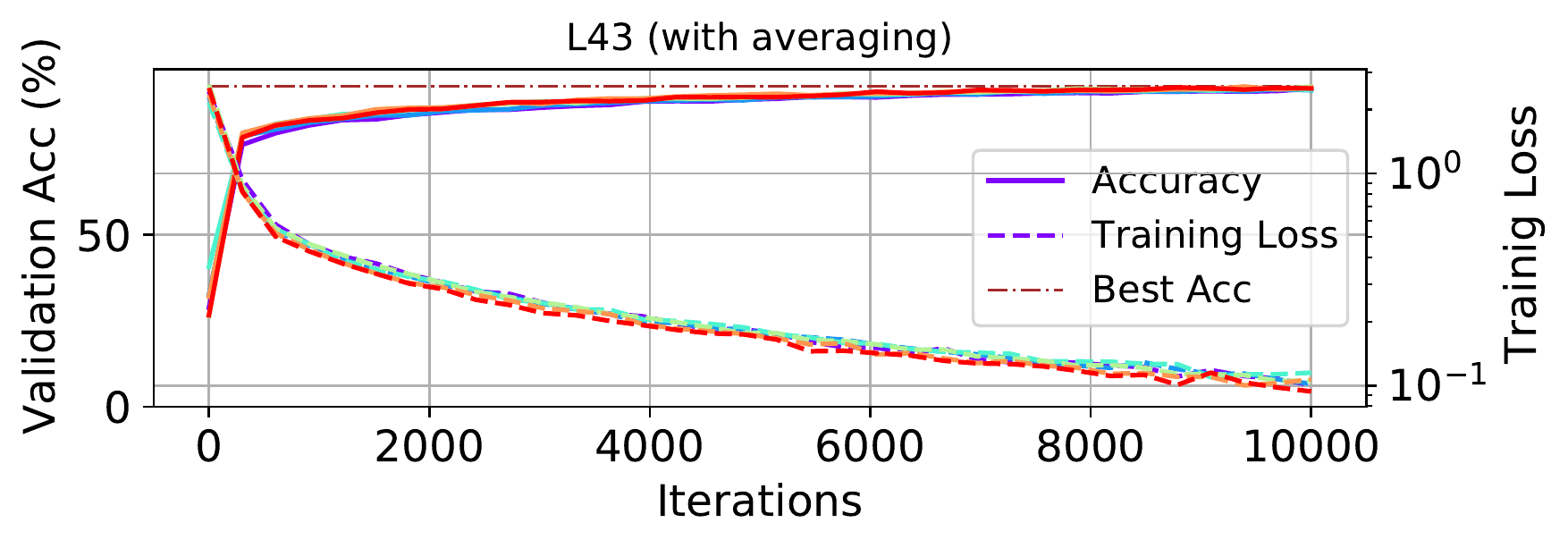}
    \hfil
      \includegraphics[width=0.36\columnwidth]{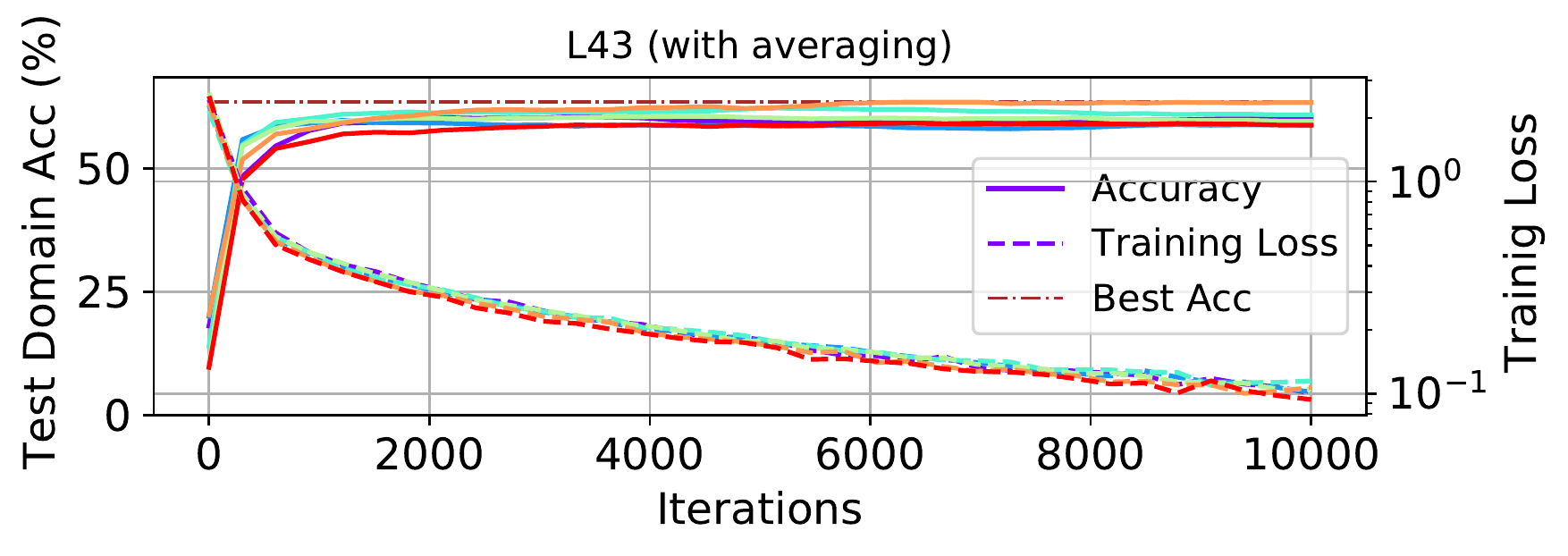}
      \vfil
      \includegraphics[width=0.36\columnwidth]{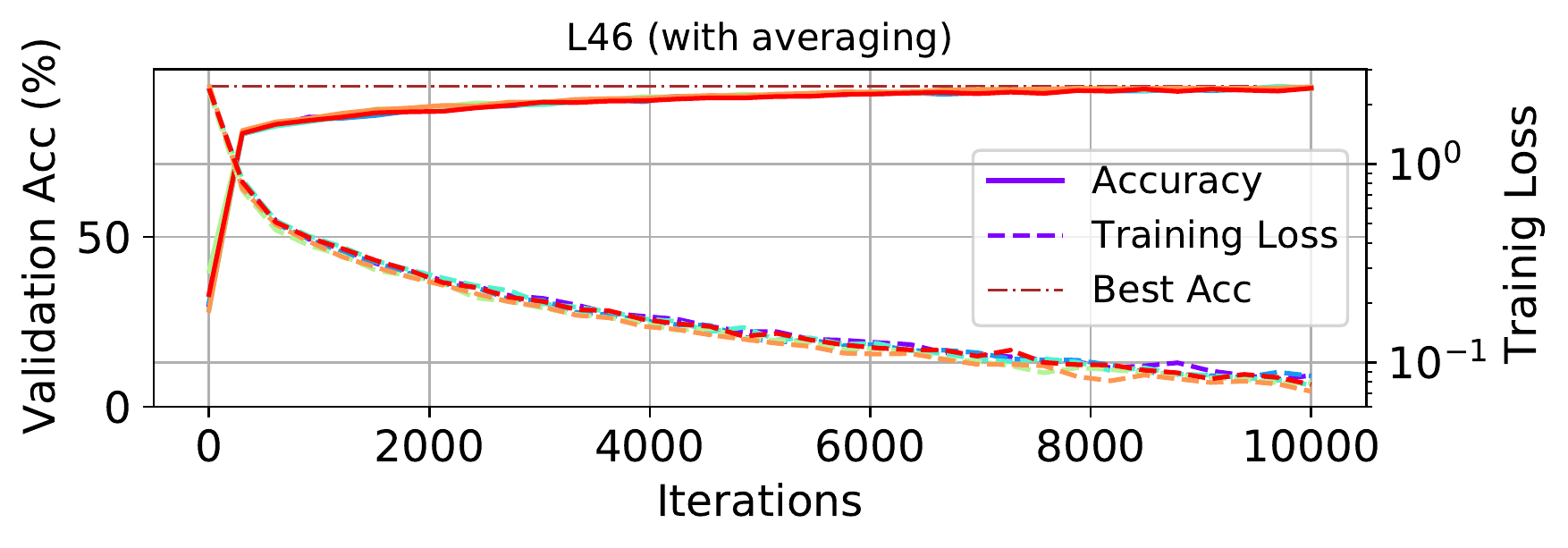}
    \hfil
      \includegraphics[width=0.36\columnwidth]{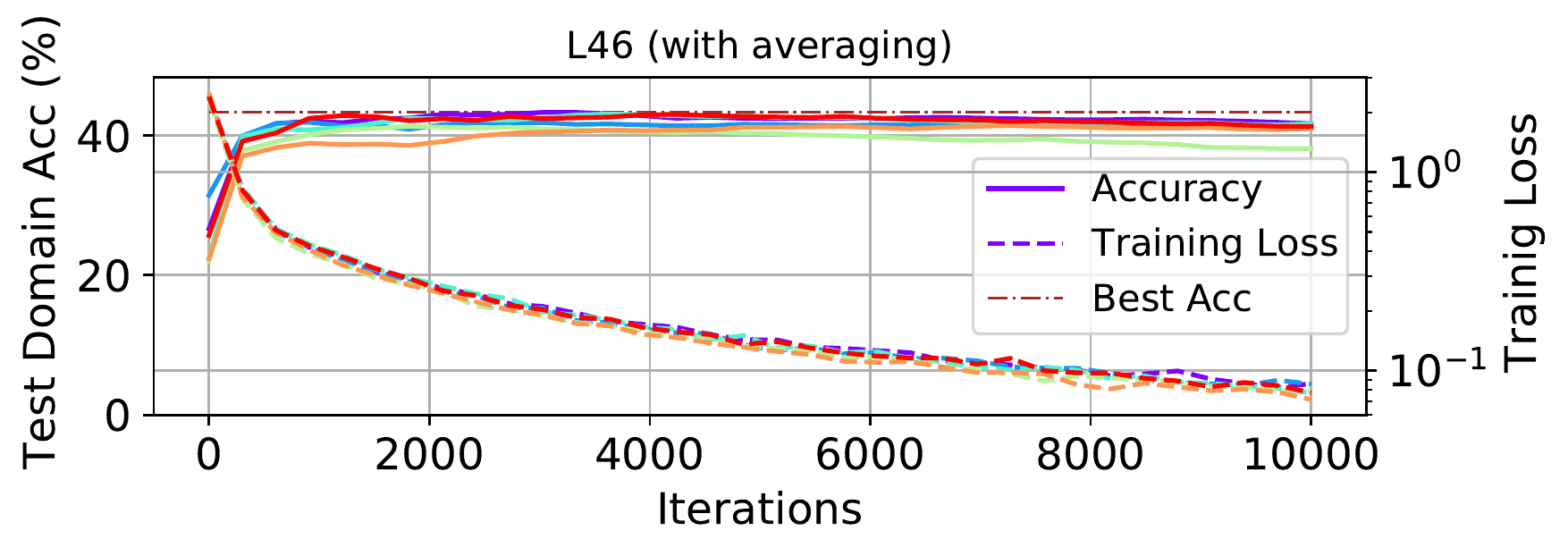}
    \caption{Evolution of training loss, in-domain validation accuracy and out-domain test accuracy for ResNet-50 (pre-trained on ImageNet) trained on TerraIncognita without model averaging (top 4 rows) and with model averaging (bottom 4 rows) for $10,000$ iterations with the domain mentioned in the title used as test domain and remain domains as training/validation data. Each color represents a different run with randomly chosen seed, hyper-parameters and training-validation split following \cite{gulrajani2020search}. \textbf{Gist}: Out-domain test performance is unstable without model averaging, which causes problem for model selection using in-domain validation performance. Model averaging is able to mitigate this instability.}
    \label{fig:terra_evolution_all}
\end{figure*}

\end{document}